\documentclass[12pt,letterpaper]{report}
\newcommand{\thesistitle}{Detecting Dead Weights and Units in Neural Networks}
\newcommand{\thesisauthor}{Utku Evci}
\newcommand{\thesisadvisor}{Professor Leon Bottou}
\newcommand{\graddate}{May 2018}
\newcommand{\thesisdedication}{Feed your head}

\usepackage{setspace}

\usepackage[final]{graphicx}

\usepackage{adjustbox}
\usepackage{subcaption}
\usepackage{listings}

\usepackage{indentfirst}
\usepackage{hyperref}

\usepackage{amsthm} 
\usepackage[tbtags]{amsmath} 

\usepackage{amsfonts}
\usepackage{amssymb}
\theoremstyle{definition}
\newtheorem{definition}{Definition}[section]

\begin{document}

\pagenumbering{roman}
%
\thispagestyle{empty}
\begin{center}
  {\large\textbf{\thesistitle}}
  \vspace{.7in}

  by
  \vspace{.7in}

  \thesisauthor
  \vfill

\begin{doublespace}
  A thesis submitted in partial fulfillment\\
  of the requirements for the degree of\\
  Master of Science\\
  Computer Science Department\\
  New York University\\
  \graddate
\end{doublespace}
\end{center}
\vfill

\noindent\makebox[\textwidth]{\hfill\makebox[2.5in]{\hrulefill}}\\
\makebox[\textwidth]{\hfill\makebox[2.5in]{\hfill\thesisadvisor\hfill}}
\newpage
\thispagestyle{empty}
\vspace*{0in}
\newpage

\vspace*{\fill}
\begin{center}
  \thesisdedication\addcontentsline{toc}{section}{Dedication}
\end{center}
\vfill
\newpage
\section*{Acknowledgements}\addcontentsline{toc}{section}{Acknowledgements}
%
During the process of working towards this work and especially towards the end, I kept asking myself how and why I decided to devote my last semester in NYC writing this thesis. I could take classes instead and have a less stressful last semester. This work was optional and at the time I had already accepted a software development engineer position and It seemed kind of unnecessary. However, I decided to do it for various reasons which are mostly related to a possible future Ph.D. application at the end. Looking back now, I think it was one of the best decisions I have made during my time at NYU.

NYC is a special city and I was one of the people floating in this urban jungle for a while. NYC is grand and there is no time to waste. You can always fail here spending a fortune or work more and more to make it here, you can't even feel comfortable. There are always things to do. Sometimes it feels like what NYC offers is not meant for you. You feel like you are missing stuff all the time and everyone else is doing better than you and you don't have any friend here. It is hard to live here and it is even harder to leave. I will miss waking up on Christopher, biking to CDS on 5th Avenue and spending all of my days on the 5th floor, in front of the computer. I will definitely miss having coffee in Blue Bottle Coffee and Joe's enjoying the first days of the spring.

I would like to thank Jason Weston for his feedback on SVM feature selection. Levent Sagun provided quite valuable comments and feedback about the first two chapter. I would like to thank him for his valuable advice and support during my studies at NYU. He has an important influence on my involvement in Machine Learning research.

Computer Vision class thought by Prof. Fergus was the first time I heard about the pruning idea and I still remember the excitement I had after his class about model compression. After his class, I decided to reproduce the recent pruning research as my class project and had valuable comments from him. I like to thank Prof. Fergus for being my the second reader for the thesis and all his support.

Finally, I would like to thank Prof. Bottou for all his patience and time advising this work. His guidance shaped this thesis and made it something readable. I learned many things about Machine Learning through our conversations and his reading suggestions. More important than that I learned how to motivate my work, how to visualize data and how to decide which experiments to do. He guided me towards the dead unit idea and introduced me to the bias propagation idea. Efficient approximation of the loss change due to this propagation (MRS) was also his idea. This work would have been quite different without his guidance and support. I feel lucky.

\newpage
\section*{Abstract}\addcontentsline{toc}{section}{Abstract}
%
Deep Neural Networks are highly over-parameterized and the size of the neural networks can be reduced significantly after training without any decrease in performance. One can clearly see this phenomenon in a wide range of architectures trained for various problems. Weight/channel pruning, distillation, quantization, matrix factorization are some of the main methods one can use to remove the redundancy to come up with smaller and faster models.

This work starts with a short informative chapter, where we motivate the pruning idea and provide the necessary notation. In the second chapter, we compare various saliency scores in the context of parameter pruning. Using the insights obtained from this comparison and stating the problems it brings we motivate why pruning units instead of the individual parameters might be a better idea. We propose some set of definitions to quantify and analyze units that don't learn and create any useful information. We propose an efficient way for detecting dead units and use  it to select which units to prune. We get 5x model size reduction through unit-wise pruning on MNIST. 

\newpage
\tableofcontents

\listoffigures\addcontentsline{toc}{section}{List of Figures}
\newpage

\listoftables\addcontentsline{toc}{section}{List of Tables}
\newpage

\pagenumbering{arabic} 
\section*{Introduction}\addcontentsline{toc}{section}{Introduction}
With increasing availability of data and computing power, Neural Networks became the most successful tool for learning complex functions. Networks with millions of parameters can be trained through first-order, gradient-based algorithms to minimize a given loss function. In the history of neural networks, many research proposed various architectures and optimization methods to come up with better results. With the increasing computation power, researchers were able to try larger architectures and utilize larger datasets to achieve significant improvements on many benchmarks and practical problems. A downside of these very successful and very big networks is that each network needs a considerable amount of storage and compute power to run. However, we can easily show that the resulting deep networks are highly redundant and had ill-conditioned parameter matrices. Can we reduce this redundancy by removing unnecessary parameters? Optimal Brain Damage paper\cite{lecun90} emphasizes the importance of removing connections (pruning) as a regularizer and a performance-enhancer. It proposes deleting parameters with small saliency and shows that second-order approximations might improve over simple magnitude based saliency measures. It also shows the similarity between weight decay\cite{hanson1989} and pruning.

In order to utilize this very large networks in resource-limited real-world environments like mobile devices, we should be able to reduce their storage and energy requirements. Han showed that the famous ImageNet winner AlexNet could be compressed up to 40x\cite{han2015a}\cite{han2015b}. However since there is no restriction on which connection to remove, pruned parameter matrices do not follow any specific pattern and therefore require sparse representation to utilize sparsity. One drawback with this approach is that the current sparse matrix operations are not as fast as their dense counterparts. Attacking this issue, various groups focused on channel pruning and introduced some structured sparsity techniques\cite{wen2016},\cite{molchanov2016},\cite{ye2018}.

It is very interesting to see that one can get around 90\% sparsity with retraining on many different networks and it looks like this number does not depend on architecture or data. Even a small network with a very small number of convolutional filters end-up having most of its filters dying out during training and we know that starting the training with an overparameterized network is crucial for the best performance. It looks like our optimization methods are not perfect and they are far from utilizing the full power of our networks.

Given a resource and time budget what is the best performance we can get? We can always use the biggest network we can run and then prune, quantize, distill or use other techniques to reduce the size of the network for the repetitive inference task. However, if we know that these low importance connections appear during the early training, we can prune the network during the training. Doing this would bring us no advantage in terms of computation time since the dense operations usually run much faster than the sparse operations below 90\% sparsity. So we are better off keeping the dense representation and doing unnecessary calculations on the pruned parameters during the training even if we prune the network early in the training. However, if we prune the whole channels or units at once, we can reduce the size of the parameter matrices right away and decrease the training time without harming the performance.

This thesis demonstrates, compares and criticizes the two main pruning approaches mentioned above: parameter-wise(unconstrained) pruning and unit-wise pruning. We also focus on finding\textbf{dead regions: weights/units that are likely to be pruned} that appear the early-training and stay dead after.

In Chapter \ref{chap1} , we provide a short introduction to neural networks and provide a literature review on pruning neural networks. In Chapter \ref{chap2} we motivate the idea of dead units by showing that low saliency parameters seem like developing around some particular units. We also compare various saliency measures and conclude that the magnitude-based saliency measure performs best at pruning a group of parameters at once. In Chapter \ref{chap3} we formally define the dead units as units with low mean replacement penalty (MRP: defined in Chapter \ref{chap3}). We propose using the first order approximation in order to detect the dead units efficiently. We generate dead units by introducing high learning rate for an arbitrary unit during some training and show that the unit most likely stays dead throughout rest of the training. After defining and generating dead units, in Chapter \ref{chap4}, we repeat the experiment in Chapter \ref{chap2}, but this time pruning entire units at once using the efficient scoring function defined in the previous chapter. As a result, we get a greater reduction in network size without requiring a sparse format. 

\chapter{Background\label{chap1}}

This chapter introduces necessary notation, concepts and presents the previous work.
\section{A Very Small list of Abbreviations\label{sec:1.1}}
\begin{enumerate}
  \item ANN: Artificial Neural Network
  \item CNN: Convolutional Neural Network
  \item GD: Gradient Descent
  \item SGD: Stochastic Gradient Descent
  \item RNN: Recursive Neural Networks
  \item MRP: Mean Replacement Penalty
  \item MRS: Mean Replacement Saliency
  \item \#: number of
\end{enumerate}

\section{Artificial Neural Networks and Notation\label{sec:1.2}}

Artificial Neural Networks (ANN) is a biologically inspired algorithm for transforming numbers
from an input space $X$ to an action space $A$.

\begin{figure}[ht]
  \begin{center}
    \includegraphics[width=8cm]{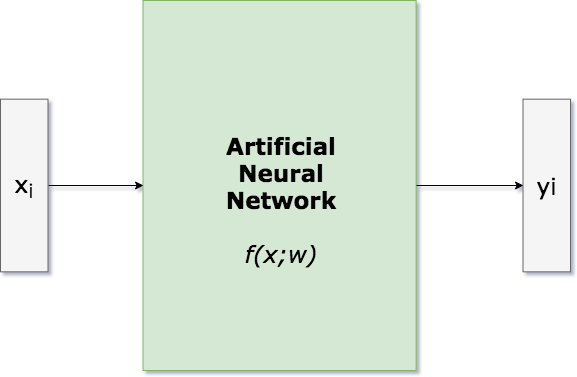}
  \end{center}%
\caption[ANN idea]{A very basic illustration of an ANN as a block diagram }
\label{fig:1.ann}
\end{figure}

With the advancements in Neuroscience and the invention of computers, the idea of mimicking neurons in our brains with the help of computers attracted quite a bit attention. In 1958, Rosenblatt introduced the building block of ANN's: the Perceptron\cite{Rosenblatt1958}. In the following years, advancements in the field made it possible to connect many perceptrons \textbf{units} together and train them jointly. With the advancements in acceleration hardware and increased availability of data; deep and large ANN's became the best performing algorithm in a wide variety of machine learning tasks. ANN's can have hundreds to millions of units and they are mostly trained through backpropagation. I would like to point out Schmidhuber's review \cite{schmidhuber2015} for further reference on the history of ANN's.

One can formulate an ANN as a function parameterized with a vector $w$, mapping an input $x_i$ to an output $\hat{y_i}$, where $i$ represent the index of the data from a dataset $X$ [Figure \ref{fig:1.ann}].

$$f(x_i;w)=\hat{y_i}$$

Then we would define a loss function with a range of non-negative real numbers calculated from the individual $\hat{y_i}$'s.
Usually there are ${y_i}$'s we wish $\hat{y_i}$'s to match.
$$L(w)=\frac{1}{m}\sum_{x_i,y_i \in D}l(f(x_i;w),y_i)$$
where $m$ is the total number of samples and $l(\cdot)$ is a non-negative function penalizing bad estimations.

In the supervised setting, this would correspond to the observed target values.
We denote them with $y_i$ and the corresponding target dataset as $Y$. In unsupervised methods, the loss function would depend on only $X$ with size $N$. We represent the loss function as $L(w)$ removing the data that the loss function intrinsically depends on from the notation.

ANN's can have many different layers/modules inside them. Recursive Neural Networks(RNN)'s allow units to have states and those states are used to calculate the output along with inputs. Units in RNNs learn both how to transform the input, given the state and how to update its state, given the input. Convolutional Neural Networks(CNN)'s are able to utilize spatial dependency in inputs. They use convolution operation and parameter sharing to learn how to extract  translation-invariant features from the inputs. The output of CNN's and RNN's are usually transformed further by some matrix transformation and this simple layer is usually named as \textbf{Linear} layer. Most of the modules mentioned above would also have nonlinearities applied to their outputs. There is not a strict line between all of these definitions and usually real-world architectures include a combination of different modules and nonlinearities; but they all have one thing in common: All the operations used in these networks are sub-differentiable; so that we can apply chain rule to get $\frac{\partial L(w)}{\partial w_i}$; i.e. gradient of the loss function with respect to every parameter in the network. Efficient calculation of $\frac{\partial L(w)}{\partial w_i}$ for every parameter in the network is named as backpropagation algorithm \cite{rumelhart1986} and it is a fundamental method for optimizing ANN's.

\subsection{Training DNN's\label{sec1.2.1} }

Optimizing the loss function of an ANN or in other words training an ANN is hard due to its non-convex energy landscape and its huge dimensionality. Almost all ANN's are trained with some version of gradient descent(GD). GD uses back-propagation algorithm to calculate $\frac{\partial L(w)}{\partial w_i}$'s and then updates the parameters with the following formula. The calculation of the $f(x_i;w)$'s for a given dataset is called \textbf{forward-pass}. Then we would calculate the average loss $L(w)$. Using chain rule we would then calculate the gradient of the loss $\frac{\partial L(w)}{\partial w}$ and this is called the \textbf{backward-pass}. After calculating the gradient we would update the parameters with the following rule.

$$w_i^t=w^{t-1}_i-\alpha\frac{\partial L(w^{t-1})}{\partial w^{t-1}_i}$$

where $\alpha$ is a small constant called \textbf{learning rate} and $t$ in the upper-script denotes the time and we call each of these updates as \textbf{step}. Remember that the loss above calculated over the entire dataset $X$. However, the size of a dataset hits easily millions in today's problems and calculating the exact gradient at each step poses technical difficulties. Therefore stochastic version of GD(SGD) is used to approximate the gradient direction over a subset of the dataset $X$ instead of the complete dataset. We name these subsets as mini-batch and denote them as $X_t$ emphasizing its dependence to step number. Number of elements in $X_t$ is called as \textbf{batchsize} and denoted with $m$. At each step optimization algorithm chooses a random mini-batch $X_t$ and performs the update with the partial derivatives calculated over this mini-batch. A complete-pass over the dataset is called an \textbf{epoch}.

\subsection{Taylor Approximation of Loss\label{sec1.2.2} }

To understand why GD or SGD might work, let's write down the quadratic approximation of our loss function around the point $w$.
Assuming the loss value is calculated each time with the complete dataset
($X,Y$) or with a mini-batch ($X_t,Y_t$) and the parameter set is represented by vector $w$ of size $M$ the approximation of $L(\dot w)$ would be:
\begin{equation} \label{eq1.1}
L(\dot w)=L(w)+(\dot w-w)^T\nabla L(w) + \frac{1}{2}(\dot w-w)^TH(w)(\dot w-w) + O((\dot w-w)^3)
\end{equation}

where H(w) is an $M$ by $M$ matrix with elements $ h_{i,j}=\frac{\delta L(w)}{\delta w_i \delta w_j}$.

Let's imagine we want to calculate how much the loss function will change if we change some weights
by $\Delta w$  such that new weights: $\dot w = w+\Delta w$. Substituting above gives us

\begin{equation} \label{eq1.2}
L(\dot w) = L(w)+\Delta w^T \nabla L(w) + \frac{1}{2}\Delta w^T H(w) \Delta w+O(\Delta w^3)
\end{equation}

and the change in the loss function $\Delta L= L(\dot w)-L(w)$is given by
bringing the original loss to the left side.

\begin{equation} \label{eq1.3}
\Delta L = \Delta w^T \nabla L(w) + \frac{1}{2}\Delta w^T H(w) \Delta w+O(\Delta w^3)
\end{equation}

We can similarly write the change in the loss using the first-order approximation and we would get following.

\begin{equation} \label{eq1.4}
\Delta L = \Delta w^T\nabla L(w) + O(\Delta w^2)
\end{equation}

Note that the error term $O(\Delta w^2)$ increases polynomially with $\Delta w$ and we need to ensure that this error term is small if we want to do anything with this approximation. Let's pick $\Delta w$ as the gradient $\nabla L(w)$ itself. Then we would have a $\|\cdot\|^2$ norm of the gradient which is guaranteed to be positive. However we want to decrease the loss and ensure that $\Delta w$ is small, so let's scale the direction with a small negative number $\alpha$. Then we get what we need.

$$\Delta L \approx \alpha \|\nabla L(w)\|$$
$$\Delta L < 0$$

Having a step update rule on the negative gradient direction guarantees a decrease in the loss for a sufficiently small $\alpha$. However when we use gradient calculated over a mini-batch note that our gradient directions are noisy and there is no such guarantee. However with some tricks and right set of tools SGD is undoubtedly the most successful tool for optimizing ANN's.

The goal of optimization is to minimize the loss over a dataset $X$ and we called this dataset as \textbf{training dataset}. However, we are usually interested
in minimizing the generalization error, which is usually approximated with a separate \textbf{test dataset}.

One can use various regularization methods to ensure a good generalization and it is usually believed/observed that ANN's trained with SGD generalizes well through some implicit(e.g. early stopping) and explicit(e.g. Dropout \cite{hinton2012}) regularizer methods. However, all of these known sources are not enough to completely explain the generalization performance of neural networks \cite{kumar2017} and why ANN's generalizes so well is one of the biggest mysteries of the ANN's.

\subsection{Units(Neurons): Building block of DNN's \label{sec1.2.3} }

\begin{table}[ht]
\begin{center}
\begin{tabular}{|c|c|c|}
  \hline
  \textbf{Parameter Name} & \textbf{Tensor Shape} & \textbf{$W_i^l$ or $b_i^l$}  \\
 \hline
 Linear.weight &$I*O$ &  $[:,i]$\\
 \hline
 Linear.bias & $O$ & - \\
 \hline
 Conv2d.weight & $I*O*H*W$ &$[:,i,:,:]$\\
 \hline
 Conv2d.bias & $O$&- \\
 \hline
\end{tabular}
\end{center}
\caption[Tensor shapes of various layer parameters]{Tensor shapes of various layer parameters.
 $I$: input features, $O$: output features, $W$: width of the filter, $H$: height of the filter}
\label{table:1.1}
\end{table}

As mentioned before ANN's consists of cascaded layers. Lets partition our parameter vector $w$ into matrices $W^l$ and $b_l$ for each layer where $l$ denotes the layer number and $b$ is the $bias$ vector. The shape of $W^l$ depends on the type and size of the layer. By convention, the first two dimensions are used for \#input features ($I$)and \#output features ($O$) respectively. For convolutional layer, the third and fourth dimension would represent the filter width ($W$) and height ($H$) (Table \ref{table:1.1}). Each part in these layers that does a scaled summation of input features called as a \textbf{unit}(also neuron) and their parameters are denoted as $W_i^l$ and $b_i^l$. We denote this summation value with $h_i^l$
$$h_i^l=W_i \cdot a_i^{l-1}+b_i^l$$

The values generated at each unit is then passed to some non-linearities denoted with $g(\cdot)$. Value after the nonlinearity called as the \textbf{activation} $a_i^l$ of a unit.
$$a_i^l=g(h_i^l)$$

Units are defined to be the independent components of a layer. However, they might get coupled as a result of some non-linearities which combine outputs from many units like SoftMax.

\subsection{Saliency and Pruning\label{sec1.2.4} }
\textbf{Saliency} $S_i$ of a parameter $w_i$ is used to capture the contribution of individual parameters to the loss.
$$S_i = L(w_{-i})-L(w)$$
where $w_{-i}$ is the same vector as $w$ except $w_i$ is replaced with 0. A high saliency would mean that removing the parameter would introduce a high penalty on the loss functions, so it has a high contribution. On the other hand, a small saliency would mean that the contribution of the parameter is small(or even negative!) and can be safely removed. So basically, saliency captures the \textbf{penalty} for removing a parameter from the network. Removing a fraction of parameters (setting them to zero) according to the ranking of saliency scores is called \textbf{pruning}. Pruning is usually done after training as an inference optimization to make the network smaller. However, one can also prune the network during the training to improve on final performance or reduced the training time. Various saliency measures trying the approximate the quantity above are compared and explored in the next chapter(Chapter \ref{chap2}).

Pruning parameters according to the parameter-wise saliency above assumes independence between parameters. However, this independence assumption is not correct. Depending on the network architecture pruning a single parameter would change the saliency ranking of the remaining parameters and therefore make the initial ranking invalid. To prevent this effect, one can perform iterative pruning, where each time the parameter with the smallest saliency score( or smallest k of them) is removed and the saliency scores are calculated again before each pruning iteration. However, this would be very time consuming considering the size of the ANN's in today's problems.

Another problem with parameter-wise pruning is that the resulting sparse parameters would be still stored in a dense tensor and the computational speed-up would not be proportional to the pruning factor. One can decide to represent the pruned tensors in a sparse format; however, this would not provide speed-up unless pruning factor becomes considerably high.

An alternative to the parameter-wise pruning is to prune units. If we can calculate saliencies for units, i.e. penalty of removing the whole unit, we can remove units from the network. Removing a whole unit would put us in a much better position at doing iterative pruning (number of units is much smaller than the number of parameters) and replace the need for fast sparse operations. However, we can do even better.

In what conditions units end up having a low penalty for removal? This may happen when the unit constantly generates 0's or all the fan-out weights are so small that the network ignores the unit. It could be also possible that all fan-out units themselves have very high magnitude. What if the unit generates a value $c$ with very low variance. If a unit constantly generates similar values then we can subtract the mean value it generates from its bias and we can propagate the same value to the fan-out units $W_j^{l+1}$ by updating the bias of the units with the following rule.
\begin{equation}\label{eq1.5}
b_j^{l} = b_j^{l}-\mathbb{E}[y_i^l]
\end{equation}

\begin{equation}\label{eq1.6}
b_j^{l+1} = b_j^{l+1}+g(\mathbb{E}[y_i^l]*W_{j}^{l+1})
\end{equation}

After doing this \textbf{bias-propagation} we can now have our saliency measures calculated again and we would get a very low penalty for removing the particular unit $i$, since it would generate always zeros. So, we can do better than just removing the units by calculating the \textbf{mean-replacement penalty} for each unit and applying bias propagation during removal. My thesis focuses on this saliency measure and we investigate our strategy in Chapter \ref{chap2} and Chapter \ref{chap3}.

\section{Literature Review\label{sec:1.3}}
The magnitude of weights proposed as a simple measure of saliency in the earlier literature and its connection to the structured weight decay highlighted in \cite{hanson1989}. \cite{lecun1990} used second-order Taylor approximation at minima along with the diagonal approximation of the hessian to calculate saliencies for network parameters and emphasized the importance of pruning as a regularizer and performance-optimizer. Later \cite{hassibi1993} used inverse of the hessian to get a better approximation of saliencies for network parameters. The experiments performed in these early works were relatively small tasks like the famous XOR problem.

Later it was the ice age for the neural networks and the majority of the research focused on the support vector machines(SVM). Various feature elimination methods are utilized to find out the minimal set of relevant features that give the same performance on the dataset. For a review on feature elimination and a reference paper, we would like to direct the reader to \cite{kohavi1997} and \cite{guyon2002}.

With increasing computational power, interest in neural networks increased again towards the second decade of 20th century. Many research attacked the difficulties we face training the neural networks and come-up with a bag of tools that helped us training bigger networks. With the increasing amount of data, tricks, and architecture, neural networks started to dominate the field again.

With the increased size of networks, many research highlighted the redundancy in the parameters. \cite{denil2013} showed that the whole network can be predicted from a small fraction of parameters. We can attack this redundancy in various ways and try to minimize the size of a network. The main motivation for doing so is to make deep neural networks available in hardware or time-limited platforms/scenarios. \cite{shazeer2017} trains very big gated network(Sparsely-gated Mixture-Of-Experts) consists of many small experts. It is more important now to learn how to make small networks as efficient as possible and we can always combine them.

One can reduce the size of a network by matrix decomposition \cite{denton2014}, pruning \cite{han2015a} \cite{han2015b}, quantized/low precision networks \cite{lin2015}, \cite{courbariaux2015}, \cite{rastegari2016}, \cite{polino2018} or distillation \cite{frosst2017},\cite{hinton2015}.\cite{denton2014} reports 2x speedup on forward propagation, \cite{han2015a} reports around 30x compression through pruning, weight quantization and Huffman coding. \cite{zhu2017} performs pruning during training and reports similar compression rates. \cite{liu2015} uses custom sparse-matrix multiplications after decomposition to speed up-convolutional layers. \cite{choi2017} compares various quantization methods and obtaining upto 50x quantization compression. \cite{achterhold2018} focuses on pruning bayesian neural networks.  \cite{galloway2018} shows that stochastic quantization helps preventing black box attacks on neural networks.

It seems like there is something that is going wrong during the training of neural networks, which high redundancy in our parameters and enables all the methods mentioned above to reduce the size of the networks.

\cite{wen2016} uses group lasso penalty to get some structured sparsity that enables reducing the size of dense tensors. \cite{molchanov2016} focuses on unit-wise pruning in the context of transfer learning and compared some first-order saliency measures using Spearman correlation coefficient. Their work focuses on pruning instead of the mean-score-replacement method, which we define in Section \ref{sec3.1} formally. \cite{ye2018} proposes the idea of forcing the units to produce constant values in a neural network and then does \textit{bias propagation} to remove them without a loss penalty which is basically the same idea we are proposing, since it is the generalized version of unit removal.

If a unit stops getting gradient (stuck in local minima) without doing anything useful. We should be able to detect this. This can help us to detect/find right architectures that align well with current optimization methods and provide us good training and validation losses. Therefore we believe that understanding behavior of units is crucial for answering other important question in neural network research. \cite{zhang2017} shows that ANN's with more parameters than the number of training samples trained with SGD show good generalization behavior. \cite{neyshabur2017} explores various complexity measures and proposes norm instead of count-based measures. \cite{morcos2018} ablate units (which mimics removal) in various units to understand their generalization behaviors. They argue that a high mutual information is crucial for a good generalization.

We would like to point the reader to \cite{cheng2017} for a review of current approaches and methods on network compression and acceleration.

\chapter{Weight-based pruning and saliency score comparison\label{chap2}}

We can measure the contribution of individual parameters to our loss value using the second order Taylor approximation (Equation \ref{eq2.1}). Following the notation in Section \ref{sec:1.2}, lets denote the change in the loss function with the value $\Delta L$ and lets denote the saliency of the $i^{th}$ parameters with $S_i$ where $w_i$ is the zero vector except  $\Delta w_i=w_i$.

\begin{equation} \label{eq2.1}
\Delta L = L(\dot w)-L(w) \approx \Delta w^T \nabla L(w) + \Delta w^T H(w) \Delta w
\end{equation}
\begin{equation} \label{eq2.2}
S_i \approx \Delta w_i (\nabla L(w)_i + \Delta w_i H(w)_{ii})
\end{equation}

This looks computationally plausible since during back-propagation, we calculate $\nabla L(w)$ and we know how to calculate an approximation of diagonal values of Hessian in O(N) time \cite{lecun1987}. However, saliencies calculated with this approximation may change if anything in these networks changes (i.e. if we prune one parameter). To be able to use this measure in the context of pruning we need to assume pairwise independency between parameters, i.e. changing one parameter would not affect another parameter's saliency. In this section, we show this is far from being true. To be able to capture dependencies and improve the saliency score we propose updating the Hessian part by including the sum of the whole row of the hessian.

\begin{equation} \label{eq2.3}
S_i \approx \Delta w_i (\nabla L(w)_i + \sum_j H(w)_{ij}\Delta w_j)
\end{equation}

The hessian term in equation \ref{eq2.3} can be calculated in O(n) time using Hessian-vector product \cite{pearlmutter1994} where n is the number of parameters in the network.

What should we set $\Delta w_i$ to? In the context of pruning, we remove weights, i.e. setting them to 0. So we need to set $\Delta w_i=-\epsilon w_i$. When $\epsilon=1$, this would correspond to weight removal. However, if $w_i$ is not small the Taylor approximation may have a high error. Therefore we may introduce an $\epsilon$ term. Note that changing $\epsilon$ would not affect the ranking in first order approximation and therefore we don't include $\epsilon$ in our first order scoring function.

In the case of general sensitivity analysis, we would set $\Delta w_i=\epsilon*sign(w_i)$ to see how much the loss would change if we make each weight a little closer to zero by a constant amount.

\begin{table}[ht]
\begin{center}
\begin{tabular}{|c|c|c|}
  \hline
  \textbf{Saliency Function} & $\boldsymbol{S_i}$ & \textbf{Runtime}  \\
 \hline
 taylor1Scorer & $\Delta w_i \nabla L(w)_i$ & 0.088s \\
 \hline
 taylor2Scorer & $\Delta w_i(\nabla L(w)_i+\lambda \sum_j H(w)_{ij}\Delta w_j)$&0.564s\\
 \hline
 hessianScorer & $\Delta w_i(\sum_j H(w)_{ij}\Delta w_j)$&0.483s \\
 \hline
 taylor1ScorerAbs & $abs(taylor1Scorer)$ & $-$ \\
 \hline
 taylor2ScorerAbs & $abs(taylor2Scorer)$ & $-$ \\
 \hline
 hessianScorerAbs & $abs(hessianScorerAbs)$ & $-$ \\
 \hline
 magnitudeScorer &$abs(w_i)$ &  0.00006s\\
 \hline
\end{tabular}
\end{center}
\caption[Saliency Score Definitions]{Summary of various saliency functions and their formula. In the context of pruning we would set $\Delta w_i=-w_i$ and $\lambda=1$. In sensitivity analysis we would set $\Delta w=-sign(w)$ and $\lambda=\epsilon$ where $\epsilon$ is a small number ensuring correctness of taylor approximation. Runtimes are averaged over 10 runs on MacbookPro 2015 calculating scores for all parameters of the network proposed in Section \ref{sec:2.1} using a validation set of size 1000 from the MNIST dataset.}
\label{table:2.1}
\end{table}

In this chapter, we show that saliency calculations based on the Taylor approximation fail at pruning a group of parameters at once. Magnitude based saliency score works better than the rest. We start by comparing the performance of various first and second order saliency measures used at pruning (Section \ref{sec:2.1}). Table \ref{table:2.1} summarizes these saliency measures derived from the equation \ref{eq2.3}. All the scores in the table are implemented in the \textsc{pytorchpruner} package (see Appendix \ref{app1}). Note that our implementation of second order scoring functions takes 6-7 times more time than a single back-propagation pass on CPU. Run times can be further optimized, but it is important to understand that the second order approximations are in the same time complexity as the first order ones due to the hessian vector product enabled by the auto-grad library of Pytorch \cite{pytorch}.

In Section \ref{sec:2.2} we demonstrate pruning a simple network trained on MNIST using magnitude-based scoring and show that pruned parameters have some kind of structure. They gather around some particular units and we leverage this observation and decrease the size of the network by half.

\section{Saliency Measures and Comparison\label{sec:2.1}}
In this section, we compare various saliency scores against each other. In theory, pruning individual weights can have a negative effect and we observe indeed this is the case. We then realize that the promised reduction in the loss can be overturned if we prune many weights together. This effect is observed more dominantly later in the training. This tells us that dependency among the individual scores increases over time.

In addition to the  scoring functions in Table \ref{table:2.1} we implemented two additional scoring functions. \textit{lossChangeScorer} calculates the empirical saliency for each parameter by setting them zero one by one and calculating the change in the loss. This is the value \textit{taylor1Scorer} approximates. \textit{randomScorer} assigns random scores to each parameter sampled uniformly from the range $(0,1)$, so that we can compare our saliency functions with random scoring.

Note that our primary goal is not to approximate \textit{lossChangeScorer}, but to have a low overall penalty when we prune a fraction $f$ of the parameters at once. To capture the dependency between parameters, we introduce \textit{taylor2Scorer} and \textit{hessianScorer}. Even though they don't approximate \textit{lossChangeScorer}, we hope these saliency measures would pick the least dependent parameters and therefore induce a smaller penalty at pruning. Comparing the penalties of various saliency measures (i.e. \textit{taylor1Scorer}, \textit{taylor2Scorer} and \textit{hessianScorer}) in the context of pruning, we observe that absolute values of these measures work better at pruning a fraction of the parameters at once. Finally, we show that \textit{magnitudeScorer} works even better than the absolute-valued measures most of the time and has much better stability.

In each run, we are training our CNN model for MNIST dataset (Appendix \ref{app3}) with 0.01 learning rate and 32 mini-batch size for 5 epochs. At every 50 step, we are calculating the pruning loss without changing the model itself for each scoring function separately. All data dependent scores are calculated on a validation set of size 1000\footnote{This might seem counter-intuitive since the validation set is not the data that we are directly optimizing for. However, our experiments and results don't focus on the comparison of various function and therefore the data we use doesn't make a difference and we confirmed that in separate experiments.}. Finally calculated scores are used to determine which parameters to prune for a given target fraction.

We generated many experiments using various models(i.e. models with ReLU and Tanh), different layers and various scoring functions. It is not feasible to share all of the results here; therefore we share only a subset of the results where we believe represents the whole. All models below uses Relu non-linearity unless stated otherwise.

\subsection{Negative Loss Changes and Correlation Between Parameters\label{sec2.1.1}}
\begin{figure}[ht]
  \begin{center}
    \includegraphics[width=.8\linewidth]{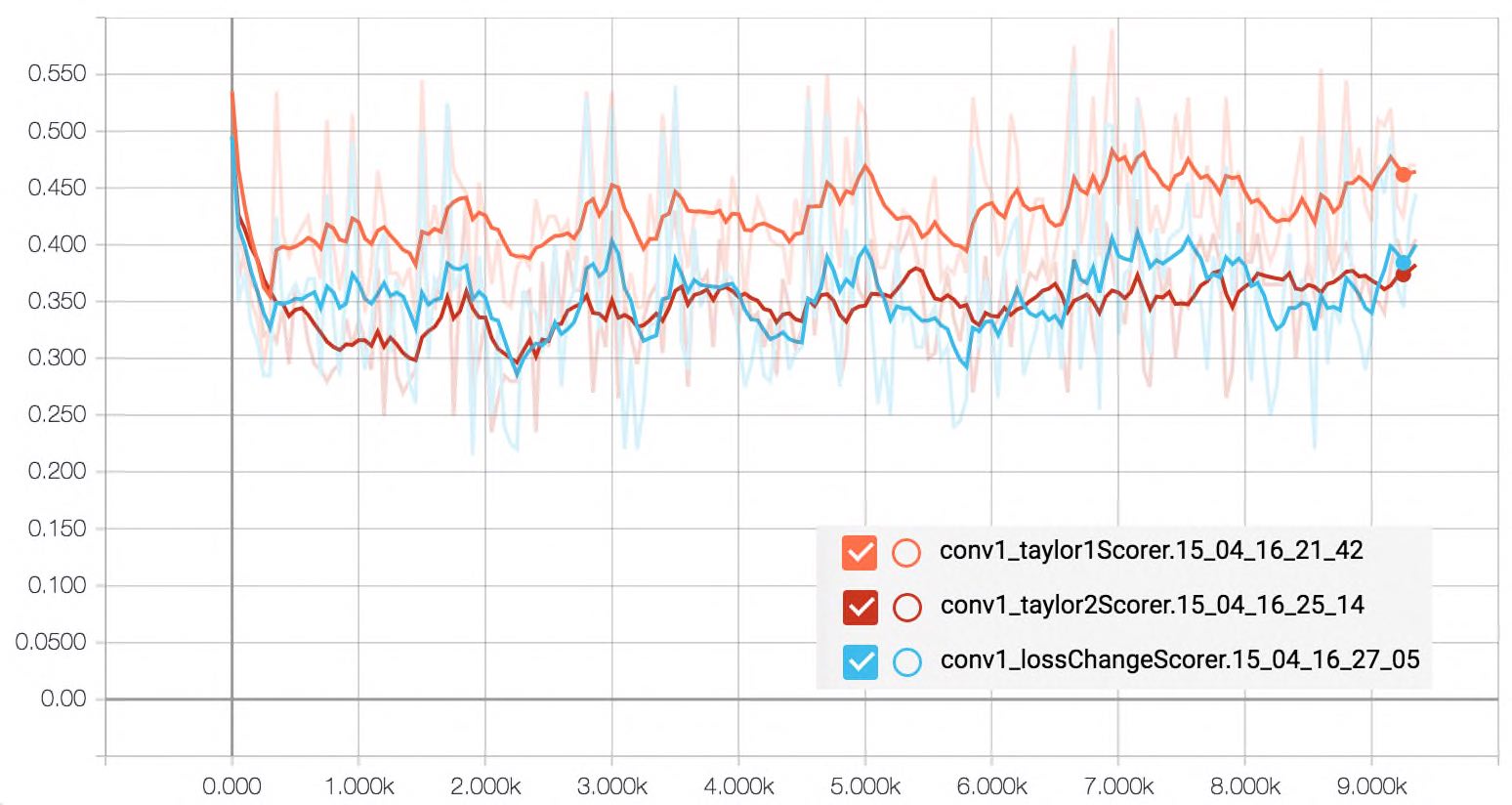}
  \end{center}%
\caption[Fraction of negative loss changes]{Horizontal axis represents the number of iterations passed. Vertical axis is the smoothed version of the fraction of negative saliencies for the first convolutional layer calculated with \textit{lossChangeScorer}, \textit{taylor1Scorer} and \textit{taylor2Scorer}. \ref{app3}}
\label{fig:2.1.1neg}
\end{figure}

Ideally, we would like to see a positive change in the loss when we prune a random parameter in a network. This would indicate that each parameter is doing something right and we can't give up any parameter. However, when we look at the results of the \textit{lossChangeScorer}, we see that constantly around 35\% of the parameters(of the first convolutional layer) cause a negative change in the loss function when removed (Figure \ref{fig:2.1.1neg}). This is very interesting since the fraction of negatively scored parameters stay constantly around 35\% throughout the training, which may hint us that the network can be pruned during the training, without waiting for the convergence and we may even get some improvements in the loss value. Even more striking, it seems like there are some parameters that are not getting the error signal they need to improve. There is something preventing them.

There might be two reasons for the negative loss change. One is that we are overshooting the minimum constantly. Which is not likely since we see the fraction of negative scores even in the early training. More likely is that the path from the parameter to the calculated loss is blocked and the parameter is not getting any gradient signal. Remember that the gradient of a parameter is given by

\begin{equation}\label{eq2.1.1}
\frac{\partial L}{\partial w_i } = \frac{\partial L}{\partial y_i} a_i^{l-1}
\end{equation}

where $a_i^{l-1}$ is the activation coming from the previous layer or the input tensor.

Looking at the back-propagation rule (Eq \ref{eq2.1.1}) there might be two reasons for very small gradient: input $a_i^{l-1}$ being zero and the output gradient $\frac{\partial L}{\partial y_i}$ being zero. Assuming that the parameter gets some non-zero input across many different samples and mini batches, the reason should be that the unit at the other end has zero gradient itself(i.e. $\frac{\partial L}{\partial y_i}=0$).

 Having no gradient signal cannot be due to converging to a global minima since a good fraction of the parameters has a negative effect on loss once removed. One thing to note here that we average gradients over a mini-batch and it is possible gradients from different samples to cancel each other and get a very small average gradient. This scenario is possible, however considering we repeatedly calculate saliencies and see the same picture it is very unlikely. Therefore we argue the scenario where individual samples introduce zero or close-to-zero gradient.

This may happen again for two reasons. One is that the unit is doing a good job and have a small gradient. Another reason could be that it doesn't get any error signal since outgoing weights are all small and the unit is dead. So how to ensure the $\frac{\partial L}{\partial y_i}$'s being zero only when the whole network is optimal and we are having gradient flowing across all connections. One way is to have skip connections \cite{resnet}.
Another possible way is to detect low $\frac{\partial L}{\partial y_i}$ regions and check whether this small gradient is caused by the fact that they learned their tasks or not. In the case of small mean replacement error, we should fix the problem, since otherwise, we would have so-called dead units; which doesn't do anything useful and prevent learning for its weights.

\subsection{Pruning can improve loss\label{sec2.1.2}}
\begin{figure}[ht]
  \begin{center}
    \begin{subfigure}{.8\textwidth}
    \centering
    \includegraphics[width=.8\linewidth]{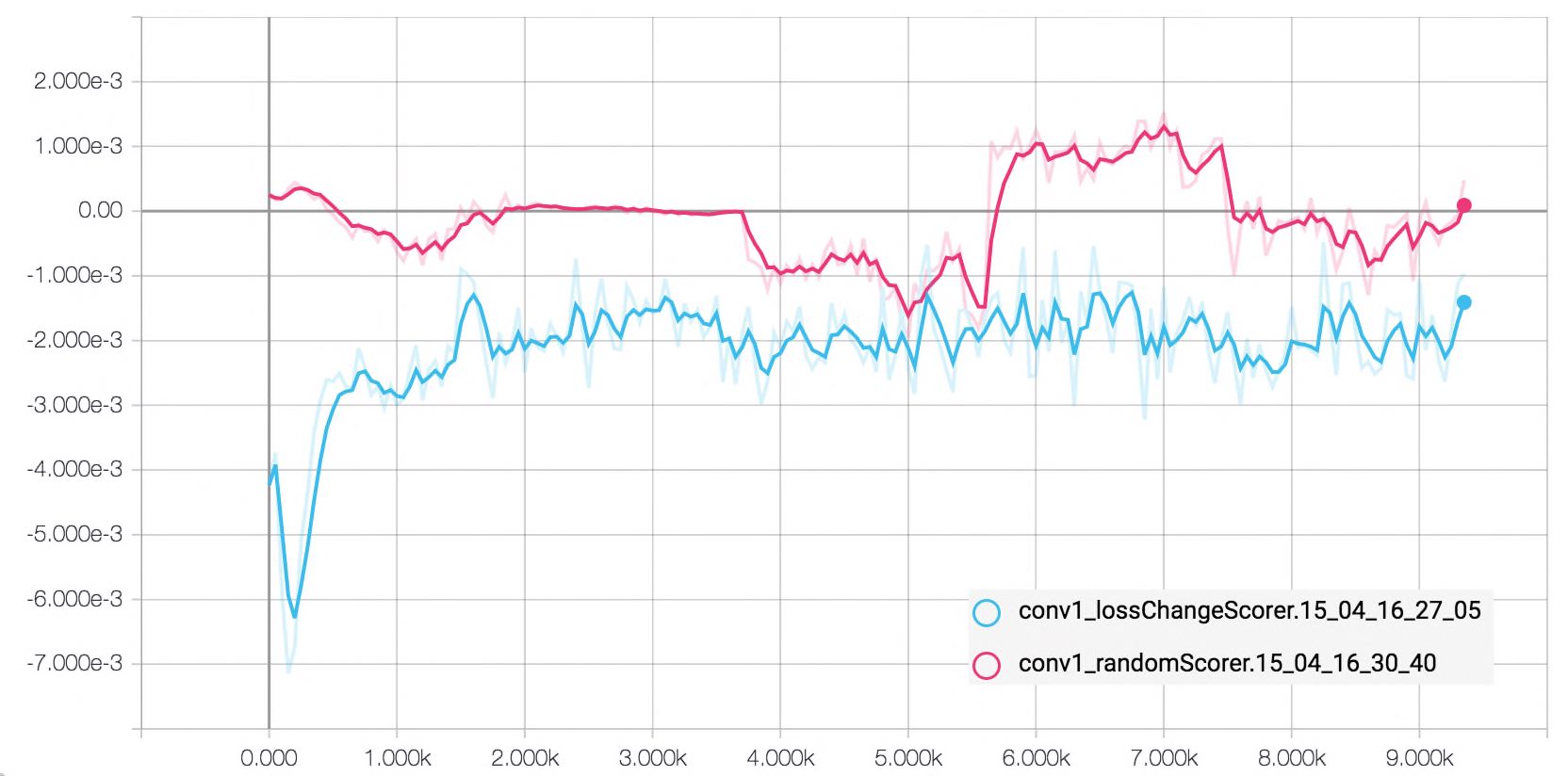}
    \caption{$f=0.01$}
    \label{fig:2.1.2a}
  \end{subfigure}
  \begin{subfigure}{.8\textwidth}
    \centering
    \includegraphics[width=.8\linewidth]{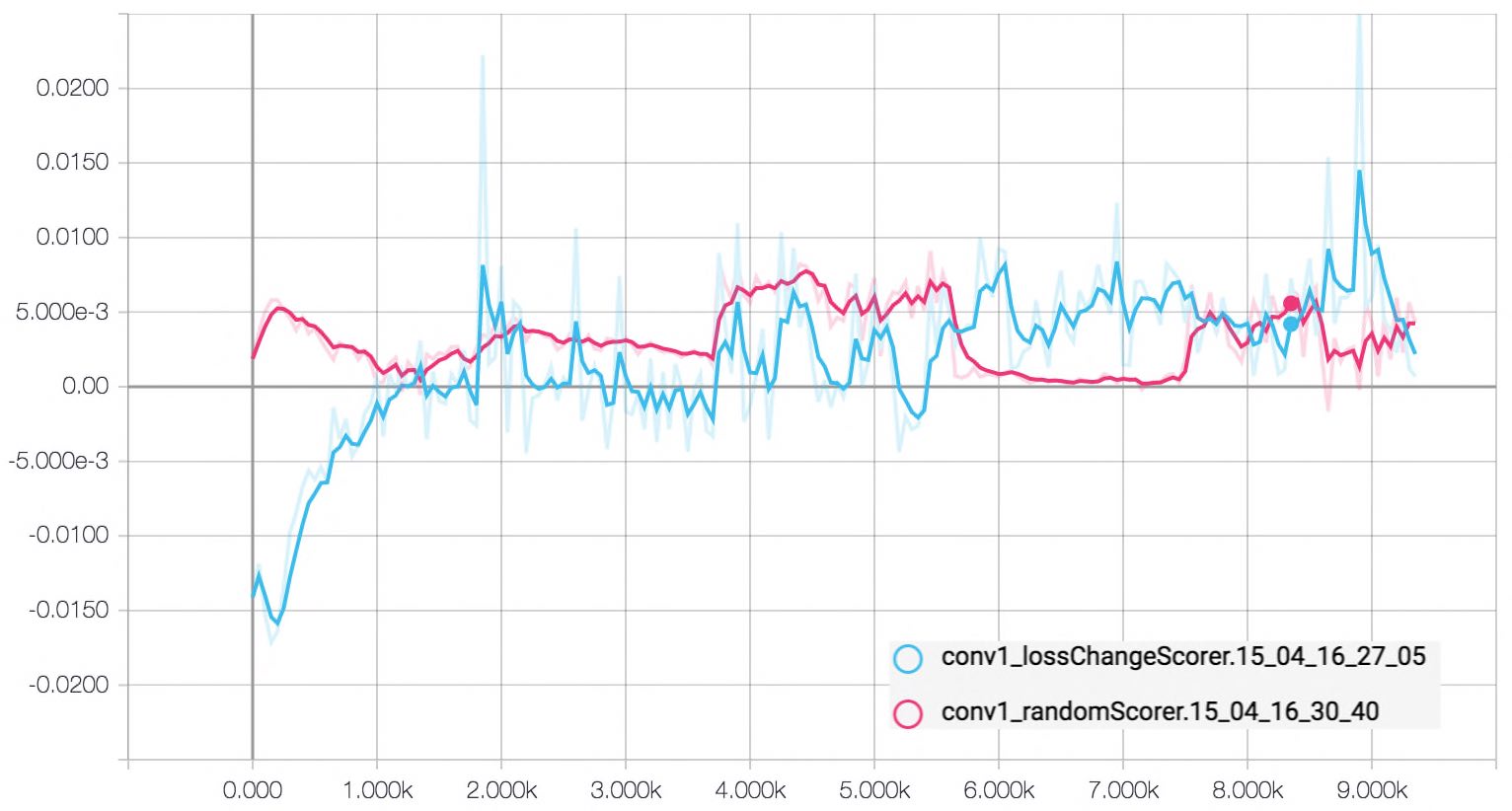}
    \caption{$f=0.05$}
    \label{fig:2.1.2b}
  \end{subfigure}
  \end{center}%

  \caption[Pruning can improve loss]{Small CNN [Appenix \ref{app3}] on MNIST with Tanh non-linearity. y-axis is the change in the loss if we prune fraction $f$ of the conv1 parameters(200 parameter total) using \textit{randomScorer} versus \textit{lossChangeScorer}. At (b) we can also see that the dependency increases over time and we start getting positive change in the loss even though we would get negative change if we prune a single parameter.}
  \label{fig:2.1.2}
\end{figure}
Since we get negative empirical saliencies for a good fraction of weights; why not expect a negative change in the loss after pruning? To demonstrate this idea we prune the first convolutional layer using the \textit{randomScorer} and \textit{lossChangeScorer} for two fractions at Figure \ref{fig:2.1.2}. When we prune only 2 parameters ($f=0.01$) we observe a constant improvement in the loss value throughout the training. Yes, we can improve our loss value if we prune some small number of weights together. However, when we increase the pruned parameters to 10, we start getting into the positive regime. In some cases, the loss value increases even more than the case where we randomly prune 10 parameters. This observation shows that one can improve the loss value, even after 5 epochs if the number of pruned parameters are small. Note that one can always calculate the saliency again after pruning let say 2 parameters. We haven't pursued this idea, however, it would be interesting to see how much one can improve loss function doing iterative pruning.

\subsection{Dependency between parameters increases over time \label{sec2.1.3}}
Looking into Figure \ref{fig:2.1.2}.b we see that pruning 10 parameters(out of 200) induces a negative change in the loss at the beginning of the training. However, after around 1000 steps, we observe that the dependency between scores increases and we start getting positive changes. This clearly shows that using individual saliency scores to prune a group of parameters is not a good strategy. We should somehow find a group of parameters that incur the smallest loss change when pruned together. This scoring function or approximation is yet to be discovered.
\subsection{Approximating loss change doesn't really work \label{sec2.1.4}}
\begin{figure}[ht]
  \begin{center}
  \begin{subfigure}{.75\textwidth}
    \centering
    \includegraphics[width=.8\linewidth]{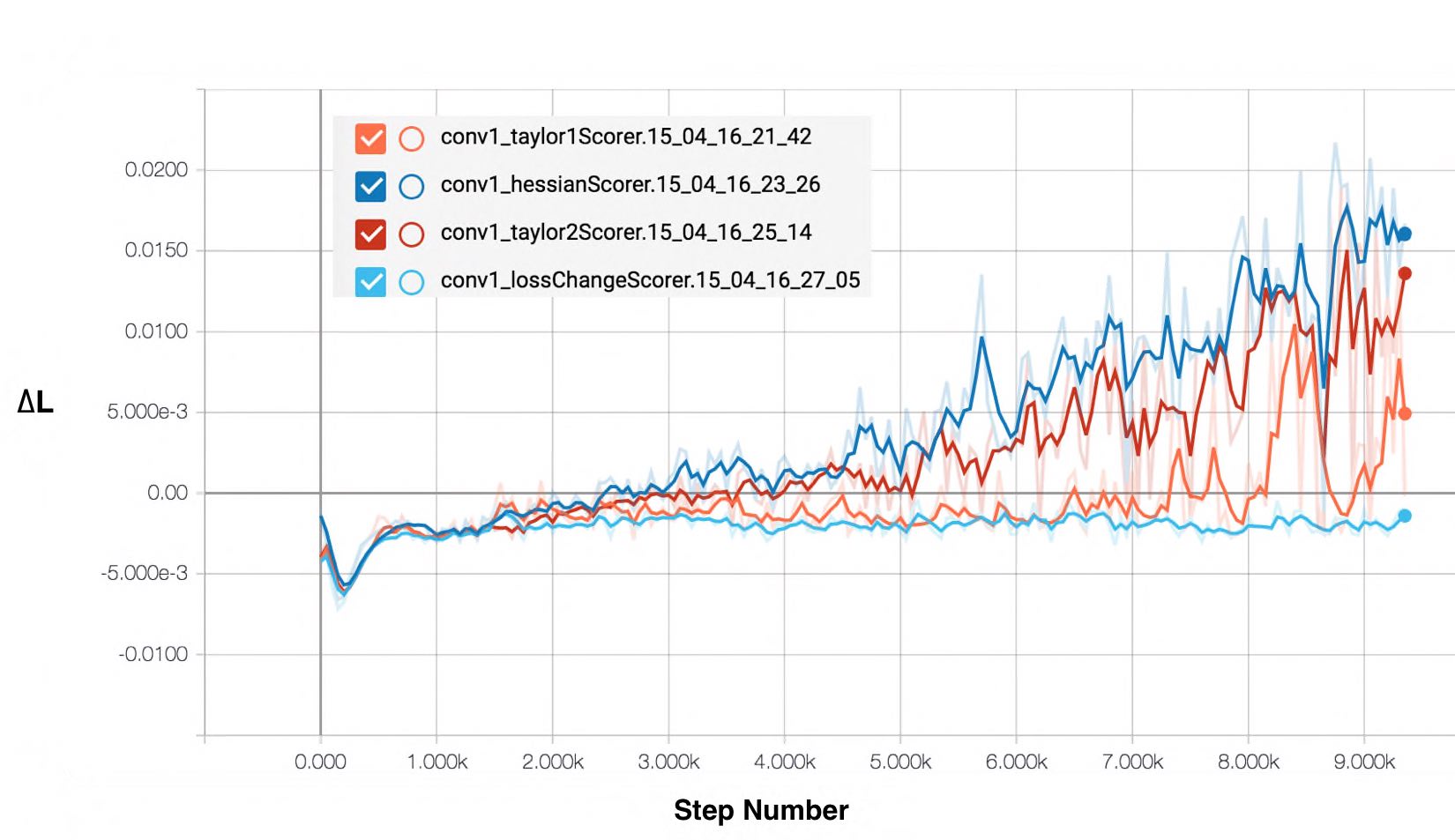}
    \caption{$f=0.01$}
    \label{fig:2.1.4.1a}
  \end{subfigure}
  \begin{subfigure}{.75\textwidth}
    \centering
    \includegraphics[width=.8\linewidth]{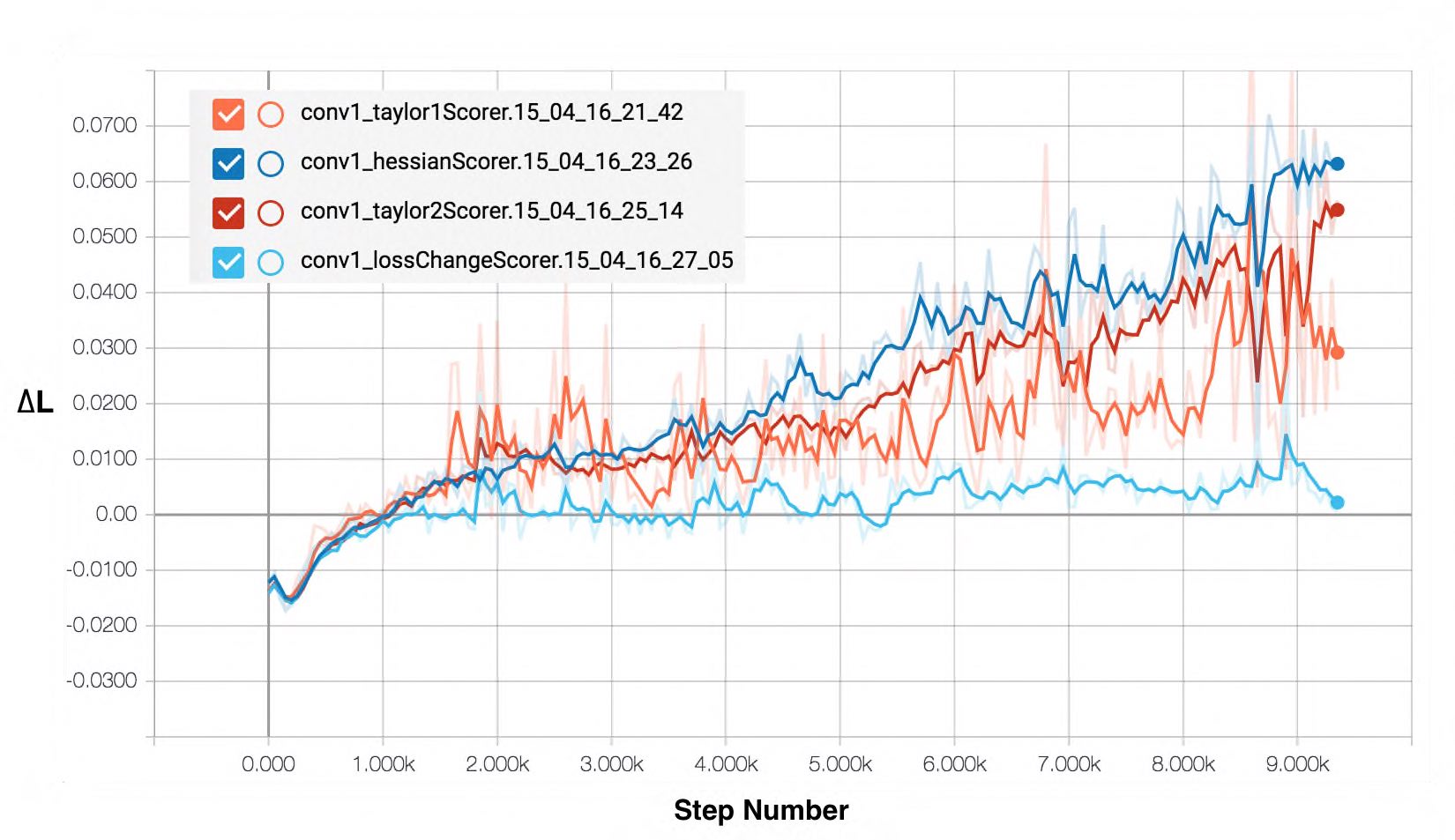}
    \caption{$f=0.05$}
    \label{fig:2.1.4.1b}
  \end{subfigure}
  \end{center}%
  \caption[Approximating loss change]{Smoothed loss change after pruning with various fractions using various scoring functions which approximates the \textit{lossChangeScorer}. The close relation at the early training disappears towards the end. The three approximation also posses some peaks, which challenges the stability of these functions.}
  \label{fig:2.1.4.1}
\end{figure}

Even though in the previous section we showed that \textit{lossChangeScorer} works poorly at pruning group of weights we still like to compare its performance with the performance of its approximations in the context of pruning. Note that one can numerically compare the approximations and measure the correlation between them. We leave this comparison as a future work. In Figure \ref{fig:2.1.4.1} we repeat the experiment in Figure \ref{fig:2.1.2} replacing the \textit{randomScorer} with the three approximations of the saliency score $S_i$ mentioned. We observe that these approximations work quite well early in the training, whereas towards the end of the training we see a relatively bad performance. We also observe that these approximations have occasional peaks which may significantly deteriorate the performance of pruning if used in practice. Remembering that all of the pruned parameters have negative scores for $f<0.35$, higher changes in the loss value is a sign for increased dependency between the scores of the parameters. The answer to the question why parameters selected by saliency approximations have higher internal dependency is yet to be discovered.

\begin{figure}[ht]
  \begin{subfigure}{.48\textwidth}
    \centering
    \includegraphics[width=.8\linewidth]{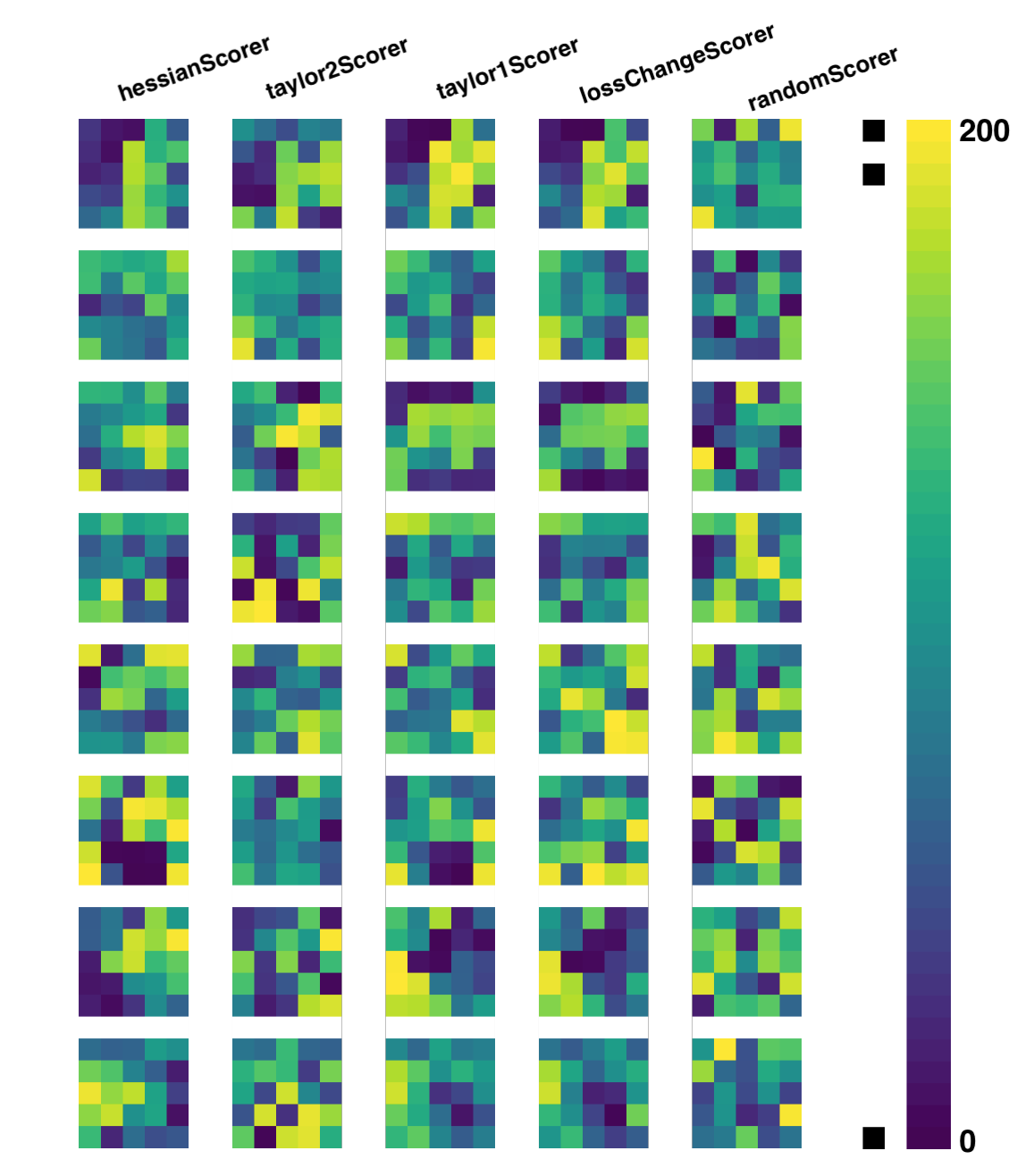}
    \caption{after step 50}
    \label{fig:2.1.4.2a}
  \end{subfigure}
  \begin{subfigure}{.48\textwidth}
    \centering
    \includegraphics[width=.8\linewidth]{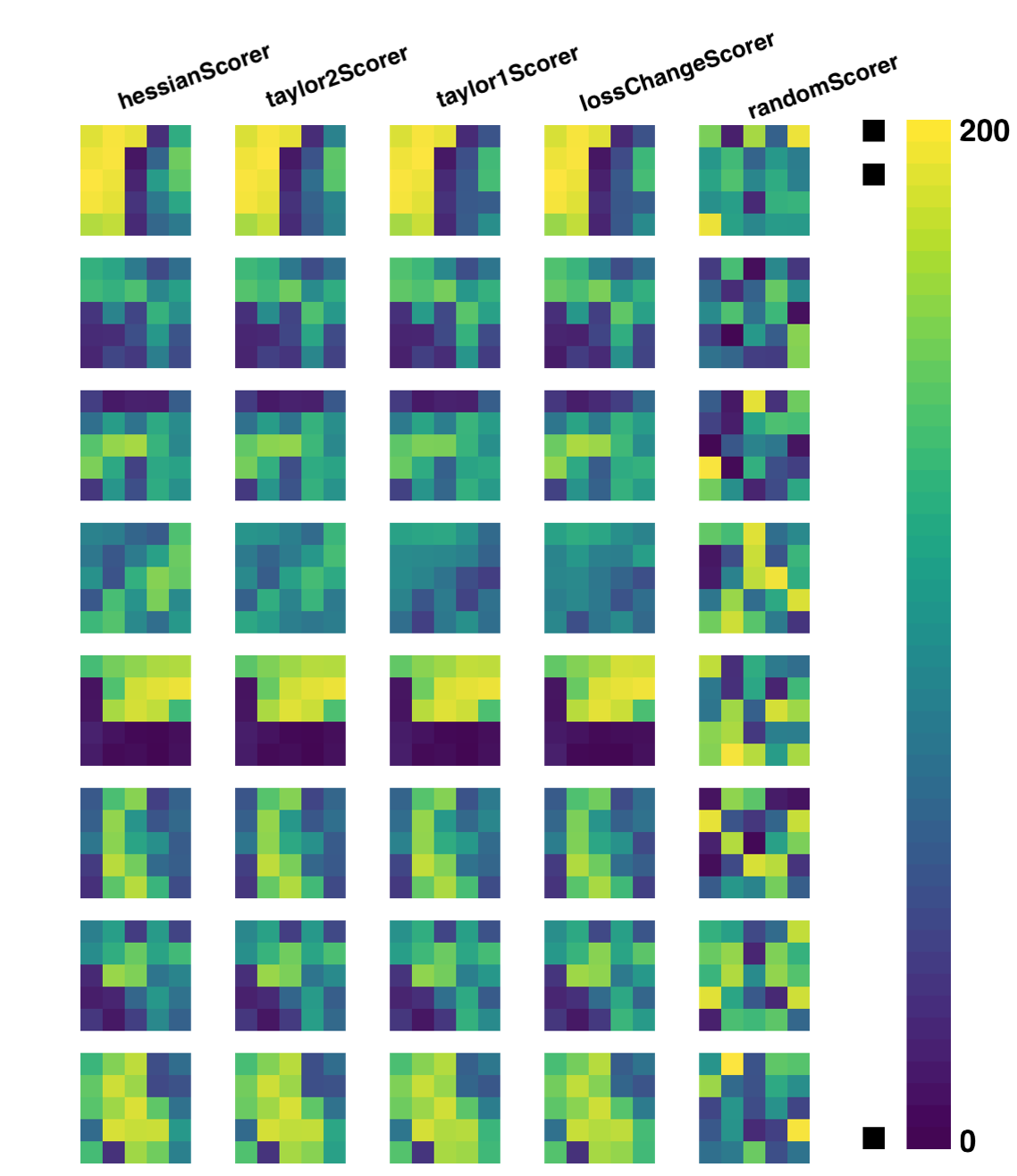}
    \caption{after step 9350}
    \label{fig:2.1.4.2b}
  \end{subfigure}
  \caption[Ranking of parameters among different scoring functions]{Ranking of parameters among different scoring functions. Note that all scoring functions are trying to approximate \textit{lossChangeScorer}. The first convolutional layer has 8 filters and yellow end of the color scale is used for high saliency ranking. We note that taylor1Scorer looks like best at approximating}
  \label{fig:2.1.4.2}
\end{figure}
One possible explanation is, even though pruning parameters with negative saliencies alone would decrease the loss, they may increase the saliencies of other parameters more than the others. It might be the case that pruning negative saliency parameters make other parameters more important for the network. In other words, their absence might affect the saliencies of other parameters somehow more than the others. We may try to find parameters with small effect. In other words, parameters with close to 0 loss-change might bring better performance. We argue this idea in Section \ref{sec2.1.5}.

We believe that some parameters are just happened to have a low saliency in the beginning and they don't change since they don't get a proper gradient signal.

In Figure \ref{fig:2.1.4.2} we plot the saliencies of the first convolutional layer for the five scoring functions used in this section. We observe that \textit{taylor1Scorer} looks like best at approximating the \textit{lossChangeScorer} based rankings. \textit{taylor2Scorer} is a combination of \textit{hessianScorer} and \textit{taylor1Scorer} and we are using $\lambda=1$ in the formula proposed in Table \ref{table:2.1}. One may get better approximations playing with the coefficient $\lambda$.

Note that in general, pruning methods use a fraction of the parameters to decide which connections to prune. Therefore the ranking of a parameter is usually what only matters. However one can also use a threshold score to decide how much to prune. This strategy is not used in practice according to our knowledge.

\subsection{Absolute valued scores works better\label{sec2.1.5}}
\begin{figure}[ht]
  \begin{center}
  \begin{subfigure}{.8\textwidth}
    \centering
    \includegraphics[width=.8\linewidth]{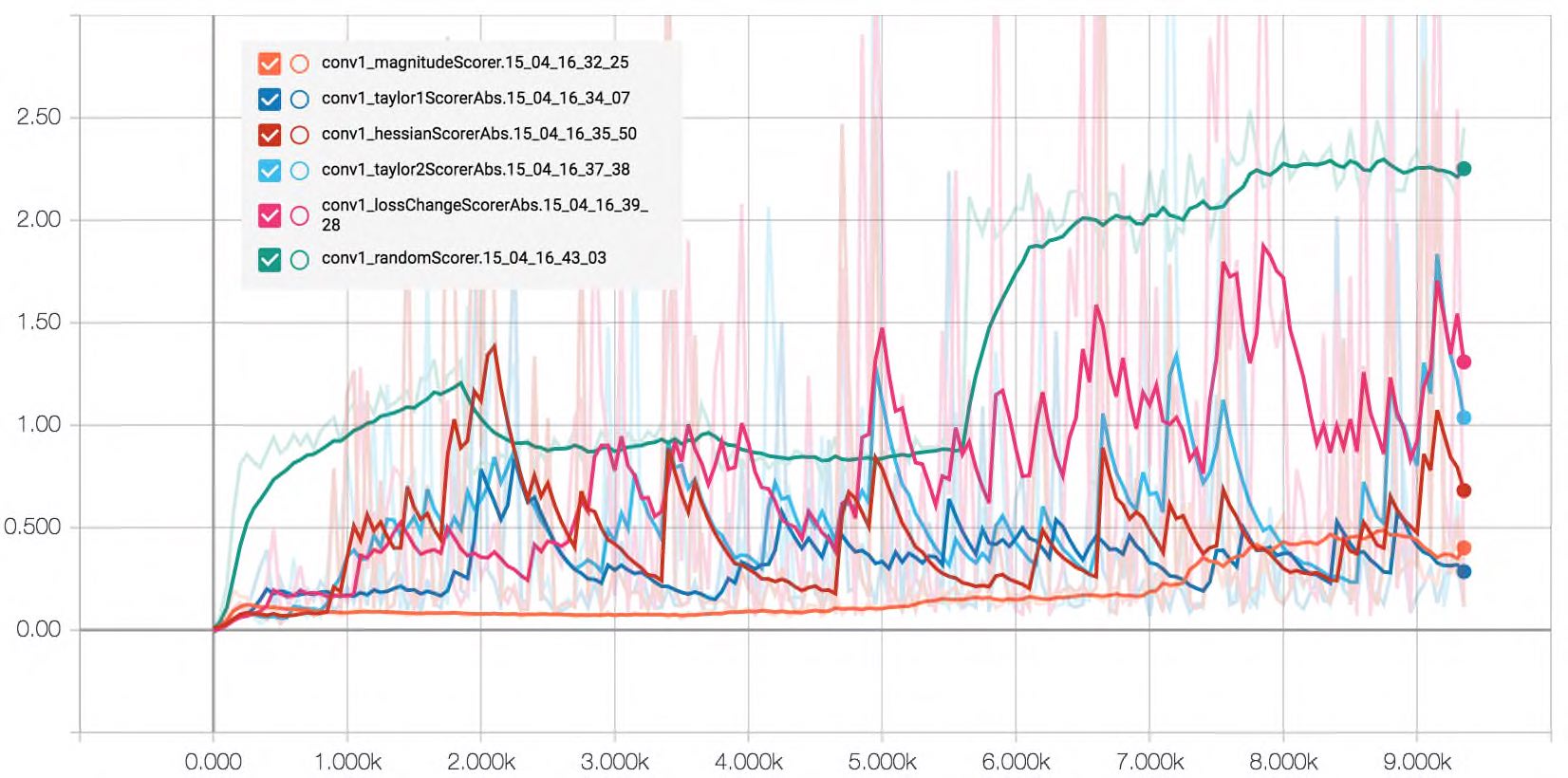}
    \caption{$relu$}
    \label{fig:2.1.5.1a}
  \end{subfigure}
  \begin{subfigure}{.8\textwidth}
    \centering
    \includegraphics[width=.8\linewidth]{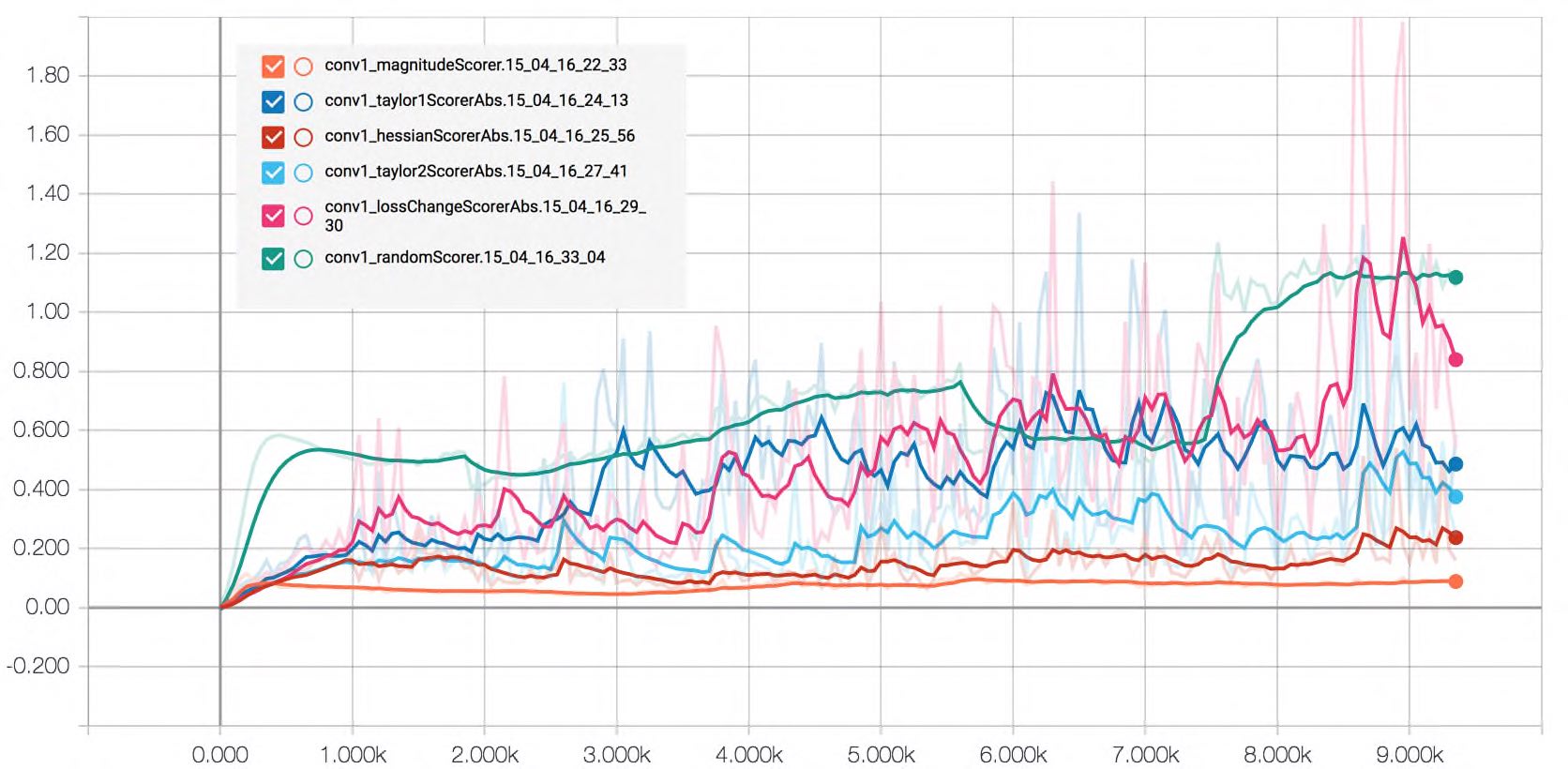}
    \caption{$tanh$}
    \label{fig:2.1.5.1b}
  \end{subfigure}
\end{center}
  \caption[Final comparison of Saliency Functions for Pruning]{Change in the loss value after pruning 160 of 200 parameters of the first convolutional layer over the training. The x-axis is the iteration number and y-axis represents $\Delta L(w)$. Curves are smoothed.}
  \label{fig:2.1.5.1}
\end{figure}

As observed in the previous section using saliency scores based on the loss-change when that individual parameter is pruned away, is a bad idea since we observe high dependency between selected parameters and get very bad pruning results. How can we find a scoring function which detects the low saliency and loosely dependent parameters? In our experiments, we found out that parameters with low absolute value scores have better results at pruning. We argue that this is due to the fact that parameters with a small effect on the loss(negative or positive) also have a weak dependency with other parameters.

Imagine a weight $w_{ij}^l$ that determines the contribution of $a_j^{l-1}$ at the summation happening in unit $i$, layer $l$. If the value generated at the unit $u_i$ is not used or in other words the outgoing weights $w_{i\cdot}$ from the $u_i$ are small we would have a very small change in the loss when we prune the $w_{ij}$ away, in other words $w_{ij}$ would have a small saliency. One can also expect to see parameter $w_{ij}$ having a similar low magnitude saliency score if we prune some other weights. In other words, $w_{ij}$ would have a very small effect on the saliency score of other weights and therefore one can prune $w_{ij}$ and similar low magnitude saliency parameters together expecting to see a smaller change in the loss value. This is observed in Figure \ref{fig:2.1.5.1} for a pruning fraction of 0.8. Absolute values of $S_i$ and its approximations work much better than the raw saliency scores in practice when used for pruning. Therefore we compare absolute valued versions of saliency functions in the next section with the magnitude based scoring.

\subsection{Magnitude based scoring works best\label{sec2.1.6}}
\begin{figure}[ht]
  \begin{center}
    \includegraphics[width=.9\textwidth]{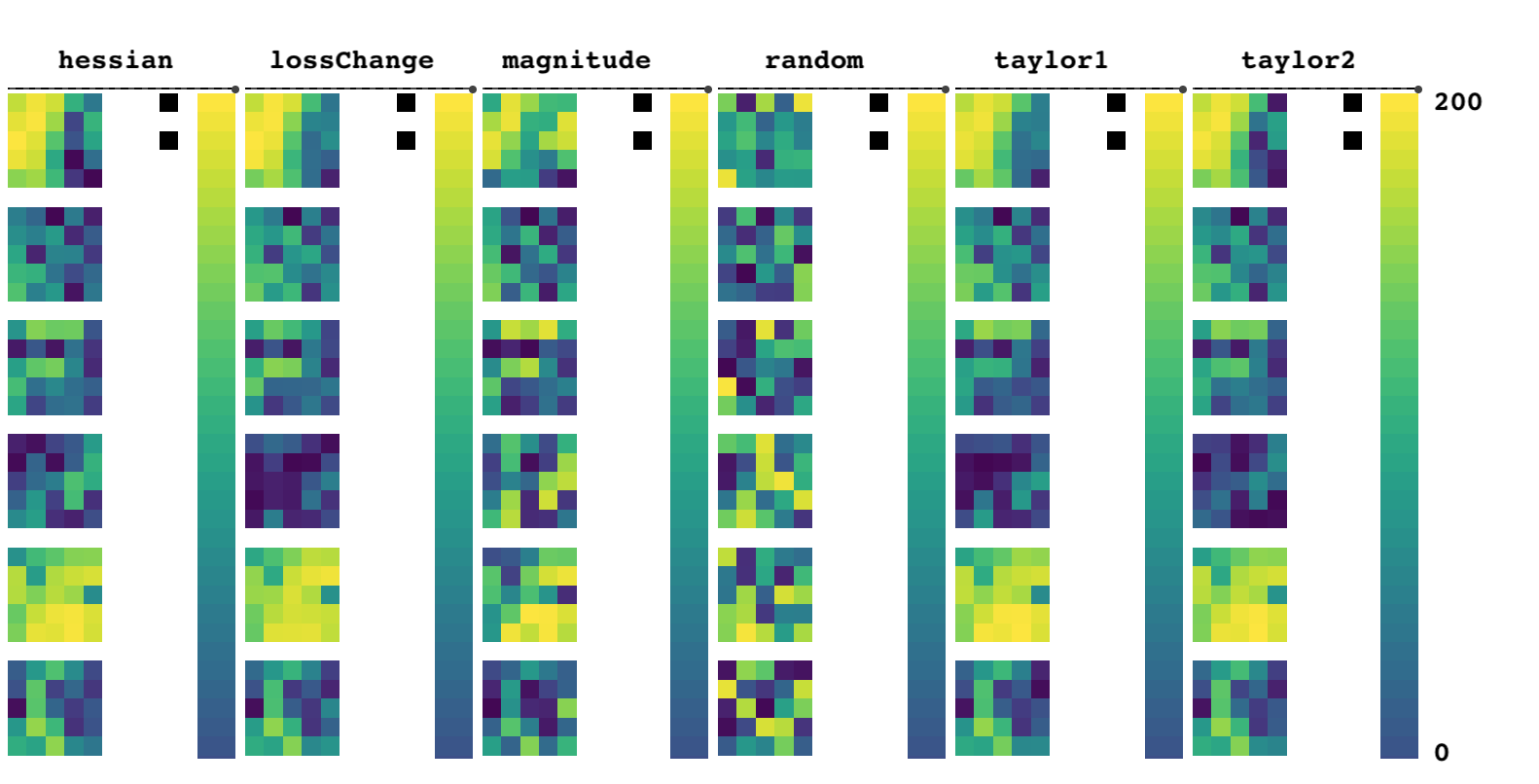}
  \end{center}%
\caption[Rankings of parameters using positive scoring functions]{Rankings of the parameters of the first convolutional layer determined with the absolute value based scoring functions. At the end of the training we calculate saliencies of filters of first convolutional layer using various saliency functions and then rank them. Yellow means high saliency and blue means low.}
\label{fig:2.1.5.2}
\end{figure}
We plotted rankings generated by the various scoring functions at the end of the training in Figure \ref{fig:2.1.5.2}. We clearly see the resemblance between the \textit{lossChangeScorerAbs} and its approximations \textit{hessianScorerAbs},\textit{taylor1ScorerAbs} and \textit{taylor2ScorerAbs}. They also have some similarities with the \textit{magnitudeScorer}.

In Figure \ref{fig:2.1.5.1} we see that \textit{magnitudeScorer} performs best among all other absolute value based scores. However one can see that other scores occasionally incur smaller loss change at the beginning and towards the end. The unexpected oscillations and peaks are problematic for most of the saliency scores and may be avoided using a running average or a bigger validation set. However \textit{magnitudeScorer} seems to be quite stable throughout the training and looks like the simplest and best scoring function we found.

\section{A Basic Pruning Experiment\label{sec:2.2}}
\begin{figure}[ht]
  \begin{center}
    \includegraphics[width=8cm]{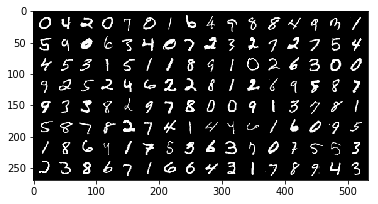}
  \end{center}%
\caption[sample images from MNIST]{some samples from the MNIST dataset\cite{mnist}}
\label{fig:2.2mnist}
\end{figure}

Pruning ANN's is not limited to after training. One can prune a network also during training as they can do that after the network is trained. To demonstrate how pruning works during training we set up an experiment using a small CNN trained on MNIST dataset (Figure \ref{fig:2.2mnist}). The CNN has two convolutional layers with 8,16 filters respectively followed by max-pooling(stride=2) and Relu nonlinearity. Following are the two fully connected layers with 64 and 10 units respectively. Details of this network and all other networks can be found in Appendix \ref{app3}.

In this example we are using the simple \textit{magnitudeScorer}(Table \ref{table:2.1}) to prune a fraction $f$ of the network parameters each time and we are doing it with the following the schedule in Table \ref{table:2.2}. So after each scheduled epoch is done, we are setting a fraction $f$ of all parameters with the smallest magnitude to zero. The gradient of these parameters is removed after pruning at every iteration to ensure that the weights stay zero and we are simulating weight removal correctly. After the run with pruning, we repeat the same experiment without pruning using the same initial random seed.

\begin{table}[ht]
\begin{center}
\begin{tabular}{|c|c|c|c|c|}
  \hline
  \textbf{Epoch} & 2 & 3 & 4 & 5  \\
 \hline
\textbf{Fraction $f$} & 0.2 & 0.5 &  0.8 & 0.9  \\
 \hline
\end{tabular}
\end{center}
\caption[Pruning schedule]{Pruning schedule for the MNIST experiment Section \ref{sec:2.1}}
\label{table:2.2}
\end{table}

\begin{figure}[ht]
  \begin{subfigure}{.8\textwidth}
    \centering
    \includegraphics[width=.8\linewidth]{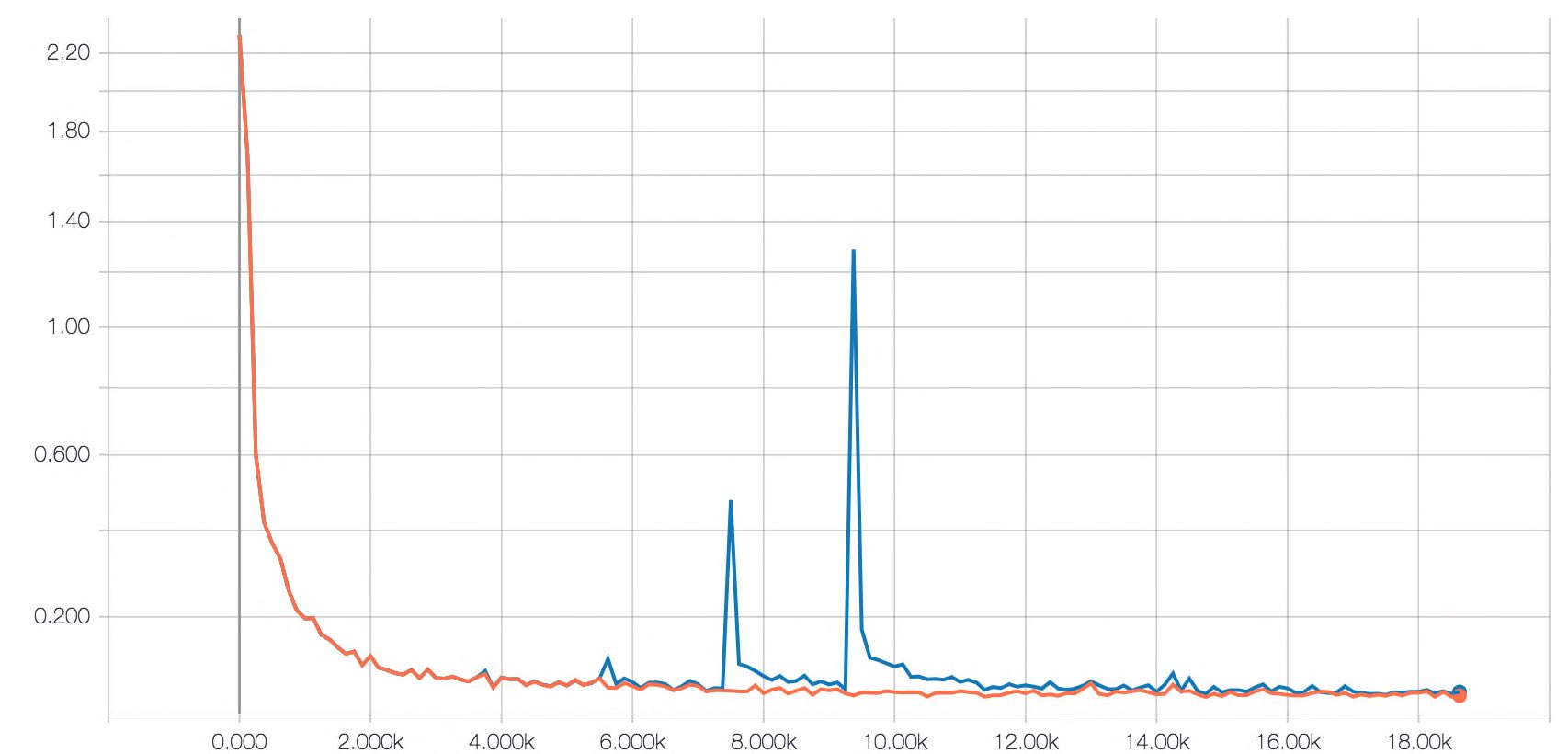}
    \caption{loss}
    \label{fig:2.2loss}
  \end{subfigure}
  \begin{subfigure}{.8\textwidth}
    \centering
    \includegraphics[width=.8\linewidth]{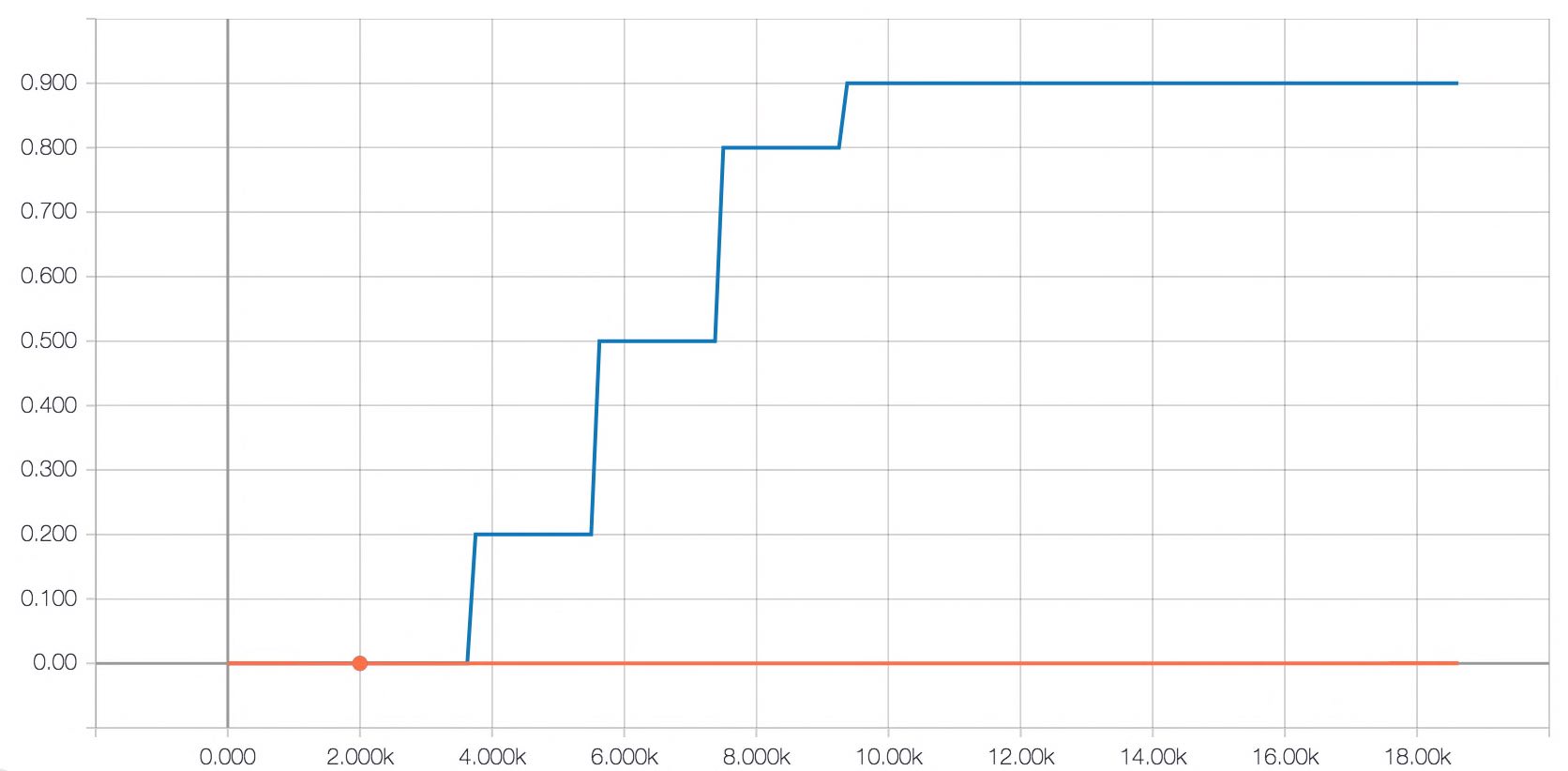}
    \caption{fraction of paramaters pruned}
    \label{fig:2.2fraction}
  \end{subfigure}
  \caption[Pruning without performence loss]{Loss and fraction of pruned parameters over time. X axis represents time and parameterized by the step number. Blue curve is the run with scheduled pruning, where the orange curve is the original run without any pruning done. We are using the same seed for random number generator to ensure consistency.}
  \label{fig:2.2lossfraction}
\end{figure}

So overall we perform 10 epochs of training and the loss value is plotted every 50 mini-batch/step(see Figure \ref{fig:2.2loss}). One can observe fractions smaller then 0.5 doesn't introduce much of a penalty in the loss. For fractions 0.8 and 0.9 we can see the jumps in the loss which is recovered during the rest of the training.

\begin{figure}[ht]
  \begin{subfigure}{.23\textwidth}
    \centering
    \includegraphics[width=.9\linewidth]{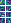}
    \caption{conv1}
    \label{fig:2.2conv1}
  \end{subfigure}%
  \begin{subfigure}{.23\textwidth}
    \centering
    \includegraphics[width=.9\linewidth]{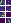}
    \caption{conv1-pruned}
    \label{fig:2.2conv1p}
  \end{subfigure}
  \begin{subfigure}{.23\textwidth}
    \centering
    \includegraphics[width=.9\linewidth]{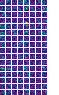}
    \caption{conv2}
    \label{fig:2.2conv2}
  \end{subfigure}%
  \begin{subfigure}{.23\textwidth}
    \centering
    \includegraphics[width=.9\linewidth]{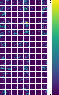}
    \caption{conv2-pruned}
    \label{fig:2.2conv2p}
  \end{subfigure}
  \caption[Pruning convolutional layers]{Convolutional filter weight magnitudes with and without pruning. Yellow is for the maximum magnitude value encountered during training and blue is for the minimum value. \textit{conv1} and \textit{conv2} are the first and second convolutional layers. Each square for \textit{conv1} is a unit. Each row for \textit{conv2} is a unit and each unit has 8 kernels for the incoming 8 channels}
  \label{fig:2.2conv}
\end{figure}

After the training is done weights of the trained network are colored with a colormap; where yellow and dark blue used for high and low values. Before applying the colormap we normalize all the weights within a layer to the range [0,1]. Colormaps of the two convolutional layers are plotted in Figure \ref{fig:2.2conv}.

We can see that the high magnitude weights in the original run are mostly preserved in the pruned network. This indicates that the regions contributing to the result most are formed within the first couple of epochs. We can also see in Figure \ref{fig:2.2conv1p} that 2 out of 8 filters are completely pruned; so that the corresponding filters in the \textit{conv2}(column 6 and 7) are also mostly 0. Similarly, all filters in two units(rows 7 and 9) of \textit{conv2} layer seems to be dead. One can also observe that usually 3-4 out of the 8 input channels for \textit{conv2} units are almost all zero. This shows us the sparse relation between the units of the two layer.

Fully connected layer \textit{fc1} takes the flattened output of the \textit{conv2} layer. Similar to the colormaps of the convolutional layers, we can spot the high magnitude weights mostly appear in the pruned network, even though we start pruning after the second epoch. We can also detect two vertical stripes at \textit{fc1} layer corresponding to the dead units(7,9) in the \textit{conv2} layer. If investigated carefully one can also relate the horizontal almost all zero(dark-blue) rows in \textit{fc1} layer to the vertical rows in \textit{fc2}.
\begin{figure}[ht]
  \begin{center}
  \begin{subfigure}{.8\textwidth}
    \centering
    \includegraphics[width=.8\linewidth]{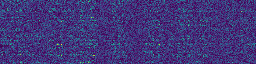}
    \caption{fc1}
    \label{fig:2.2fc1}
  \end{subfigure}
  \begin{subfigure}{.8\textwidth}
    \centering
    \includegraphics[width=.8\linewidth]{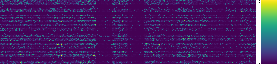}
    \caption{fc1-pruned}
    \label{fig:2.2fc1p}
  \end{subfigure}
  \vspace{0.5cm}
  \begin{subfigure}{.8\textwidth}
    \centering
    \includegraphics[width=.8\linewidth]{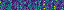}
    \caption{fc2}
    \label{fig:2.2fc2}
  \end{subfigure}
  \begin{subfigure}{.8\textwidth}
    \centering
    \includegraphics[width=.8\linewidth]{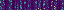}
    \caption{fc2-pruned}
    \label{fig:2.2fc2p}
  \end{subfigure}
\end{center}
  \caption[Pruning fully connected layers]{Fully connected weight magnitudes with and without pruning. Yellow is for the maximum magnitude value encountered during training and blue is for the minimum value. \textit{fc1} and \textit{fc2} are the first and second fully connected layers. Each row of the images represent individual units.}
  \label{fig:2.2fc}
\end{figure}

At this point, one can expect us to convert the network to the sparse format and enjoy the storage and computational gains. However, at the time of writing this thesis, acceleration libraries written for dense operations are much faster than their sparse counterparts and therefore converting a network with 90\% sparsity to a sparse format would not improve the runtime if not harming it.

\begin{figure}[ht]
  \begin{center}
    \includegraphics[width=10cm]{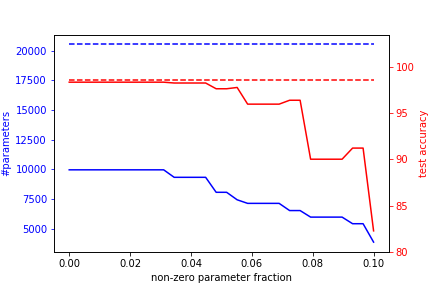}
  \end{center}%
\caption[Unit removal penalties after pruning]{Left axis and blue curves represent the number of parameters after removing units for the given $f_{nz}$ on the x-axis. The dashed blue line is the parameters of the original dense network. Right axis with red curves are the test accuracies of the networks after unit removal.}
\label{fig:2.2removal}
\end{figure}

As the last part of this experiment, we count non-zero incoming and out-going weights for each unit. Then we remove the units with less than a fraction $f_{nz}$ of non-zero incoming or outgoing weights (Figure \ref{fig:2.2removal}). Fraction 0 would correspond to removing units with all zero incoming or outgoing weights. Note that a unit with all zero outgoing weights has no effect on the result and we can safely remove it. Similarly, a unit with all zero incoming weights would always generate the bias value and one can remove the unit after propagating the bias to the next layer. In this experiment, we haven't implemented the bias propagation and therefore we get slightly smaller test accuracy with $f_{nz}=0$ and this can be easily avoided. We observe that we can make the dense network half size without any performance penalty. This shows that parameters pruned using the magnitude score have some kind of a structure; i.e. they are not completely random.

\chapter{Generating Dead Units Through High Learning Rate\label{chap3}}
Pruning individual parameters of an ANN works quite well if we look at the fraction of parameters that is pruned. However, in practice, 90\% pruning does not correspond to 10x speedup or 10x compression. This is due to the inefficiency of sparse data structures and operations. However, when we prune the whole unit, we can remove the whole row or filter from the parameter tensor and reduce the size of the network without switching to sparse representations.

In this section, we switch our attention to individual units. Units or in other words neurons are the summing nodes in each layer, where the combination of previous activations happens. In other words, units are the meeting points where the information is generated and distributed. The forward information and the gradient flow through individual units and a flaw might affect the whole group of parameters connecting the unit. Are there units that are useless, preventing the network from learning new information? Is it possible that gradient-based learning algorithms create these dead units? Are there any tricks we use training deep networks help individual units to stay alive? In Section \ref{sec3.1} we introduce \textit{mean replacement penalty} and \textit{mean replacement score} to quantize and approximate the importance of a unit in terms of their effect on the loss if removed. In Section \ref{sec3.2} we introduce the \textit{dead/frozen unit} concept and tie the definition of a dead unit with \textit{mean replacement penalty}. Then we argue, that a high learning rate might one of the reasons causing dead units during optimization and define an attack in Section \ref{sec3.3} to simulate high learning rates on individual units. Finally in Sections \ref{sec3.4},\ref{sec3.5},\ref{sec3.6} and \ref{sec3.7}, we discuss results of our experiments.

\section{Mean Replacement Penalty and its First Order Approximation \label{sec3.1}}
We defined the saliency for a single parameter in Chapter \ref{chap1} as the change in the loss value, when the parameter is removed. Now in this section, we would like to do the same for individual units. \cite{molchanov2016} defines the saliency of a unit as the change in the loss when the entire unit is removed. This is equivalent of replacing the output of a unit with zeros. However, this approach fails for units that generate constant values.

So how should we define the saliency for units? Following the pruning idea, we can try to find units which produce zeros most of the time. If the output of a unit sums to zero for all samples, then we can safely remove it along with all of its outgoing weights. What if the unit produces a constant value instead of a zero? This is still okay, since we can propagate this constant value to the next layer(bias propagation Section \ref{sec:1.2}) and remove the unit without introducing any penalty. Even if, the unit does not generate a constant value, we can still remove the unit and propagate its mean to the next layer and expect to see a smaller change in the loss by doing so compared to the naive zero replacement. We call this removal process as \textbf{Mean Replacement} and define the \textbf{Mean Replacement Penalty (MRP)} as the following:

\theoremstyle{definition}
\begin{definition}{\textbf{MRP: Mean Replacement Penalty}}
  is the change in our loss value due to replacing the value $[h_k]_i$ generated at unit $i$ for the sample $k$ with its mean value $E_{k \in D_{val}}[[h_k]_i]$ calculated over a validation set of $D_{val}$.
  \begin{equation} \label{eq3.1}
\text{MRP}_{k,i} = \sum_{k \in D_{val}} |L(\tilde h_{k,i}) - L(h_k)|
\end{equation}
  where
  \[ [\tilde h_{k,i}]_j=\begin{cases}
        E_{k \in D_{val}}[[h_k]_i] & j = i \\
        [h_k]_j & j\neq i
     \end{cases}
  \]
\end{definition}

Note that the definition of MRP involves the absolute value operator, which is different than the naive sum. This detail is important since it makes our penalty even more strict and captures the sensitivity against the mean replacement correctly.

Calculating MRP empirically would require us to run a forward-pass for each unit. Even though number of units are less then number of parameters, it is not feasible for deep networks (i.e. AlexNet has almost 300,000 units) to calculate the MRP for every single unit. Therefore we propose using the first-order Taylor approximation of the MRP. Like we did in Equation \ref{eq1.4}, we can write down the Taylor approximation of the loss value using the output of a given layer. Let's denote the output of some layer for the sample $k$ with $h_k$ vector.

\begin{equation} \label{eq3.2}
L(\tilde h_k) = L(h_k)+\Delta h_k^T \nabla L(h_k)+O(\Delta h_k^2)
\end{equation}

By setting $\Delta h_k=\tilde h_{k,i}-h_k$  we get the first order approximation for replacing the output at unit $i$ with its mean. $\Delta h_k$ would have zeros in all indices expect the $i_{th}$ index. We can vectorize this calculation and calculate MRS for every unit of a given layer by setting $\Delta h_k=\tilde h_k-h_k$ where $\tilde h_k$ is the output vector whose elements are replaced by the corresponding mean values. Lets formally define this.

\theoremstyle{definition}
\begin{definition}{\textbf{MRS: Mean Replacement Saliency}}
  for the units in a given layer calculated by
  \begin{equation} \label{eq3.3}
  MRS_{k} = |L(\tilde h_k) - L(h_k)|= |(\tilde h_k-h_k)^T \nabla L(h_k)|
  \end{equation}
  where
$$[\tilde h_k]_j= E_{k \in D_{val}}[[h_k]_j]$$
\end{definition}

Our implementation of MRS calculates the mean values of all units and saves the zero-centered output vector(or matrix if it is a minibatch) during the forward-pass. Then, during the back-propagation, we multiply the output gradient of the layer with these centered output values to obtain the saliency vector. Note that since we are taking the absolute values at the end we don't need to negate the centered output values.

We would like to put a side note here. We calculate the mean of the output before the nonlinearity. However, we can also calculate the mean value after the nonlinearity and this would probably introduce a smaller loss penalty. We replace the values with the mean before the nonlinearity, since it was straightforward to implement and as the variance of the output goes to zero the two different approaches become equal to each other. Two different approaches become identical when MRP goes to zero. We also like to point out the possibility of using other operators than the mean, like median, min, max; etc. For example, using the mean output value for a unit that uses \textit{TanH} might not be the best idea.

\section{\textit{Frozen} and \textit{Dead} Units \label{sec3.2}}
In this section we are going to introduce two terms, hoping that they would help us to analyze and understand what is happening in neural networks over the training.

Let's imagine a unit in a fully connected layer. This unit generates a number doing a summation. It scales each of the previous activations, sums them and passes the result to a non-linearity like rectified unit(ReLU) or sigmoid function. The value after the nonlinearity is then named as the activation of the unit. These activations are then used by the units of the next layer in a similar manner.

During the back propagation, the unit gets error signals from the units of the next layer. The gradient of the unit is calculated by scaling and summing these signals as follows where $h^l_i$ denotes the output of $i^{th}$ unit in $l^{th}$ layer.
\begin{equation} \label{eq3.4}
\frac{\partial L}{\partial h^l_i} = \sum_{j}\frac{\partial L}{\partial h^{l +1}_j}w_{ij}g'(h^l_i)
\end{equation}

We define frozen units as following.
\theoremstyle{definition}
\begin{definition}{\textbf{Frozen Unit}}
\\Unit that receives negligible output gradient $\frac{\partial L}{\partial h^l_i}$ for all different samples of the distribution.
$$\frac{\partial L}{\partial h^l_i} \approx 0 \quad \forall    x \in D_{val}$$

\end{definition}

A unit might be \textit{frozen} in various situations. Overall loss can be zero and all network does not get any gradient signal. This would correspond to the first term in Equation \ref{eq3.4} being 0 $\forall$ $j$. However, this case is not really interesting since a zero loss would mean we are at the minima and there is not really much we can do. The interesting case is when the unit does not get any gradient signal even when there is some gradient signal available in next layers. Thinking the chain rule this might happen due to weak connections (small weights) that the unit has with the next layer (second term in Eq. \ref{eq3.4}). And another possibility is when the third term is zero, which would correspond to the case when the entire $h^l$ lies in the flat region of the activation function. This would make the third term $g'(h^l_i)$ zero.

It is also possible that the scaled gradients coming from the next layer cancel each other and the unit gets zero gradient. However, this is also unlikely to happen thinking there are many mini-batches and different combination of samples.

Note that all the layers except the first layer, get their input from other layers. Since we are constantly updating parameters during the training, input distributions of layers may change over time. However, once the input distributions stabilize we know that the frozen units are going to stay frozen. After early training, we expect the input distribution of a frozen unit not to change much. Therefore we also expect a frozen unit to stay frozen and we observed such a behavior in our experiments.\\
\textbf{When a unit is frozen, we expect it to stay frozen.}\\
Lets define the concept of dead units. Dead units are the units that doesn't effect the loss value when replaced by a constant. We can employ various strategies at picking the constant and in our definition we employ the definition at Section \ref{sec3.1}
\theoremstyle{definition}.
\begin{definition}{\textbf{Dead Unit}}
\\Units that are replaceable with a constant.
$$MRP_{k,i} \approx 0 \quad \forall    k \in D_{val}$$
\end{definition}

Let's identify the three situations where a unit introduces a low MRP.
\begin{enumerate}
  \item Unit generates a constant value
  \item All values that the unit generates mapped into a constant value by the activation function.
  \item All outgoing weights are zero.
\end{enumerate}

The first situation might happen if we have all near-zero weights or input-weight combinations that cancel each other in the summation. This situation is not very stable. The unit might get gradient signal and get out of this situation. Therefore we don't expect to see this state often during training.

The second situation is particularly interesting to consider since values at the flat region of activation functions would also mask the incoming gradient and therefore the unit would always have zero or near-zero $\frac{\partial L}{\partial h_i}$ for every sample. If a unit gets into this state, it also gets \textit{frozen} since all the gradient is masked by the activation function (the third term in Equation \ref{eq3.4} being 0). So we should definitely detect these units and either remove or revive since we know they are frozen and they would most likely stay frozen.

The third situation might happen when the network ignores the values unit generates. Assuming the unit generates non-constant values, this situation is stable only when the units in the next layer are frozen. Otherwise, the weights would get some gradient and become non-zero.

Units that don't generate constant values might end up having very low MRP, if the values created has a low variance or/and the outgoing weights are nearly zero.

In the remaining of this text we will use these definitions to explain some concepts and observations efficiently. We omit setting hard thresholds for these two definitions since thresholds depend on the number of units in a layer or the loss function used.

\section{How can a unit die?: High Learning Rate\label{sec3.3}}
There could be various reasons why we may end up having dead units. In this chapter, we would like to investigate one possible reason. We suspect that having a big step size during the training can put the output distribution of the unit into the flat region of the subsequent non-linearity. Most of the practical/popular optimization algorithms use the same learning rate for all the parameters in the network. During the training, this single learning rate may end up being too much for some units and cause them to die. We would like to artificially create this situation by attacking some of our units.

\textbf{An attack} at iteration $i$ on unit $j$ is done by multiplying the default learning rate for the weight parameters $W^l_i$ of unit $i$ without changing the learning rate for the rest of the network and the bias of the unit.

Each update on network parameters using the gradient calculated from the given mini-batch is called an \textit{iteration}. So we would pick a starting iteration number $S$ and perform our first \textit{attack}. Then we would repeat this attack $N$ times with $I$ many iterations in between. So we basically don't modify the learning rate for $I-1$ many iterations after each attack.

In the upcoming sections, we are going to investigate various non-linearities and how they respond to this learning rate attack. For all experiments below we train a small CNN (Appendix \ref{app3}) with CIFAR10 \cite{cifar10} dataset. We use SGD without momentum and a constant learning rate of 0.01. We calculate all data-dependent measures (like the loss and MRS) on a separate validation set of size 1000. We train the network in each experiment for 10 epochs and 30 epochs for experiments using \textit{TanH}. We use PyTorch's default parameter initialization and initialize the network with a uniform distribution from a range ($-\sigma_{l}^2,\sigma_{l}^2$) where the $\sigma_l$ is the variance calculated for each layer separately and equals to $\sigma_{l}^2= \frac{1}{n_{unit}}$. $n_{unit}$ is equal to the number of incoming connections for each unit.

\section{Killing a Unit with RELU+Batch Norm\label{sec3.4}}
Let's start this section by looking what happens during the training of a small CNN with RELU non-linearities and batch normalization. All experiments in this section run for 10 epochs and the networks reach to an accuracy of 67\%.

\begin{figure}
  \begin{adjustbox}{addcode={\begin{minipage}{\width}}{\caption[Attacking network with RELU+Batch Normalization]{What y-axis represents in each figure is written on top of each column. Each row represents different experiments with different units attacked. For vector quantities (row 3,4,5,6), it is the l1 norm of the vector that we are plotting. The x-axis represents the iteration number in every plot. We attack each unit separately starting from 50th step 10 times with 10 step intervals using a learning rate multiplier of 100. For each unit, we plot the \textit{original} behavior as indicated in each legend.}\label{fig:3.1}\end{minipage}},rotate=90,center}
    \includegraphics[scale=.15]{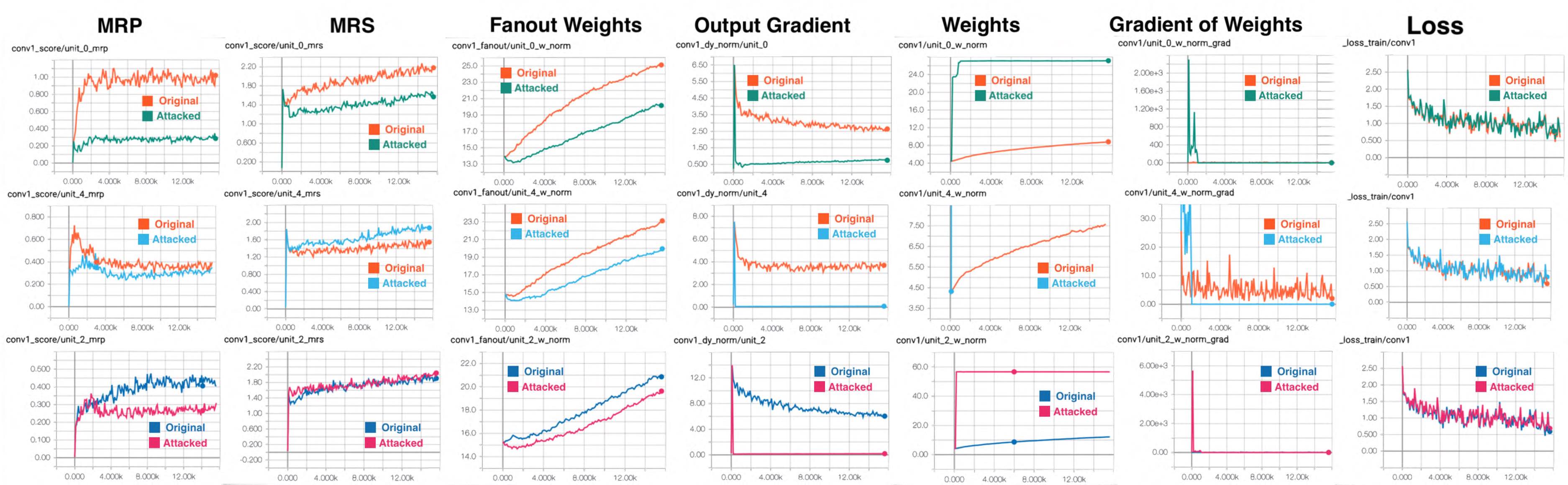}
  \end{adjustbox}
\end{figure}

In Figure \ref{fig:3.1} we see various plots for the attacks on 3 different units of the first convolutional layer. Each row represents a different unit. The two orange and the blue curves in the 3 rows respectively, represent the original training without any learning rate change. In common, we see that the norms of the weights increase. We also see that the MRP generally increases over time. It looks like after around 2000th step, units reach their final MRP values and they oscillate around these values afterwards.

Three units in Figure \ref{fig:3.1} have different responses to the attack that starts on 50th step and repeats 10 times with 10 step intervals. During each attack, we multiply the gradient for the attacked unit with 100. This cause unit to follow a different optimization path. The first thing we notice is various units responded to the attack differently. Let's start with common responses we see. As expected we see a jump in the norm of the weights(column 5). MRP of Unit-0 (first row in Figure \ref{fig:3.1}) reduces significantly without affecting training loss (last column). We observe that none of the attacks change the loss curves. This is very interesting, since if these high-learning-rate-updates happen during the training causing units to die, they pass undetected.

\begin{figure}
  \begin{center}
  \begin{subfigure}{.8\textwidth}
    \centering
    \includegraphics[width=.95\linewidth]{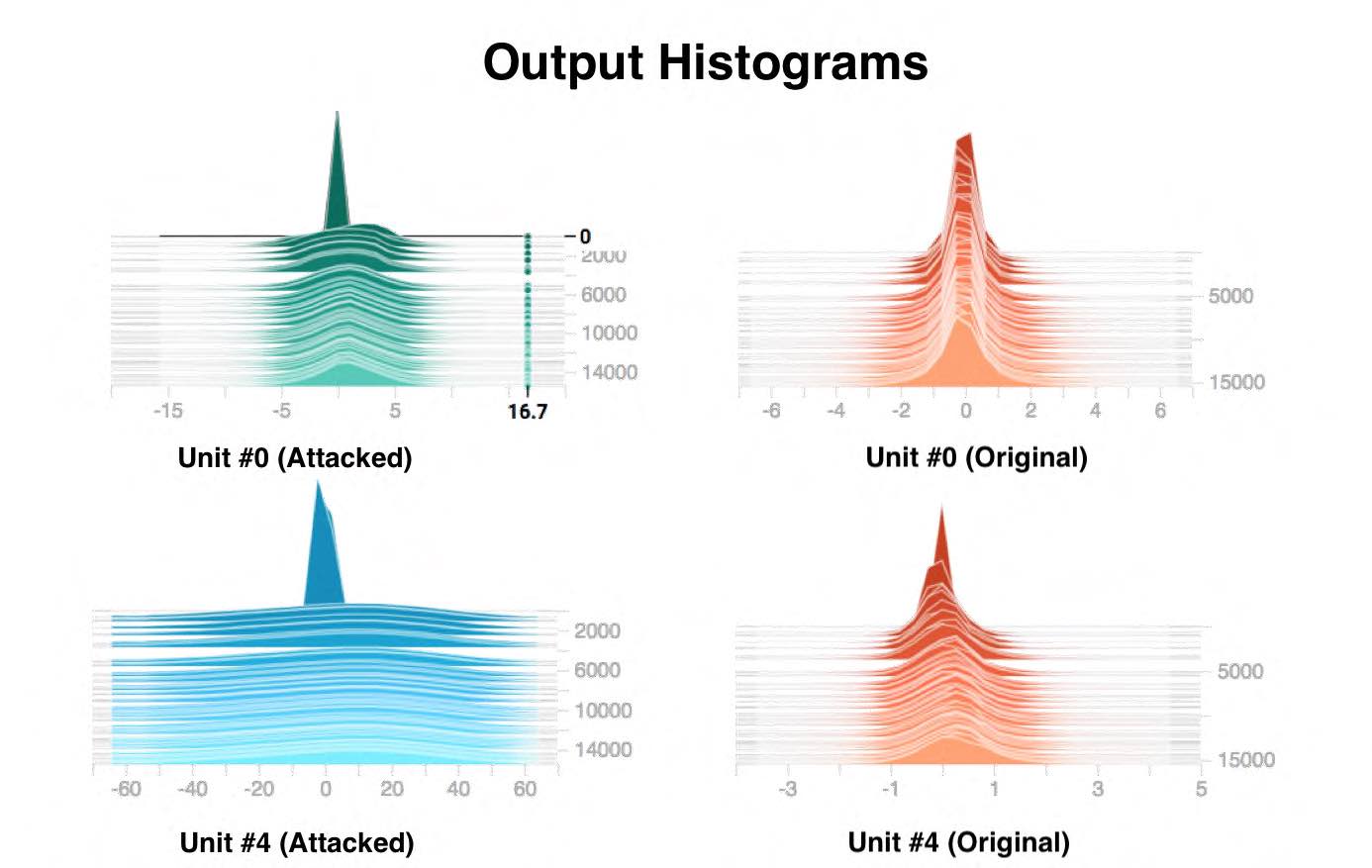}
    \caption{Histogram of outputs}
    \label{fig:3.2.1}
  \end{subfigure}
  \begin{subfigure}{.8 \textwidth}
    \centering
    \includegraphics[width=.95\linewidth]{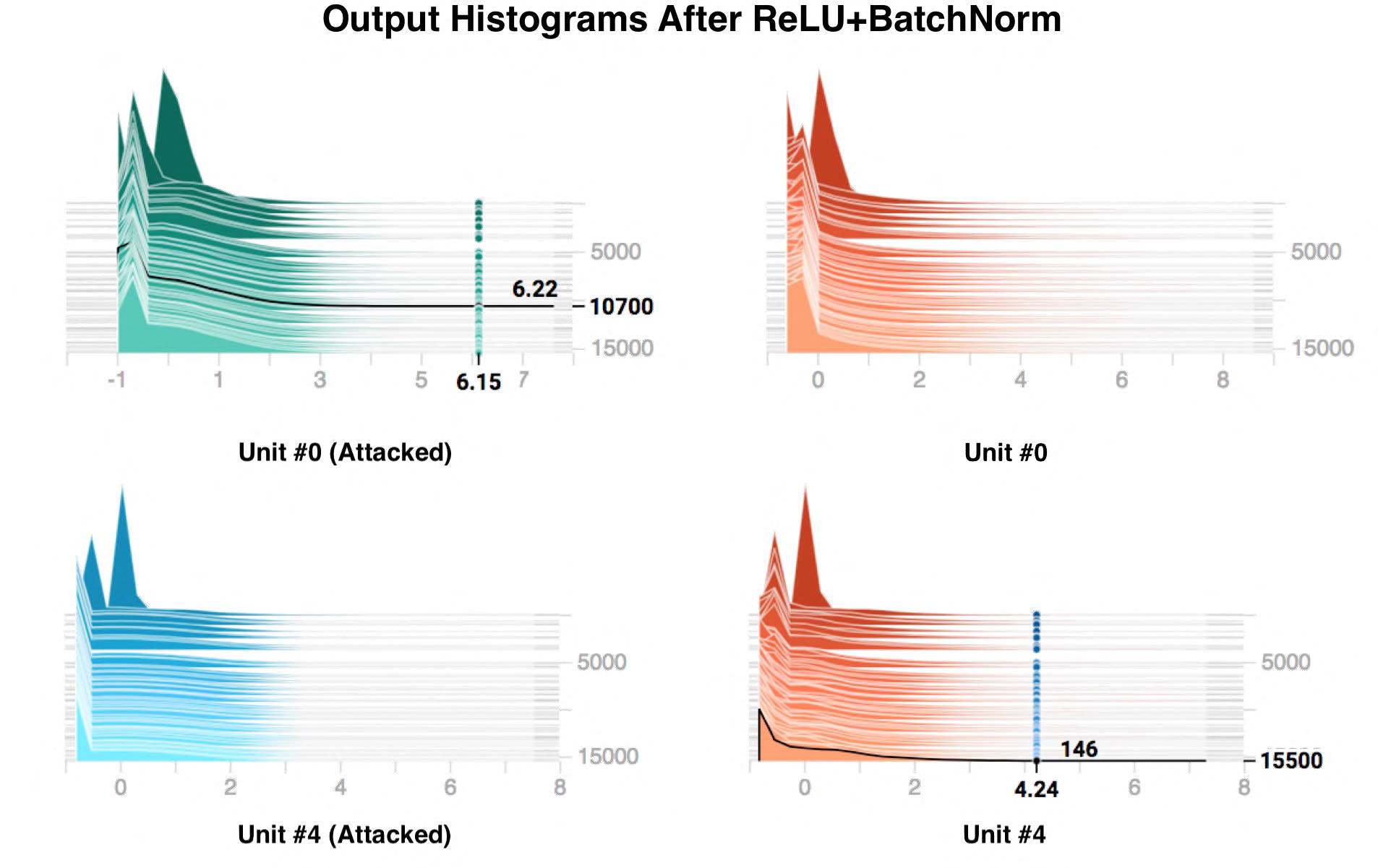}
    \caption{Histograms after ReLU+BatchNorm }
    \label{fig:3.2.2}
  \end{subfigure}
  \end{center}

  \caption[Effects of lr-attack on output distributions]{Output histograms are created from a validation set of size 1000. First row of histograms(unit 0) is an example for a unit dying due to the attack. Whereas the second row(unit 4) doesn't die after the attack and we can see that the output distribution of the unit 4 is not changed after attack and looks very similar to the original distribution.}
  \label{fig:3.2}
\end{figure}

Increased norm of the weights cause output values to increase and we get bigger values out of the linear part of the ReLU. This will cause the mean and the variation calculated by the batch normalization to increase. As a consequence, zero activations and nearby values would squeeze further to the left edge. In a sense, we might squeeze what unit learned so far towards left due to the increased values on the linear side(right).

The transformation of the output distributions for a successful and an unsuccessful attack are plotted on Figure \ref{fig:3.2}. We see that the same multiplier has two different effects on the two different units and this is very much expected considering the differences in gradients and parameters of two different units.

An abruptly shrunk left side of the output distribution is also going to alter the gradients of the outgoing weights since their gradient is given by

\begin{equation} \label{eq3.5}
\frac{\partial L}{\partial w_{ij}^{l+1}} = \frac{\partial L}{\partial h_j^{l+1}} a_i^l
\end{equation}

where $w_{ij}$ is the weight connecting our unit(i) to the unit(j) in the next layer. $a_i^l$ is the output of our unit and as reasoned above their distribution is expected to shift towards the negative side causing the weight $w_{ij}$ to receive a smaller gradient signals. Therefore we expect the outgoing weights in the attacked version to be smaller. We can see this effect in the third column of the Figure \ref{fig:3.1}.

Note that after a successful attack(like unit-0), batch normalization would have a high variance $\sigma_i^2$ dividing each value and it would shrink the incoming gradient with the same value $\sigma_i^2$. This makes $\frac{\partial L}{\partial h_i}$ to decrease (column 4 of the Figure \ref{fig:3.1}). As a result $\frac{\partial L}{\partial W_i}$ decreases, too (column 6 of the Figure \ref{fig:3.1}). This clearly shows us that units might not be dead even when they are frozen. Units may have a small gradient signal coming to them, however they might still create some useful information used by the rest of the network and therefore having a high mean replacement penalty.

\subsection{Effect of the attack is time dependent\label{sec3.4.1} }
\begin{figure}
  \begin{center}
  \begin{subfigure}{.8\textwidth}
    \centering
    \includegraphics[width=.95\linewidth]{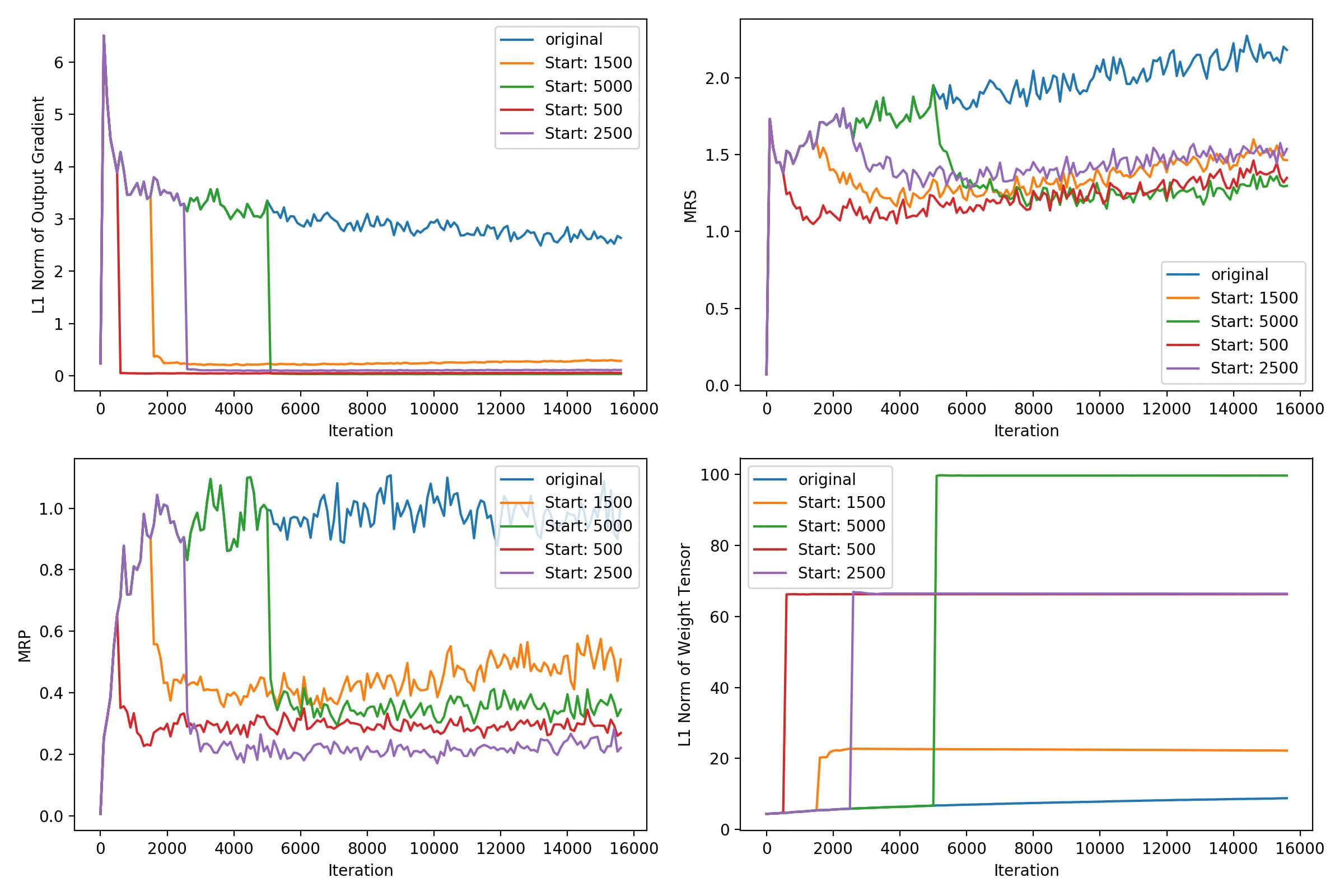}
    \caption{Layer-1 Unit-0}
    \label{fig:3.3.1}
  \end{subfigure}
  \begin{subfigure}{.8\textwidth}
    \centering
    \includegraphics[width=.95\linewidth]{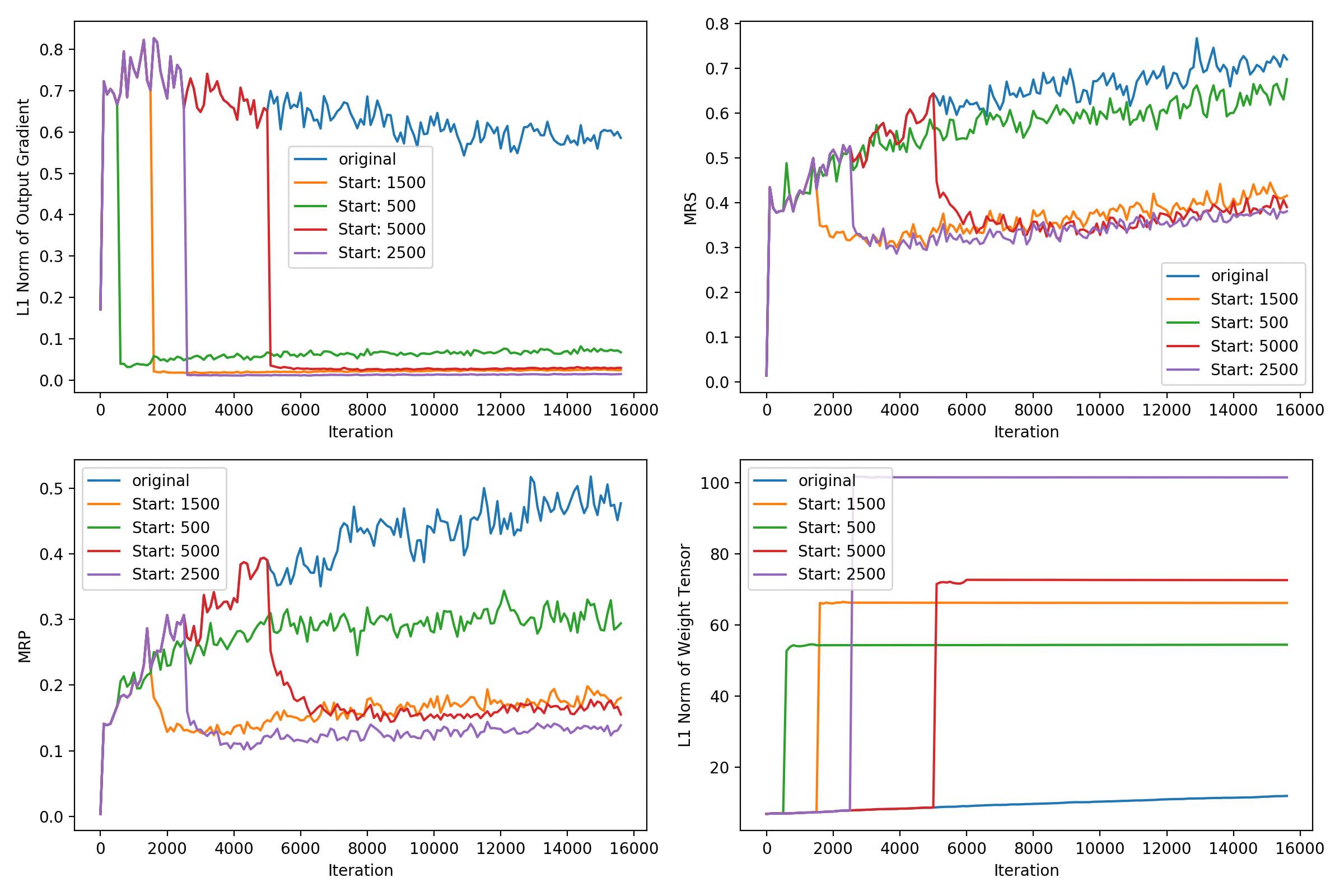}
    \caption{Layer-2 Unit-0}
    \label{fig:3.3.2}
  \end{subfigure}
  \end{center}

\caption[Effect of the starting iteration of the learning rate attack]{Various metrics observed during 4 different attacks each starting at a different time-step. The x-axis represents the time (iteration) and all vector values are plotted after taking their l1-norm. We can clearly see that there is no clear relation between time and the success of a learning rate attack. Corresponding loss curves plotted in Figure \ref{fig:A4.1} and \ref{fig:A4.2}}
  \label{fig:3.3}
\end{figure}
In the previous section, we showed the responses of unit 0,4 and 2 of the first convolutional layer against an attack with an early start (attacks starting at 50th step). In this section, we like to continue our experiments by changing the time we start attacking the unit. We are attacking the unit 0 in the first convolutional layer 10 times with 10 iteration intervals starting from 500th, 1500th, 2500th and 5000th steps in different runs and results are plotted in Figure  \ref{fig:3.3}\subref{fig:3.3.1}. We clearly see that the most successful attack is the one starting on the 2500th iteration (step) and the second most successful attack is the one start at 500th iteration, whereas the attack starting on 5000 and 1500 gets the last two spot. We observe that MRS does not capture the ordering perfectly even though it is successful at separating the original unit from attacked ones.

On Figure \ref{fig:3.3}\subref{fig:3.3.2}, we have a very similar picture, where the success of an attack seems to be not correlated to the time we start the attack. However this time we see that the MRS makes a very good job separating low MRP units from the rest.

Another interesting observation that we can make in Figure \ref{fig:3.3} is the negative correlation between the weights and gradient signals coming to the unit(output gradient). It looks like output gradient of a unit decreases proportional to the increase in the norms of the corresponding weights and this aligns well with our explanation in the previous section.

\subsection{Increasing learning rate multiplier causes units to die\label{sec3.4.2} }
\begin{figure}
  \begin{center}
  \begin{subfigure}{.8\textwidth}
    \centering
    \includegraphics[width=.94\linewidth]{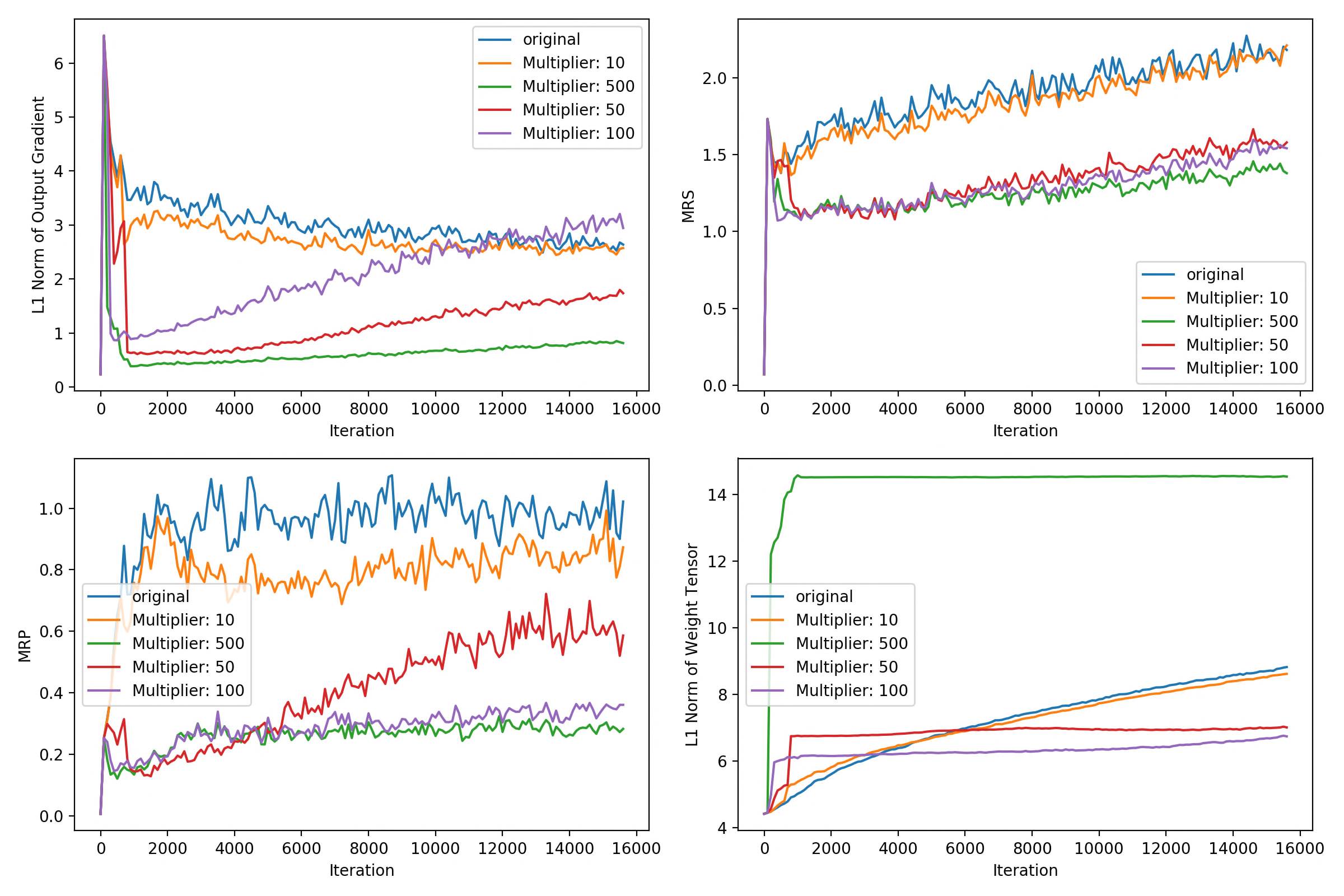}
    \caption{Layer-1 Unit-0}
    \label{fig:3.4.1}
  \end{subfigure}
  \begin{subfigure}{.8\textwidth}
    \centering
    \includegraphics[width=.94\linewidth]{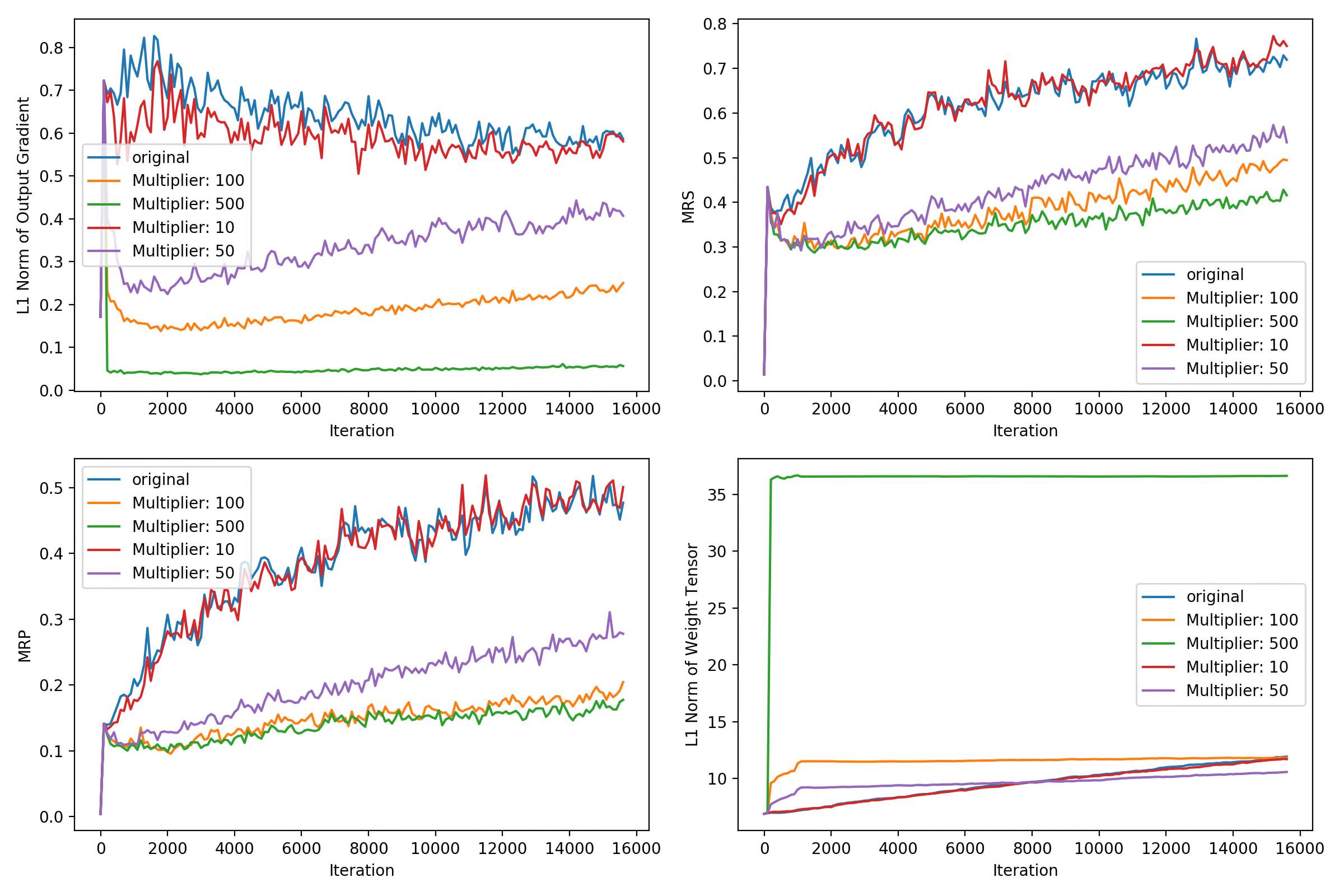}
    \caption{Layer-2 Unit-0}
    \label{fig:3.4.2}
  \end{subfigure}
  \end{center}

\caption[Effect of the strength of the attack]{Various metrics observed during 4 different attacks with different learning rate multipliers, each starting at 50th step and repeats 10 time with 100 step intervals with . X-axis represents the time(steps taken) and all vector values are plotted after taking their l1-norm. The success of the attack increases with the multiplier. Corresponding loss curves can be seen in Figure \ref{fig:A4.3} and \ref{fig:A4.4}}
  \label{fig:3.4}
\end{figure}

In Figure \ref{fig:3.4} we see two groups of plots corresponding to the experiments where we attack the first units of the first and second layers respectively. Looking to the Figure \ref{fig:3.4}\subref{fig:3.4.1}, one can clearly see that the success of an attack (drop in MRP) increases with the learning rate multiplier. We don't see any clear relation between the MRP and magnitude of the weights. However, an interesting observation is that the norm of the output gradient is not necessarily small for low MRP units. Looking to the blue curve we see that the norm of the output gradient exceeds the original run towards the end of the training despite having a much lower MRP.

In Figure \ref{fig:3.4}\subref{fig:3.4.1}, the first unit of the second convolutional layer responds to the increasing multiplier in a similar manner, i.e. MRP's drops more, weights jump more and MRS do a very good job at approximation the MRP. This time we also see a very correlated picture between output gradients and weights.

Results of the experiments in this section clearly shows us that that it is the attack(high learning rate) that makes units die.

\section{Killing a Unit with Tanh\label{sec3.5}}
In this section, we switch to the \textit{TanH} nonlinearity and perform 10 attacks starting from the 2000th step with 10 step intervals between each attack and using a learning multiplier of 1000. Since it takes more time to train network with \textit{TanH}, we let the experiment run for 30 epochs instead of 10. The network reaches to an accuracy of 64\%.
\begin{figure}
  \begin{center}
    \includegraphics[width=.9\linewidth]{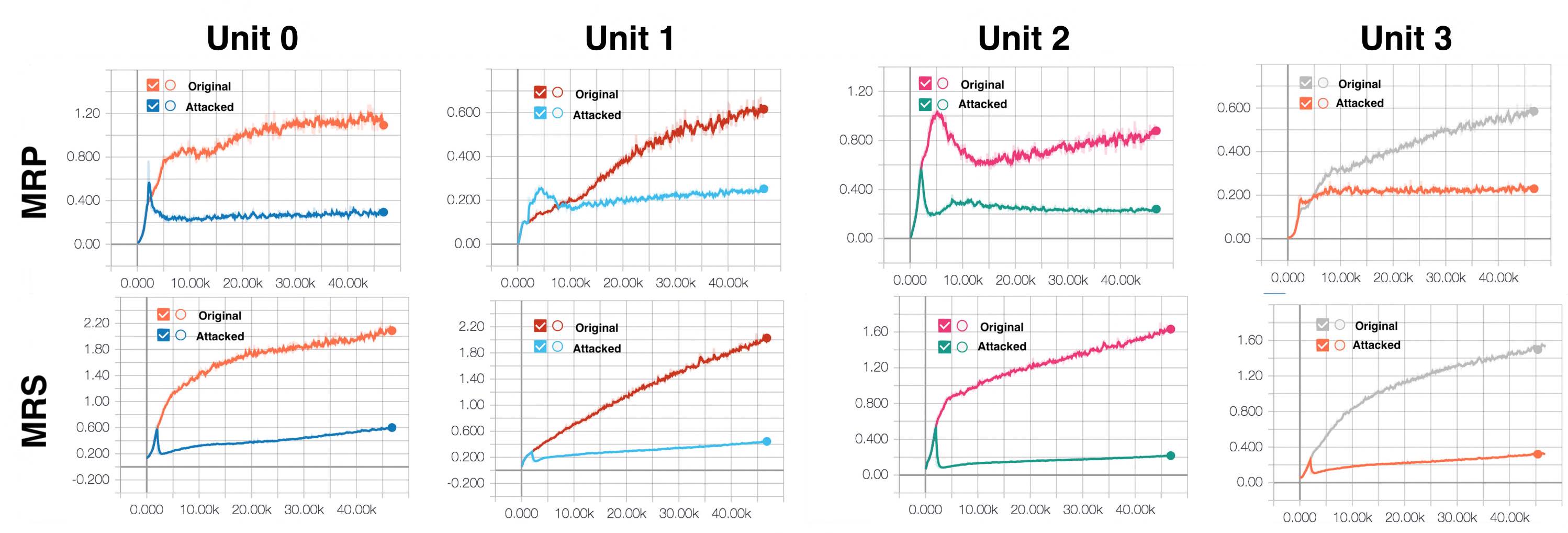}
  \end{center}%
\caption[MRP and MRS of attacked units with \textit{TanH}]{MRP and MRS of various units plotted against time X-axes represents the number of steps (iteration) taken. We see all attacks are somewhat successful at limiting the MRP of a unit.}
\label{fig:3.5}
\end{figure}

Figure \ref{fig:3.5} shows MRP and MRS of four different units from the first convolutional layer with and without the attack performed. The experiments without an attack labeled as \textit{Original}. Two of the units have final MRP's around twice more than the other two, however, they all drop to an MRP of 0.2-0.3 after the attack and MRS does a very good job approximating the MRP.

\begin{figure}
  \begin{center}
    \includegraphics[width=.9\linewidth]{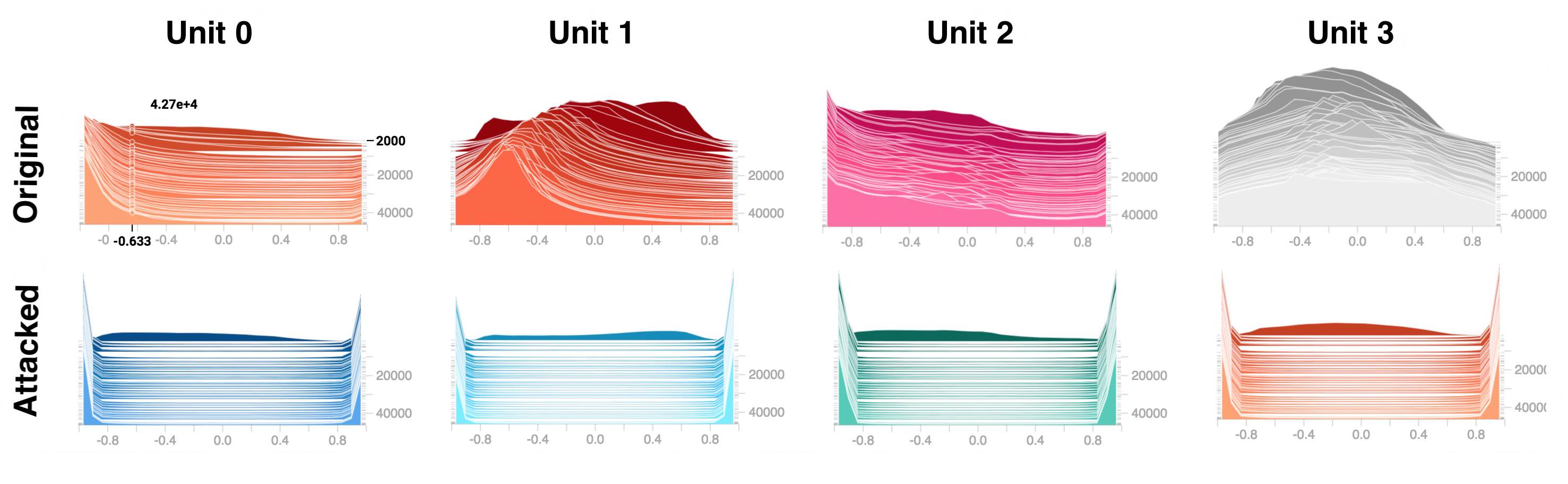}
  \end{center}%
\caption[Activation Histograms of attacked units with \textit{TanH}]{First row shows the normalized histograms of activations of each unit on the validation set of size 1000 without an attack. Second row is the same histogram when the unit is attacked 10 times between (2000,2100)th steps. The z-axes of the histograms represent the iteration number.}
\label{fig:3.6}
\end{figure}

As happened before the output distribution of the attacked units scaled considerably due to the relatively large updates caused by the attack. This shifts the activation distribution to the edges. In other words, most of the input distribution is mapped into the flat region of \textit{TanH}. In a very basic sense, the amount of information available at the end of the layer decreases considerably. We can see this behavior in Figure \ref{fig:3.6} where we plot the activation distribution's of the attacked units using the same units as in Figure \ref{fig:3.5} and same colors. The output of first three units (0,1,2) in Figure \ref{fig:3.6} shifts toward the left towards the end of the training. This gives us a clue, that if we perform the attack at a different time we might end up in an extremely uneven activation distribution and therefore have even lower MRP's.

\subsection{All you need is the right timing\label{sec3.5.1} }
\begin{figure}
  \begin{center}
  \begin{subfigure}{.8\textwidth}
    \centering
    \includegraphics[width=.94\linewidth]{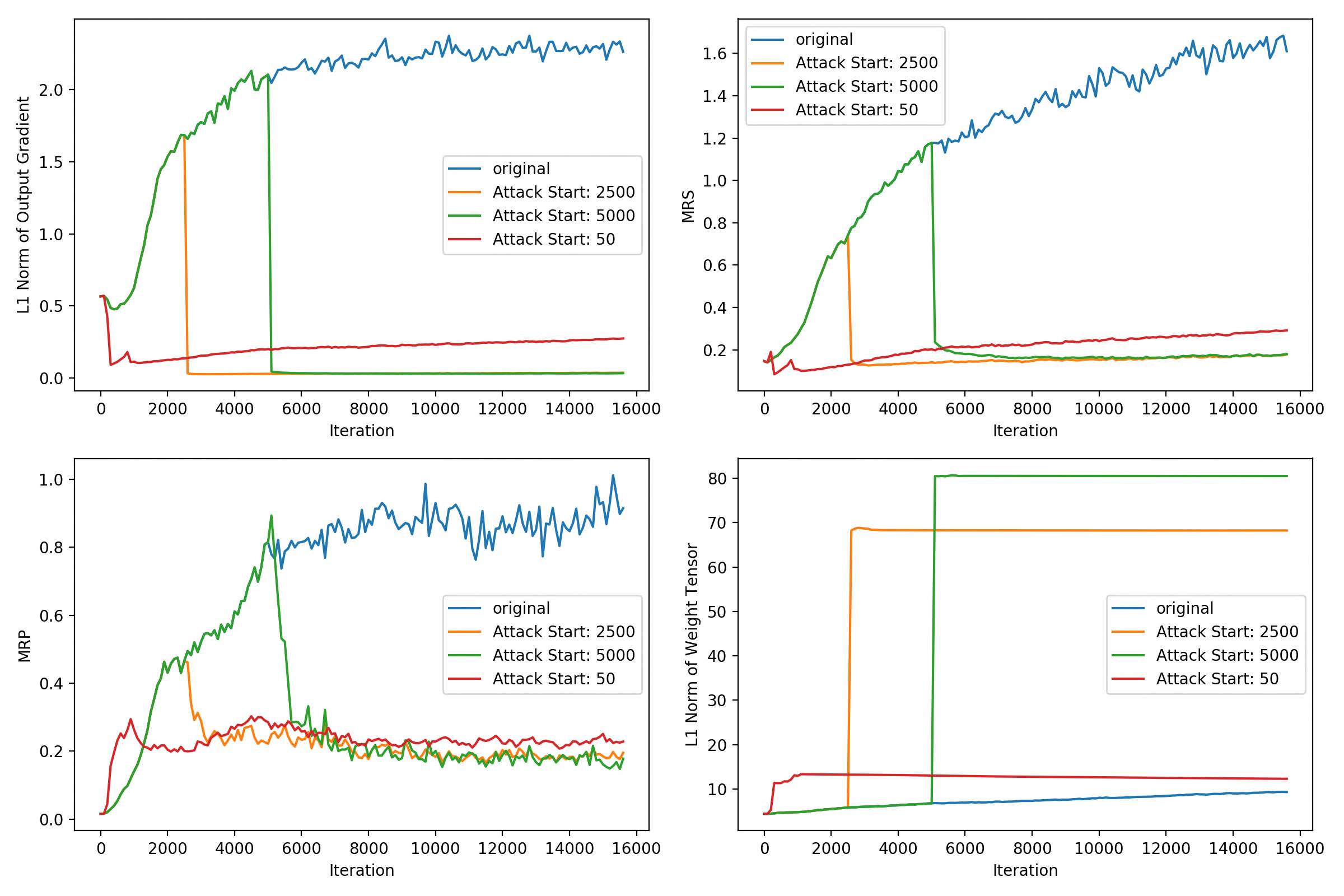}
    \caption{Layer-1 Unit-0}
    \label{fig:3.7.1}
  \end{subfigure}
  \begin{subfigure}{.8\textwidth}
    \centering
    \includegraphics[width=.94\linewidth]{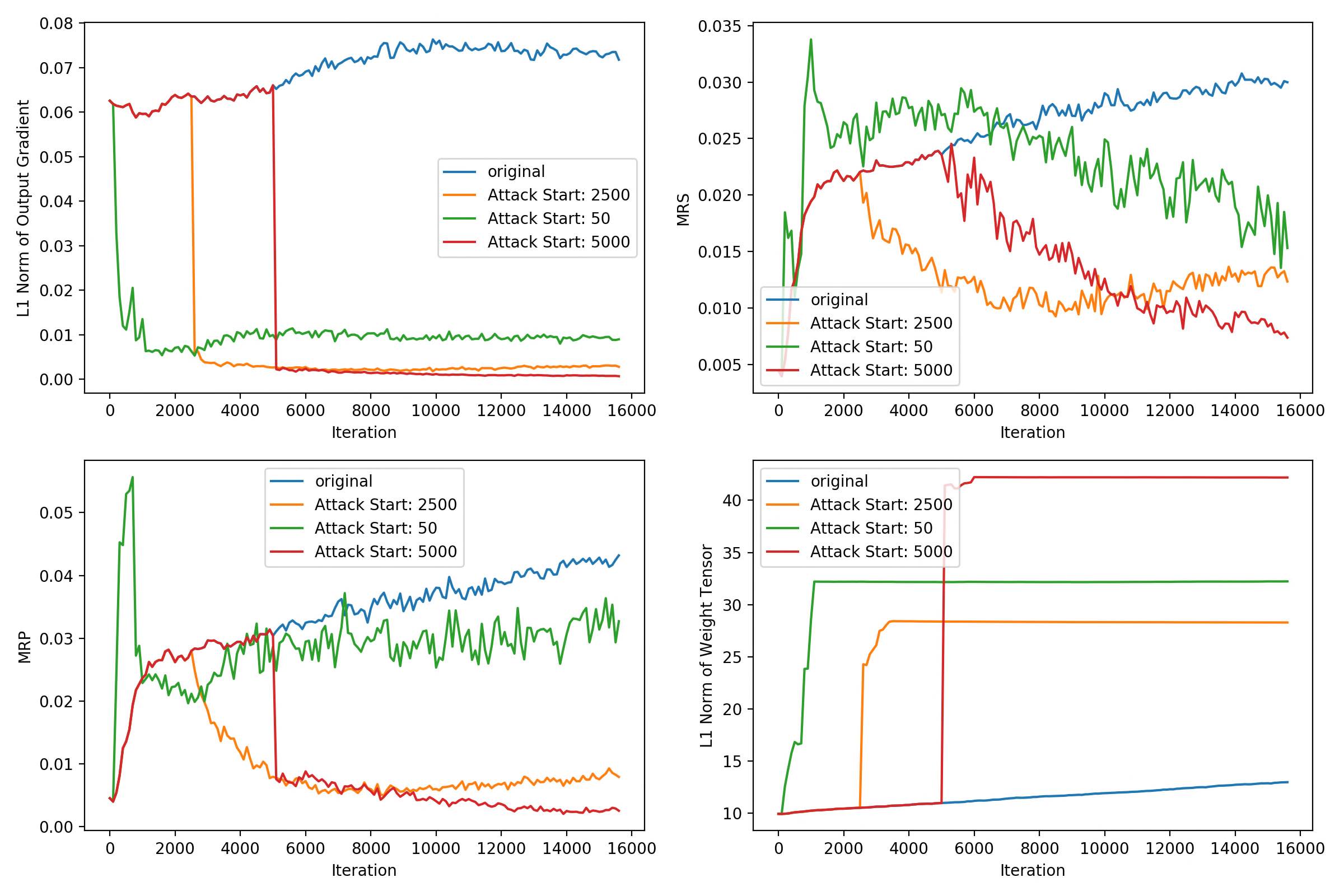}
    \caption{Layer-3 Unit-16}
    \label{fig:3.7.2}
  \end{subfigure}
  \end{center}

\caption[Effect of the start of the attack on units with \textit{TanH}]{Various metrics observed during 3 attacks with different starting points. Each repeats 10 times with 10 step intervals. X-axis represents the time(steps taken) and all vector values(Output Gradient and Weights) are plotted with l1-norm. Corresponding loss curves plotted in Figure \ref{fig:A4.5} and \ref{fig:A4.6}}
  \label{fig:3.7}
\end{figure}

In this section, we attack the unit-0 of the first layer and unit-16 of the third layer (first fully connected layer) 10 times with different starting times. Results taken from these two experiments are plotted in Figure \ref{fig:3.7}. Once again we observe correlation between \textit{Output Gradient} and \textit{Weights} after the attack in both experiments.

In Figure \ref{fig:3.7}\subref{fig:3.7.1} we see that all three attacks causes similar drops on the MRP score. However, we see significant differences on the norms of weights and output gradients. One subtle observations is that the attack starting at 50th step increases the MRP at the beginning, however the increase stops after a while (red curve). However, we observe that MRS captures the ranking perfectly throughout the training.

\begin{figure}
  \begin{center}
  \begin{subfigure}{.8\textwidth}
    \centering
    \includegraphics[width=.94\linewidth]{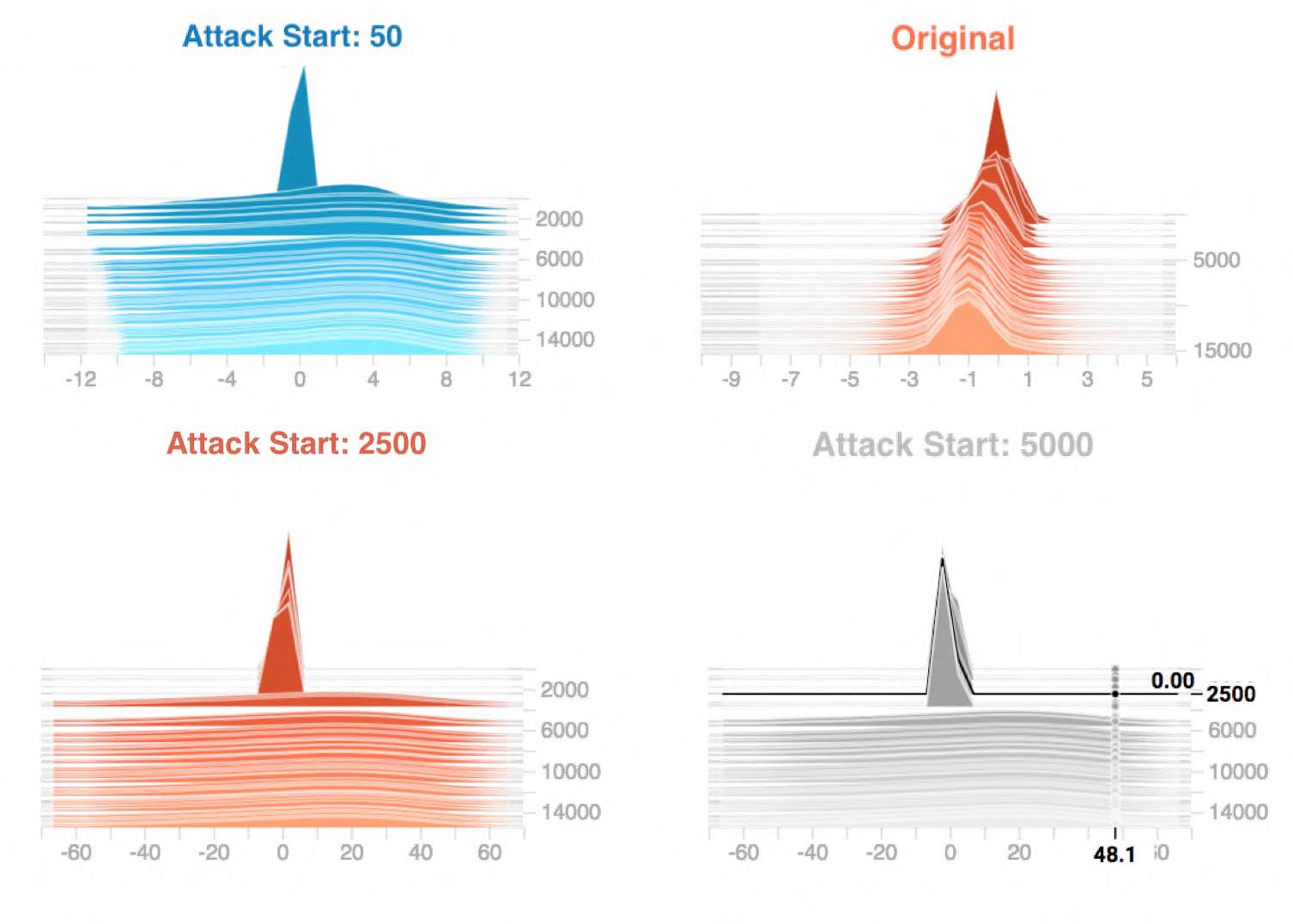}
    \caption{Layer-1 Unit-0}
    \label{fig:3.8.1}
  \end{subfigure}
  \begin{subfigure}{.8\textwidth}
    \centering
    \includegraphics[width=.94\linewidth]{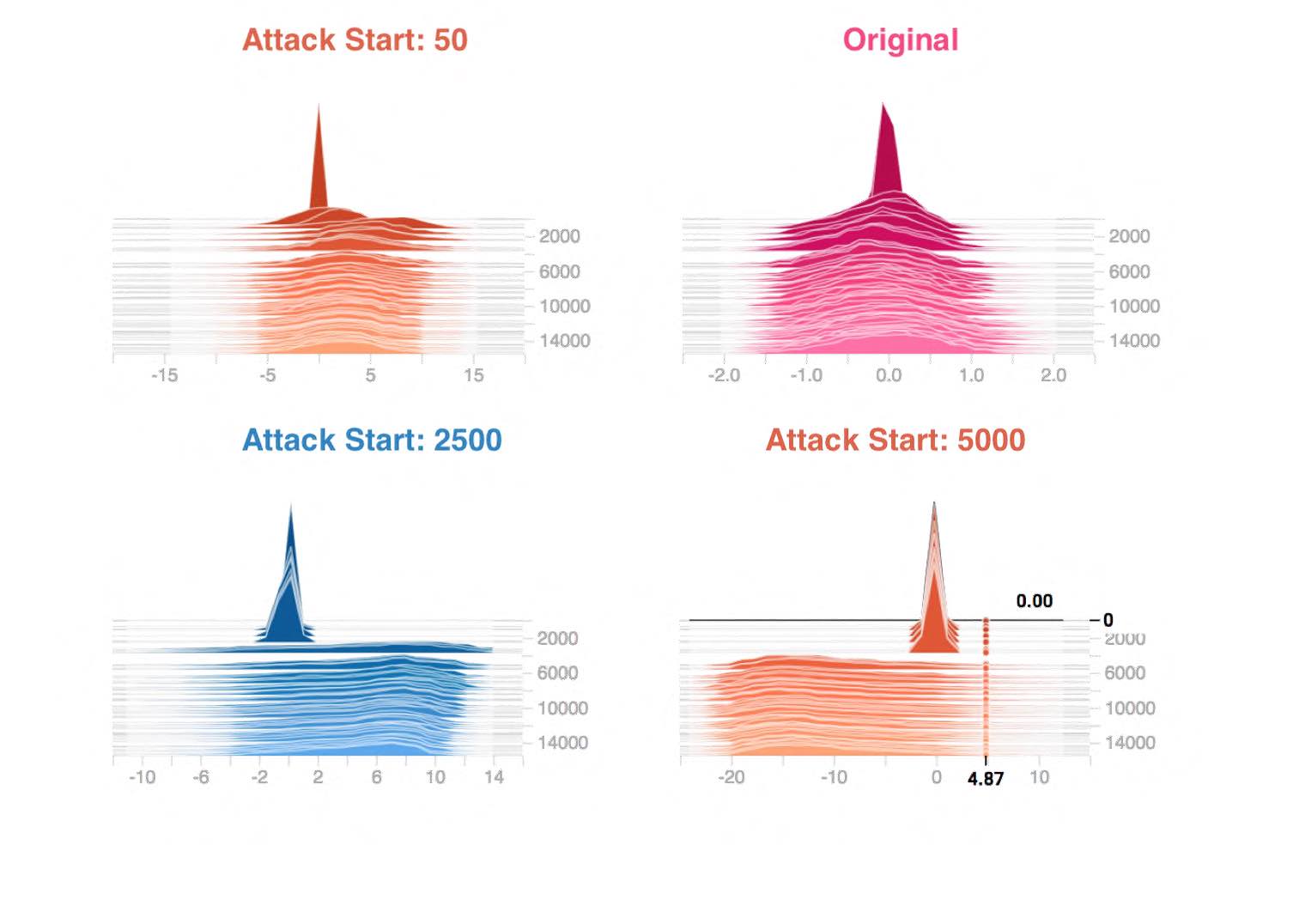}
    \caption{Layer-3 Unit-16}
    \label{fig:3.8.2}
  \end{subfigure}
  \end{center}

\caption[Output histograms of units with different attack start using \textit{TanH}]{Output histograms of two units. x-y-z axes of each histogram represents value, normalized count and step-taken (itereation) respectively. Attacks started later during the training seems like generating more values on the flat regions(left-right edges(-1,1) of the histogram). }
  \label{fig:3.8}
\end{figure}

In Figure \ref{fig:3.7}\subref{fig:3.7.2} attacks starting at 50th step increase the MRP considerably, however, these effects are fadeout later during the training. We suspect this behavior being due to the small-magnitude-gradients. An early attack may not be enough to saturate the activations. To understand this phenomenon further we plot the histogram of the unit outputs and activations of unit-16 in Figure \ref{fig:3.8}.

The attacks started at 50th step in both experiments doesn't scale the output distribution considerably and therefore the MRP of the unit increases. However attacks performed later (at 2500 and 5000) cause a wider distribution and cause more outputs to saturate. Note that the attacks starting at 2500th and 5000th step causes activation distributions leaned towards the right and left respectively. This shows that the effect of the learning rate attack depends heavily on the step it is performed.

\section{Killing a Unit with Relu \label{sec3.6}}
In the previous two sections, we showed that the learning rate attack disturbs the activation distribution and often causes a drop in MRP. We also reasoned that this effect seems like due to the reduced information produced by the unit. In this section, we would like to remove the batch normalization and repeat our experiment on a unit with ReLU activation. All experiments in this section run for 10 epochs and the network reaches to an accuracy of 61\% independent of whether an attack performed or not. We perform learning rate attacks starting at 2000th step of the training again 10 times with 10 iteration intervals.

In a network with ReLU nonlinearities, it is harder to lose information since one side of the activation function is not bounded. If our output distribution is centered around 0, a learning rate attack would not saturate the non-zero activations; it would only scale them such that the information is preserved. Moreover increased magnitudes may generate bigger gradients on the outgoing weights and make the information provided by the unit more important for the task. We may even see faster convergence (for real).

\subsection{Whatever doesn't kills you make you stronger \label{sec3.6.1} }
\begin{figure}
  \begin{center}
  \begin{subfigure}{.8\textwidth}
    \centering
    \includegraphics[width=.94\linewidth]{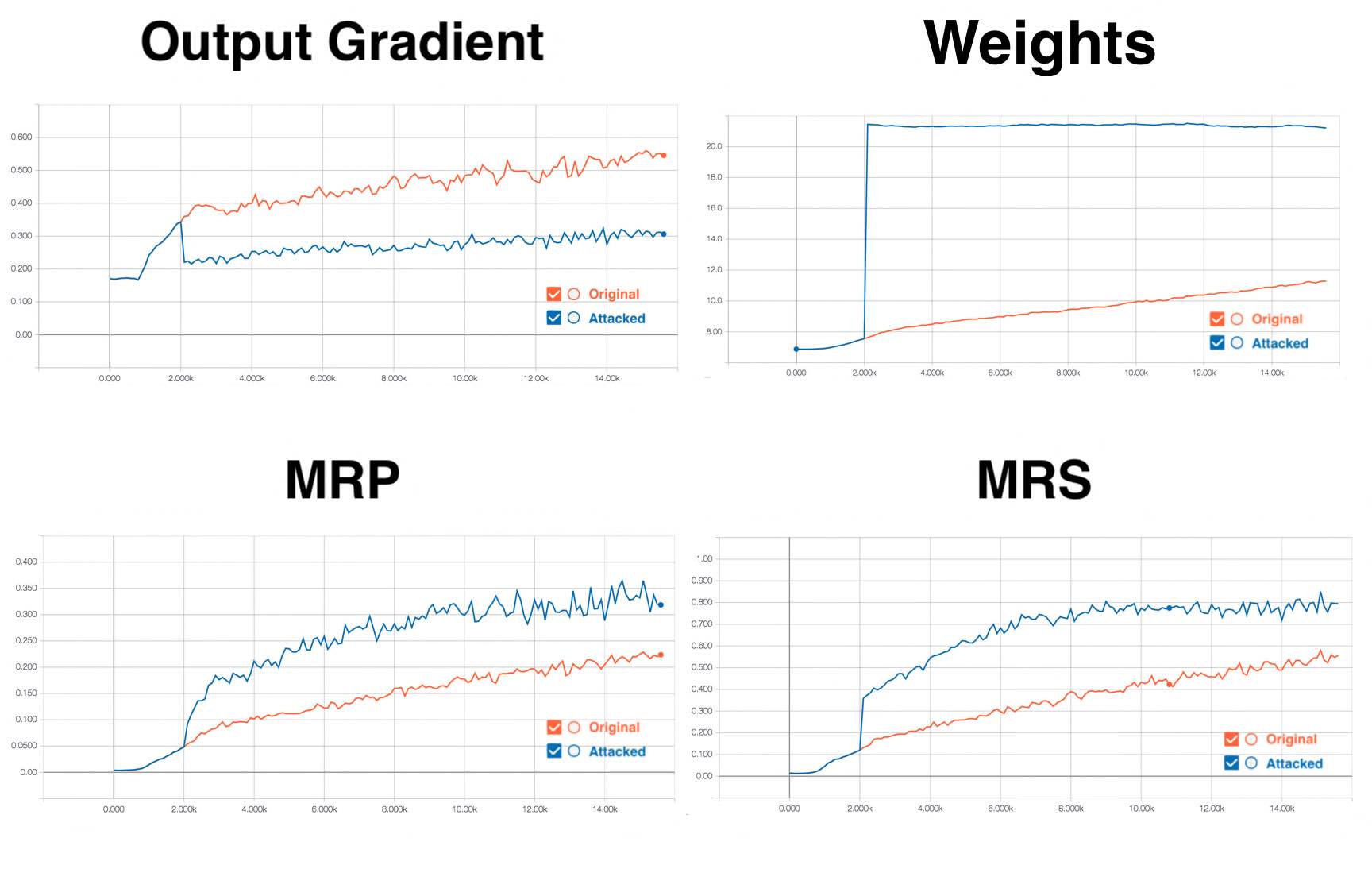}
    \caption{Layer-2 Unit-0}
    \label{fig:3.9.1}
  \end{subfigure}
  \begin{subfigure}{.8\textwidth}
    \centering
    \includegraphics[width=.94\linewidth]{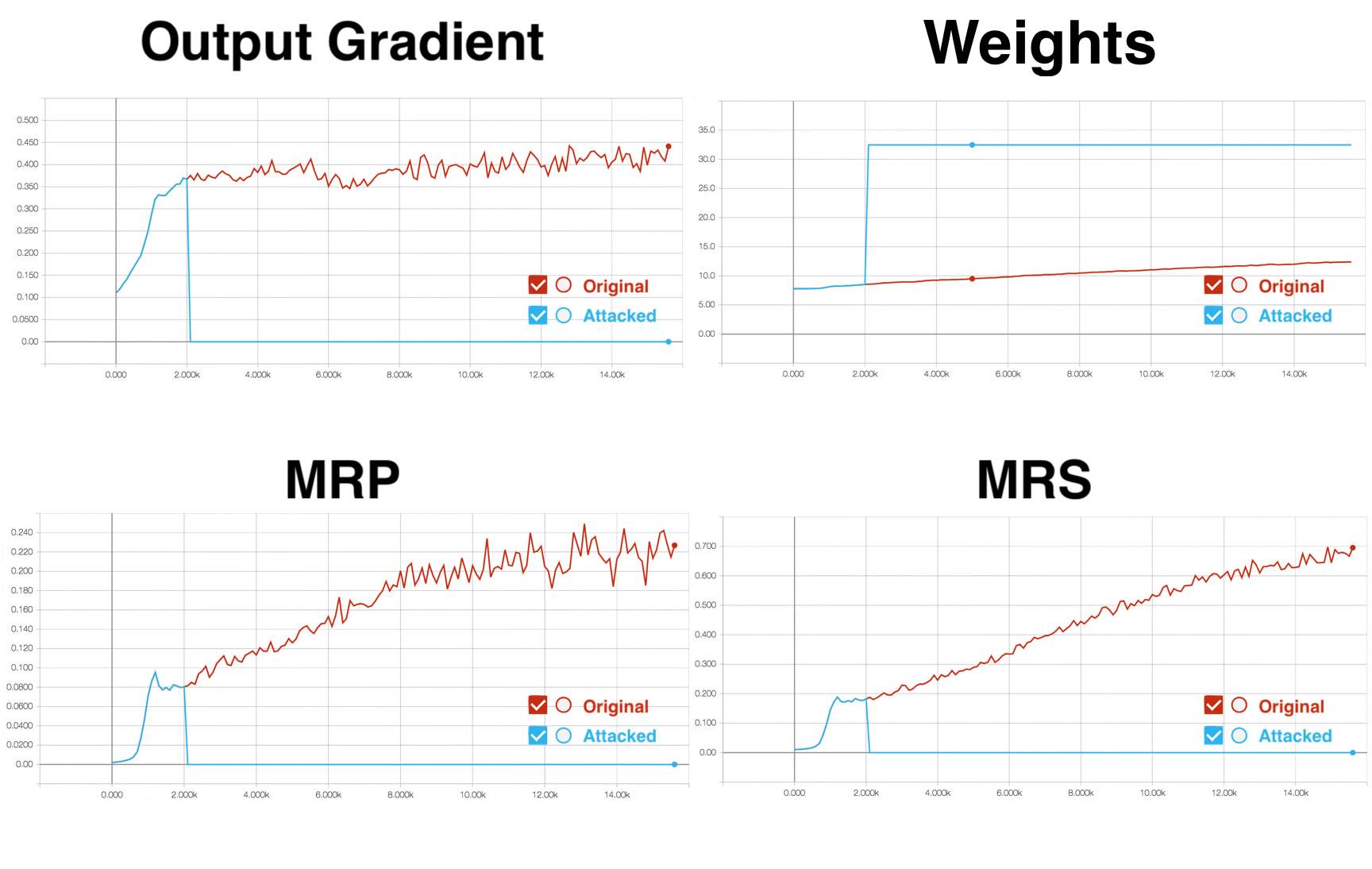}
    \caption{Layer-2 Unit-4}
    \label{fig:3.9.2}
  \end{subfigure}
  \end{center}

\caption[1 successful and 1 unsuccessful attack on Relu units]{Various metrics observed during an attack on 2 different units. Both attacks start on 2000th step. The attack on unit 0 makes the unit more important for the network (increased MRP), whereas unit 4 is completely dead after attack with 0 MRP.}
  \label{fig:3.9}
\end{figure}
Due to its unbounded nature, we had hard time attacking ReLU. Often, we observed that attacking a ReLU unit cause an increase in MRP. An example of this effect is given in Figure \ref{fig:3.9}\subref{fig:3.9.1}. The attack on Unit-0 of Layer-2 reduces the output gradient, however, increases the MRP. Most of the units we attacked responded the attack in this manner.

\begin{figure}
  \begin{center}
    \includegraphics[width=.8\linewidth]{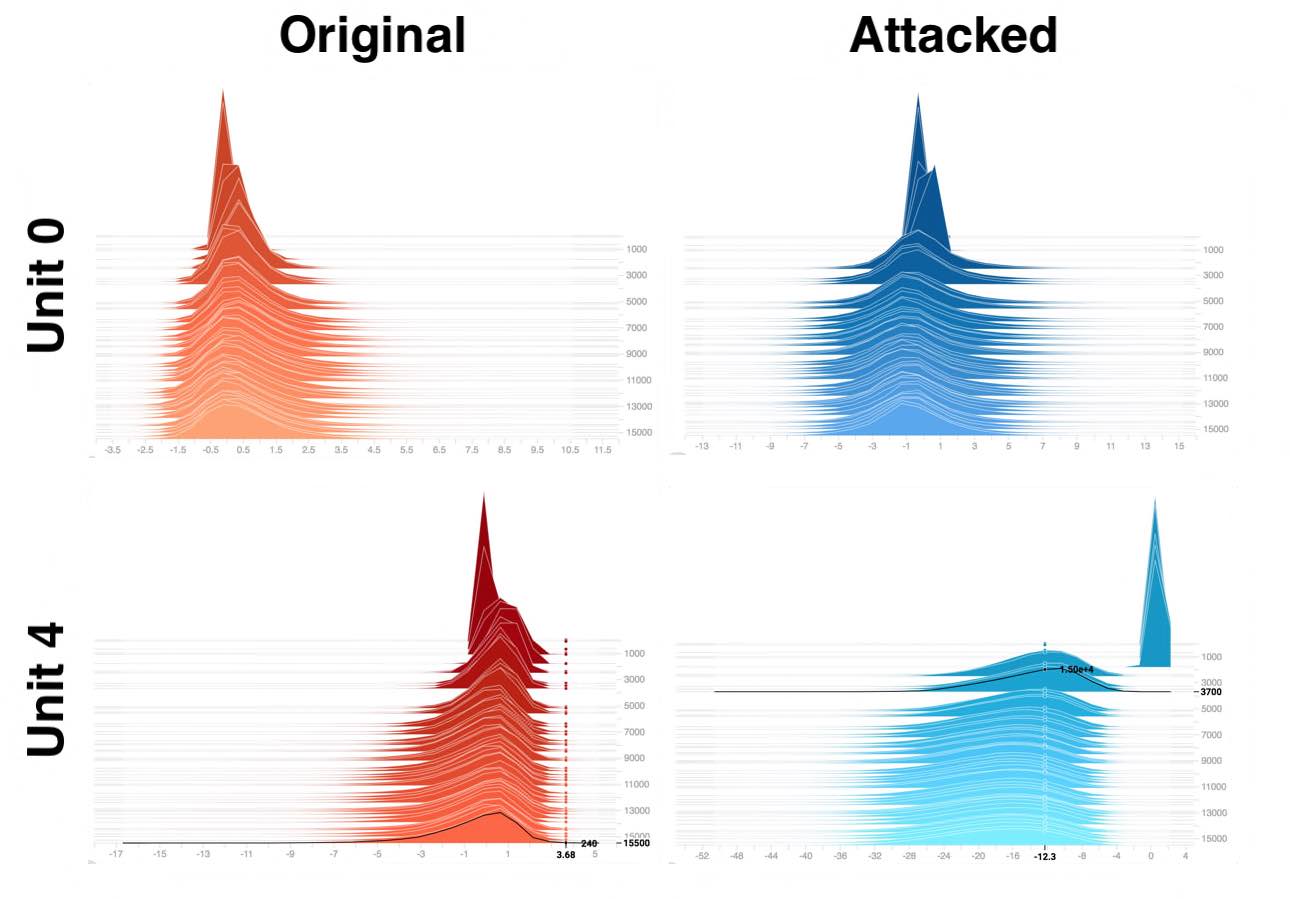}
  \end{center}%
\caption[Output Histograms of attacked units with Relu]{x-y-z axes of each histogram represents value, normalized count and step-taken respectively. The attack on the unit makes the original distribution wider with a factor of roughly 2, whereas the attack on unit 4 has a different effect, where the whole distribution is shifted to the negative side.}
\label{fig:3.10}
\end{figure}

However, we observed some units(2 out of 4 at second layer) responded to the attack in a different way. In Figure \ref{fig:3.9}\subref{fig:3.9.2} we see a unit that died completely after a learning rate attack. This is extremely interesting since the whole output distribution of the unit is shifted to the negative side after the attack (see Figure \ref{fig:3.10}).

This shows us that even though it is less likely to see a successful learning rate attack in networks with ReLU, it is still possible. It is all possible.

\section{Dead units may develop and MRS is an efficient measure to detect them\label{sec3.7} }
In this section, we observed unit with 3 different activation functions. For all of them, we were able to perform successful attacks where we see a considerable drop in MRP. Except for couple instances, MRS did a very good job approximating the MRS and never failed at choosing the smallest MRP unit. We expect MRS to perform even better with smaller MRP's due to the decreased error term in the Taylor decomposition.

We showed that ReLU+Batch Normalization activation behaves like a bounded activation function similar to \textit{TanH}. Therefore their units are more likely to get affected by the learning rate attack. On the other hand, ReLU units are harder to harm with learning rate attack, however possible. We suspect that increased MRP on the attacked units causes slight decrease on MRP of other units. However, we haven't experiment with this hypothesis.

Measures like \textit{Output gradient} and \textit{l1-norm of the weights} are definitely related to the MRP, however, they are worst than the MRS at predicting low MRP units. Contrary to our expectations, we showed that a low MRP unit might have a very big norm (of weights) and pruning a unit according to its norm might be a very bad idea.

We also showed that the success of an attack is very related to the iteration it is made and the learning rate multiplier. We haven't investigated the effect of the outgoing connections on the success of an attack.

Note that we investigated only a small subset of all possible learning rate attacks. However, we believe that our findings are enough to convince the reader that a fix learning rate might cause some units to die and we may detect them efficiently using MRS.

\chapter{MRS based pruning with bias propagation\label{chap4}}
In Chapter \ref{chap2} we defined and compared various saliency scores for individual parameters. However, when we look at the parameters of the pruned network, we observed that the individual units gather around some specific units. That motivated us to focus on units rather than the individual parameters. In Chapter \ref{chap3} we defined the dead units and the Mean Replacement Penalty(MRP). Our motivation was that the MRP is the right measure for detecting units that either doesn't create any useful information or ignored by the rest of the network. We argued that the dead units are unlikely to change over the course of the training and we empirically observed this behavior. We also mentioned that replacing a unit with its mean and propagating that to the next layer is better than a bare removal. MRP directly measures this penalty of replacing a unit with its mean.

Calculating MRP for all units in a network is not feasible for large networks. However, we can calculate the MRS, first-order approximation of MRP, quite efficiently and use it to prune units. We can calculate MRS during the backward pass in linear time and in Chapter \ref{chap3} we observed that MRS works pretty good at approximating the MRP, at least for the experiments we performed. In Section \ref{sec4.1} we would like to compare MRS with norm-based saliency scores using the \textit{randomScorer} as our baseline. In Section \ref{sec4.2} we demonstrate unit pruning using the MRS and get superior results compare to the results of Chapter \ref{chap1}.

\section{Comparing MRS with norm-based saliencies\label{sec4.1}}
In this section, we compare MRS saliency score with MRP and norm-based scoring functions. Our investigation is somehow limited since we only compare the scoring functions on a small network trained for MNIST classification. However, we believe that MRS would perform as good as norm-based saliency measures if not better in other networks and settings, too.

\cite{molchanov2016} compares various saliency scores in the context of unit pruning without bias propagation and concludes first-order Taylor approximation performs best. \cite{ye2018} uses bias propagation(explained in Chapter \ref{chap1}) idea to remove units which generate constant values. As stated earlier, mean replacement is a better way of removing units and should be employed. Combining the idea of mean replacement and the superiority of first-order Taylor approximation, we expect the MRS to perform better than the rest.

Saliency measures considered in this section are MRS, MRP and L1 and L2-squared based norms of the parameter tensors. Table \ref{table:4.1} summarizes these scores along with their function names in \textbf{pytorchpruner.unitscorers} module. We included MRP to see how well the MRS approximates the MRP.

\begin{table}[ht]
\begin{center}
\begin{tabular}{|c|c|}
  \hline
  \textbf{Saliency Score} & \textbf{Definiton}  \\
 \hline
 mrpScorer &see Chapter \ref{chap3}\\
 \hline
 mrsScorer &see Chapter \ref{chap3}\\
 \hline
 normScorerL1 & $||w||_1$\\
 \hline
 normScorerL2 & $||w||_2^2$\\
 \hline
 randomScorer & uniform\\
 \hline
\end{tabular}
\end{center}
\caption[Saliency scores for units]{Saliency scores that we compare in Section \ref{sec4.1}.}
\label{table:4.1}
\end{table}

In our experiments we use the same experimental setup as the one in Section \ref{sec:2.1}: small CNN on MNIST trained with SGD, batch-size of 32, the constant learning rate of 0.01  and without using momentum or weight decay for 10 epochs (18k+ iterations). We use ReLU as our activation function unless stated otherwise. We calculate the loss changes after performing the bias propagation calculated on a subset of the training set with size 1000. We expect these 1000 random samples to successfully approximate the training set accurately. Our small CNN reaches to 0.046 training loss and shows a pretty good generalization with 0.039 test loss and 98.8\% accuracy. After every pruning followed by the calculation of the loss change, we revert all the changes such that the network trains as normally.

\subsection{If you have few units, there is not much difference\label{sec4.1.1} }
\begin{figure}[ht]
  \begin{subfigure}{.48\textwidth}
    \centering
    \includegraphics[width=.9\linewidth]{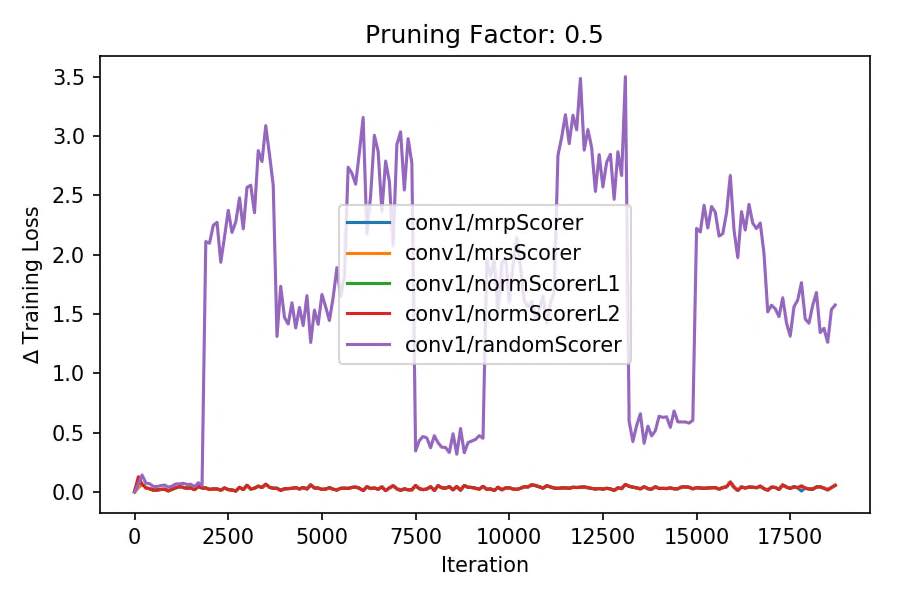}
    \caption{randomScorer included}
    \label{fig:4.1.a}
  \end{subfigure}
  \begin{subfigure}{.48\textwidth}
    \centering
    \includegraphics[width=.9\linewidth]{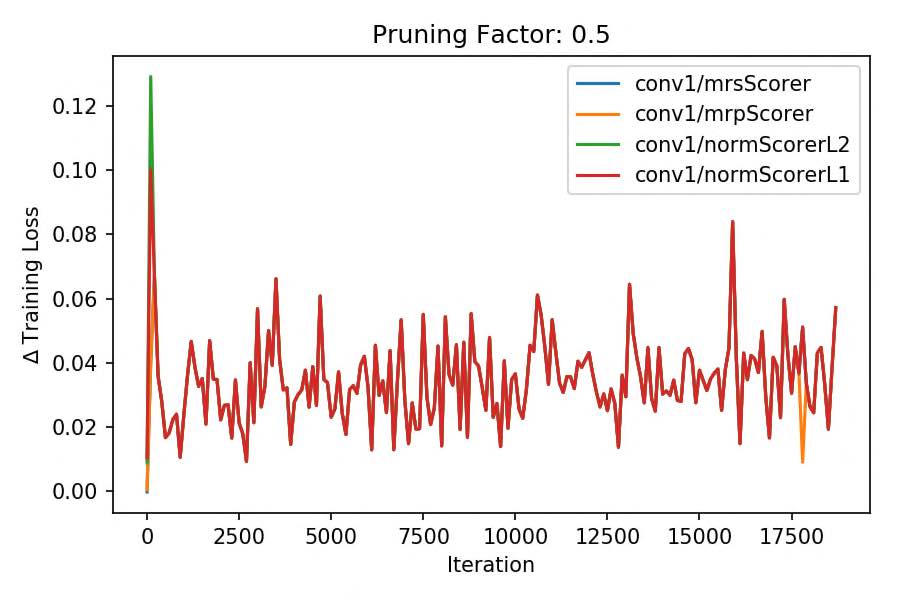}
    \caption{randomScorer not included}
    \label{fig:4.1.b}
  \end{subfigure}
  \caption[Comparison of unit saliency scores: first layer]{Change in the loss after replacing half of the units with their mean values in the first convolutional layer according to various measures explained in Section \ref{sec4.1}. Figure (a) is the same as the (b) except it includes our baseline \textit{randomScorer} in addition to the other four.}
  \label{fig:4.1}
\end{figure}

We would like to start our comparison by looking at the very first convolutional layer. Since MNIST is a relatively easy task, even a small CNN with 8 units in the first layer obtains a pretty good accuracy. We prune 4 out of 8 units and calculate the change in the loss every 100 iterations. After every measurement, we revert the changes so that our measurement address the question of what if we prune at given iteration using a particular saliency measure. Changes in the training loss measured every 100 iterations during the 10 epochs of training for the first convolutional layer plotted in Figure \ref{fig:4.1}\subref{fig:4.1.a}. Since there are only 8 units in the first layer, we don't see much difference between different saliency measures. They all seem like selecting the same 4 units during the entire training. But there is a significant difference between the \textit{randomScorer} and the rest (Figure \ref{fig:4.1}\subref{fig:4.1.b}). Some units have a very significant influence on the result and therefore pruning them with mean replacement causes a very significant change in the loss value(up to 100x more).

\subsection{Dead units having small norm\label{sec4.1.2} }
The results in \ref{fig:4.1}\subref{fig:4.1.a} also suggests that units with low MRP also have small norm. In Section \ref{sec3.2}, we argued that dead units would most likely also be frozen (having no gradient signal). We also know that the parameters of the neural network tend to increase during the training and we also confirmed this tendency in our experiments. These two observations lead us to the opinion that the dead units(low MRS) score units may prevent the norm to increase and therefore making the norm-based saliency scores selecting the same units as the MRP selects.

\subsection{MRP and MRS works better in general with ReLU \label{sec4.1.3} }
\begin{figure}[ht]
  \begin{subfigure}{.48\textwidth}
    \centering
    \includegraphics[width=.9\linewidth]{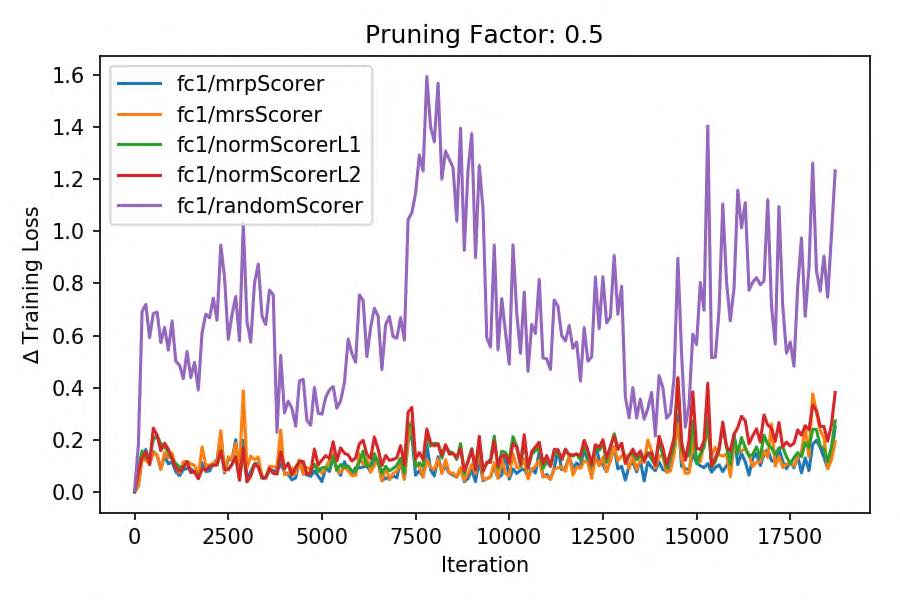}
    \caption{randomScorer included}
    \label{fig:4.2.a}
  \end{subfigure}
  \begin{subfigure}{.48\textwidth}
    \centering
    \includegraphics[width=.9\linewidth]{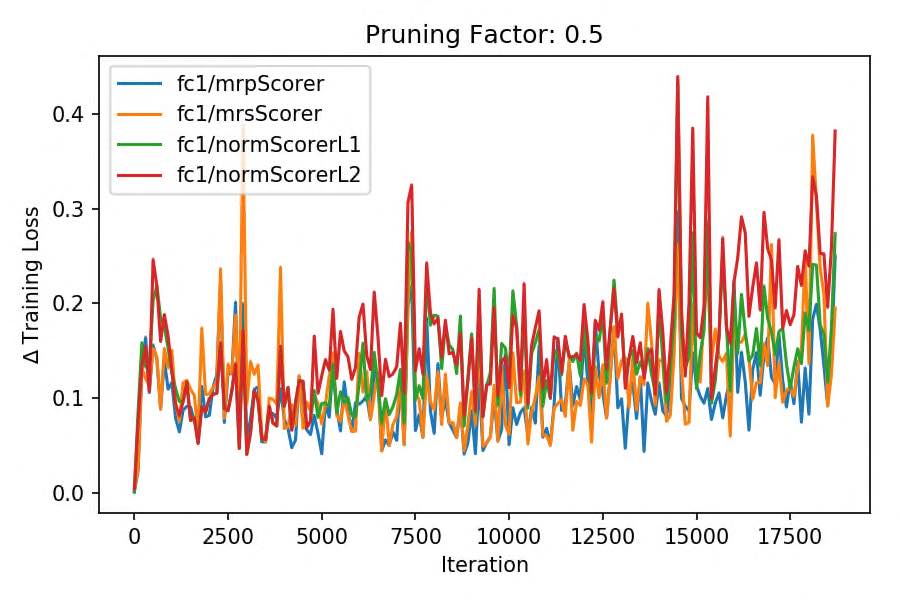}
    \caption{randomScorer not included}
    \label{fig:4.2.b}
  \end{subfigure}
  \caption[Comparison of unit saliency scores: third layer]{Change in the loss after replacing half of the units with their mean values in the first fully connected layer according to various criteria. Measurements are made over the course of the training at every 100 iteration(step).}
  \label{fig:4.2}
\end{figure}

We see a more meaningful picture when we repeat the comparison we made in the previous section for the third layer(first fully connected layer). The results are plotted in Figure \ref{fig:4.2}. Again in \ref{fig:4.2}\subref{fig:4.2.a} we see a significant difference between the baseline and the rest. So our saliency measures all kind of work and we expect them to give reasonable results.  However, if we look closely to \ref{fig:4.2}\subref{fig:4.2.b}, we all see that MRP claims the lowest penalty in general and it is followed by the MRS. We also see that all the scoring functions performs quite similarly within the first 5000 iterations and then the norm based scoring functions start to diverge.

\begin{figure}[ht]
  \begin{center}
    \includegraphics[width=.8\linewidth]{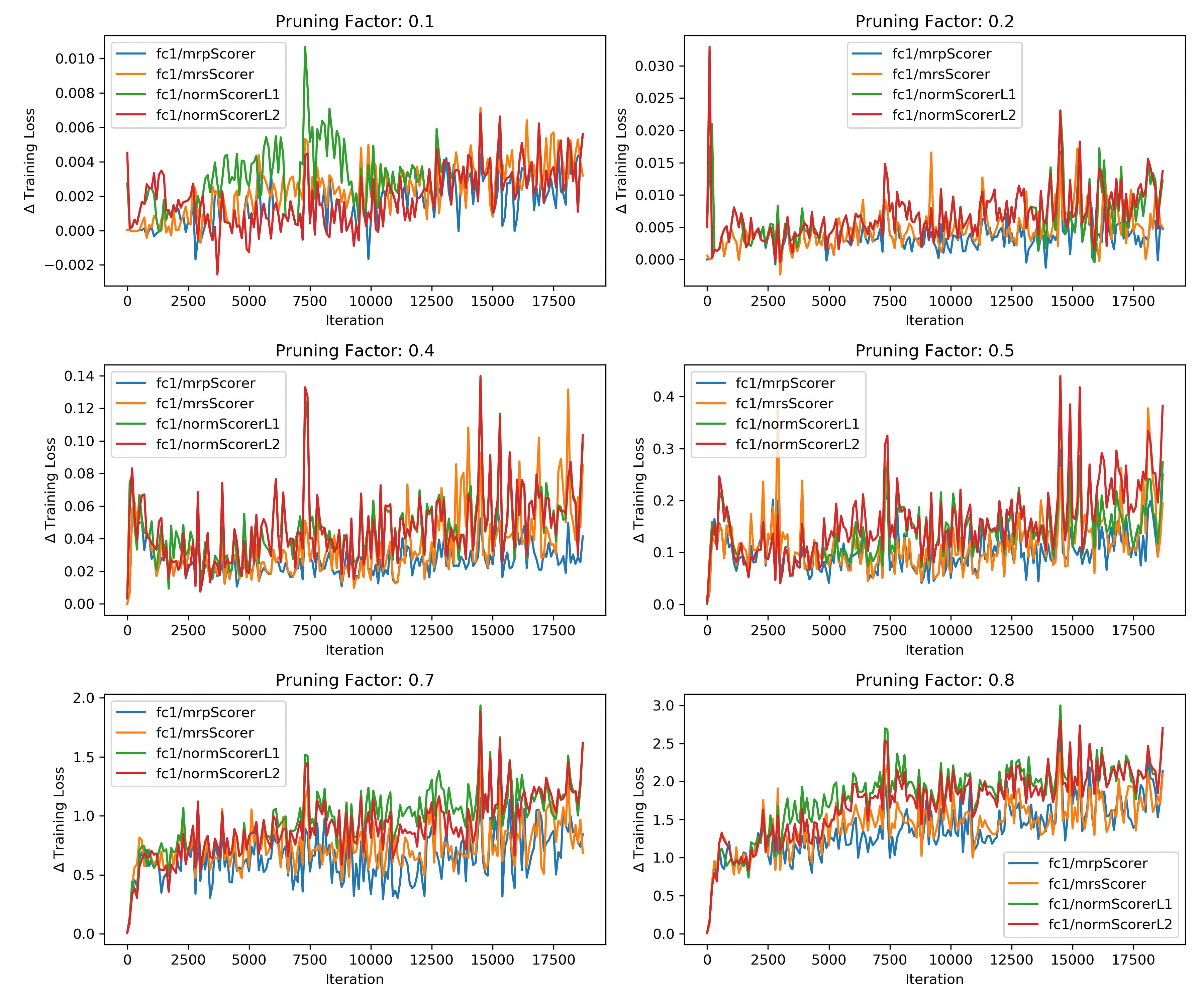}
  \end{center}%
\caption[Comparison of unit saliency scores: for different pruning factors]{We calculate the change in the loss value after pruning the third layer(first fully connected layer) using various saliency measures. Each plot employs a different pruning factor, i.e. fraction of parameters in that layer to be pruned. $\Delta$ Training Loss is calculated on a randomly sampled training set of size 1000.}
\label{fig:4.3}

\end{figure}

Since there are more units(256 units) in the third layer, selected units diverge more compare to the first layer and therefore it is a good candidate for performing additional experiments. In Figure \ref{fig:4.3} we compare the saliency measures for different pruning factors. Pruning factor is a number between 0 and 1 telling us the fraction of the parameters to be pruned. If we have 256 units and a pruning factor of $0.1$, we prune $\lfloor 256/10 \rfloor =25 $ units with lowest saliency scores.

The first thing to observe is that occasionally we improve the loss value through pruning (negative loss change). When the pruning factor is $0.1$, we observe the norm-based scoring functions working significantly worst at the beginning, however, they catch up towards the end and we don't see any significant difference between the 4 saliency measures. Increasing the pruning factor we observe the superiority of the MRP and MRS(blue and orange curves) at pruning. They almost constantly get the lowest scores.

\subsection{Mean replacement might not be the best strategy for Tanh\label{sec4.1.4} }
In the rest of this section, we continue our comparison with Tanh activation function (it was ReLU in the previous subsections). The small convolutional network with Tanh activation gets 0.053 training loss at the end of the training and gets a loss value of 0.056 and an accuracy of 98.31\%  on the test set.

\begin{figure}[ht]
  \begin{center}
    \includegraphics[width=.8\linewidth]{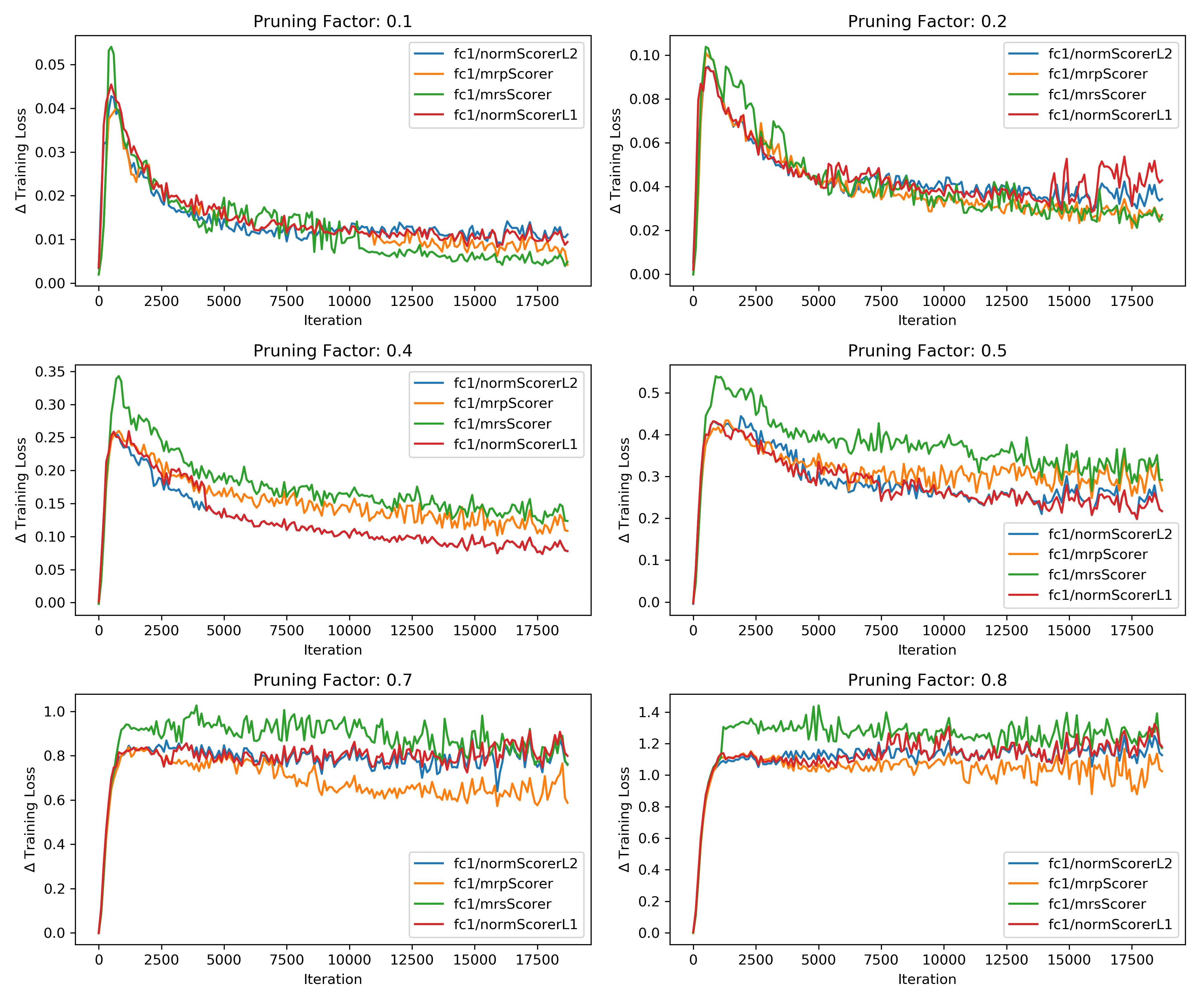}
  \end{center}%
\caption[Comparison of unit saliency scores at pruning a layer with Tanh]{This plot has the exact same setting as the Figure \ref{fig:4.3} expect activation function is changed to Tanh. We calculate the change in the loss value after pruning the third layer(first fully connected layer) using various saliency measures. Each plot employs a different pruning factor and $\Delta$ Training Loss is calculated on a randomly sampled training set of size 1000.}
\label{fig:4.4}
\end{figure}

Figure \ref{fig:4.4} shows how our saliency measures perform at pruning the first fully connected layer(256 units). For pruning factors of $0.1$ and $0.2$ MRP and MRS seem like working better. However, for $0.4$ and $0.5$ norm-based pruning seems like inducing less loss penalty. The picture is slightly different in the last two plot, where the MRP gets the first place and MRS being the last.

More important than the rankings, the main observation here is that we get up to 7x more loss penalty (the positive change in the loss after pruning) compare to Figure \ref{fig:4.3}(ReLU activation). This result is very striking since the final training and test loss values are pretty similar after 10 epochs for both networks. These results suggest that maybe the mean replacement is not the best removal method for all kind of activations. Maybe we shouldn't replace a mostly saturated Tanh unit that generates zeros or ones consistently with a value in the middle.

\subsection{Units with Tanh activations are more tolerating\label{sec4.1.5} }
\begin{figure}[ht]
  \begin{subfigure}{.48\textwidth}
    \centering
    \includegraphics[width=.9\linewidth]{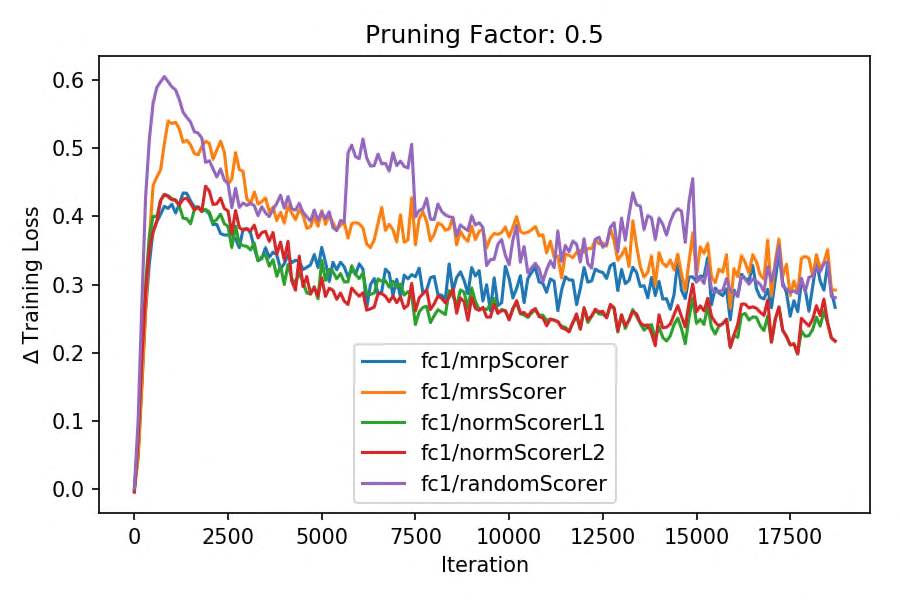}
    \caption{randomScorer included}
    \label{fig:4.5.a}
  \end{subfigure}
  \begin{subfigure}{.48\textwidth}
    \centering
    \includegraphics[width=.9\linewidth]{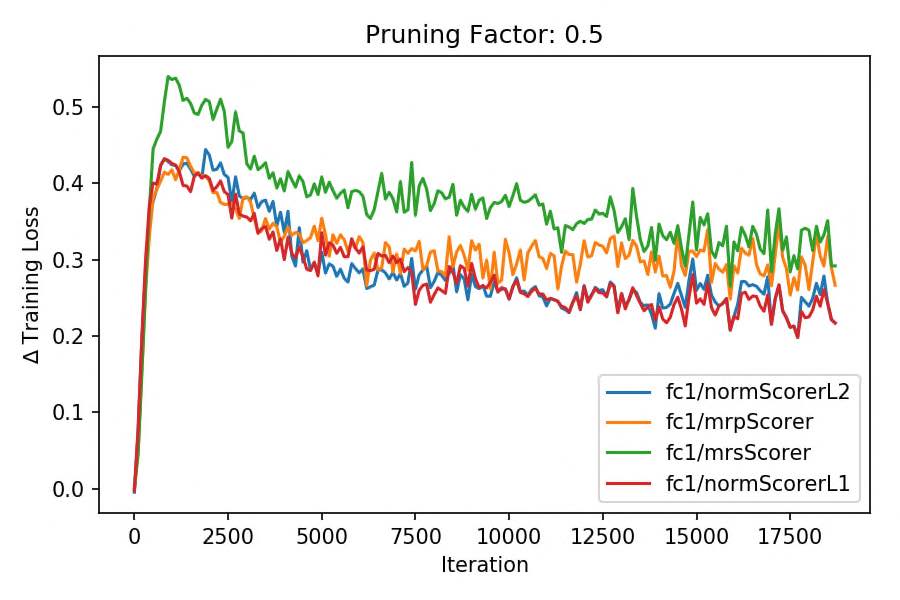}
    \caption{randomScorer not included}
    \label{fig:4.5.b}
  \end{subfigure}
  \caption[Comparison of unit saliency scores at pruning a network with Tanh]{Change in the loss after replacing half of the units with their mean values in the first fully connected layer according to various saliency measures. Measurements are made over the course of the training at every 100 iteration(step) using a randomly selected subset of the training data(size 1000).}
  \label{fig:4.5}
\end{figure}

Figure \ref{fig:4.5} shows how our saliency measures perform at pruning the first fully connected layer(256 units) with a pruning factor of $0.5$. Compare to Figure \ref{fig:4.2}, our baseline(\textit{randomScorer}) induces much smaller loss penalty (around 3x less) when we use Tanh. Our network with Tanh tolerates the absence of a random half of the units much better than the network with ReLu.

We also see that our saliency measures somehow have higher loss penalties over the course of the training. Our saliency measures do not perform significantly better than the baseline (random selection). This observation further supports our previous claim that the removal after bias propagation might not be the best method of pruning for networks with Tanh.

\section{Pruning a small network with MRS\label{sec4.2}}
In this section, we perform unit pruning using the MRS saliency measure. We use the same experimental setup in Section \ref{sec:2.2} (same learning rate, model,random seed,etc.). We slightly changed our pruning schedule reducing the target pruning factors(see Table \ref{table:4.2}). We perform bias propagation before removing units at every pruning round. Removal of a unit is simulated through masking the parameters of the units. We calculate the MRS score using a randomly sampled subset of the training data.

\begin{table}[ht]
\begin{center}
\begin{tabular}{|c|c|c|c|c|}
  \hline
  \textbf{Epoch} & 2 & 3 & 4 & 5  \\
 \hline
\textbf{Fraction $f$} & 0.1 & 0.2 &  0.4 & 0.6  \\
 \hline
\end{tabular}
\end{center}
\caption[Schedule for unit-pruning]{Pruning schedule for the unit pruning experiment on MNIST}
\label{table:4.2}
\end{table}

\begin{figure}[ht]
  \begin{subfigure}{.48\textwidth}
    \centering
    \includegraphics[width=.9\linewidth]{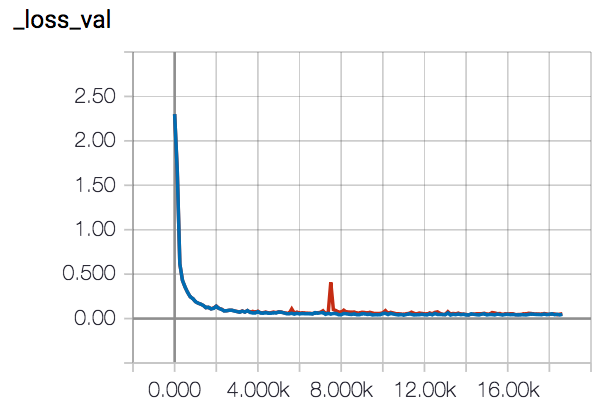}
    \caption{loss}
    \label{fig:4.6.a}
  \end{subfigure}
  \begin{subfigure}{.48\textwidth}
    \centering
    \includegraphics[width=.9\linewidth]{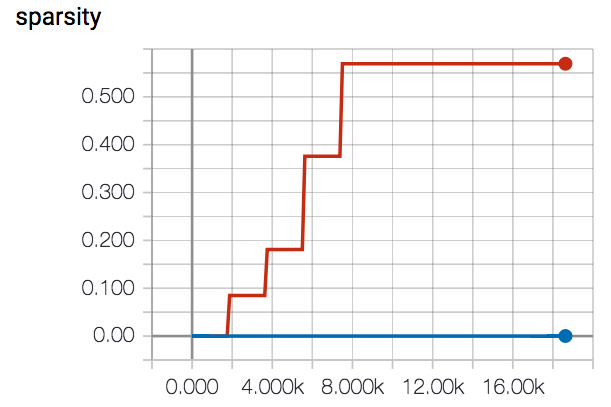}
    \caption{fraction of parameters pruned}
    \label{fig:4.6.b}
  \end{subfigure}
  \caption[Unit Pruning without performance loss]{Loss and fraction of pruned parameters over time. The x-axis represents time and parameterized by the iteration (step) number. The red curve is the run with scheduled pruning, where the blue curve is the original run without any pruning done. We are using the same seed for random number generator to ensure consistency and reproducibility.}
  \label{fig:4.6}
\end{figure}

Figure \ref{fig:4.6} shows the training loss and pruning fractions for 2 different runs of the same experiment: one with pruning(red) and another without(blue). Right after the last two pruning round (fractions $0.4$ \& $0.6$), we observe small peaks in the loss curve (Figure \ref{fig:4.6}\subref{fig:4.6.a}). However, both runs approach to zero after 10 epochs and get 98.42\% accuracy on the test set. The pruned network has only 4019 parameters, where the original network has 20522 (5x reduction!). Note that, since we remove entire units, pruned parameter tensors can be copied into smaller ones. Therefore pruning units don't require sparse representations and therefore we expect to see greater speed ups and smaller memory footprint.

\subsection{Magnitude Colormaps of Parameters\label{sec4.2.1} }
As we did in Section \ref{sec:2.2}, we plot the magnitude colormaps of our layers. Since we use the same exact random seed, the original colormaps should match with the ones in Chapter \ref{chap2}. The reader is encouraged to compare the two pruning strategies and enjoy similarities between the two experiment.

\begin{figure}[ht]
  \begin{center}
    \includegraphics[width=.8\linewidth]{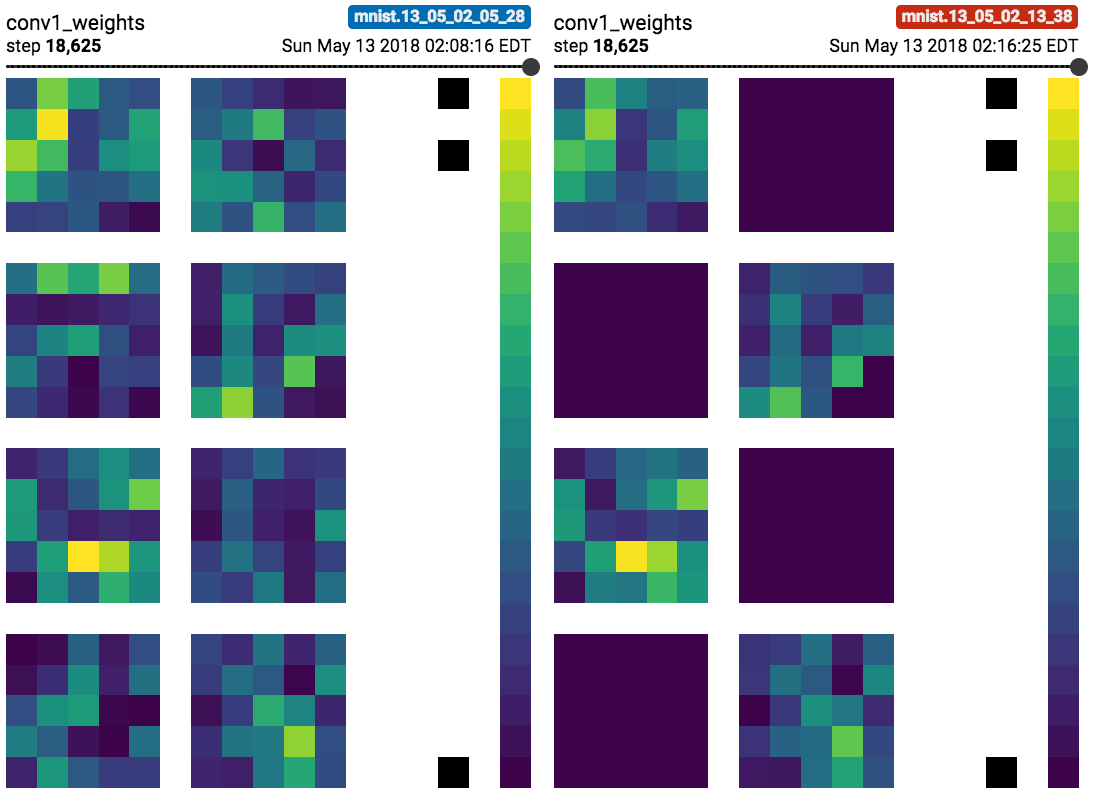}
  \end{center}%
\caption[Magnitude colormap for conv1]{Magnitude colormaps of the weight tensor of the conv1 layer with(right) and without{left} pruning. Colormaps are generated after the training and all values are normalized such that yellow represents the max-value and blue represents the minimum value.}
\label{fig:4.7}
\end{figure}

\begin{figure}[ht]
  \begin{center}
    \includegraphics[width=.8\linewidth]{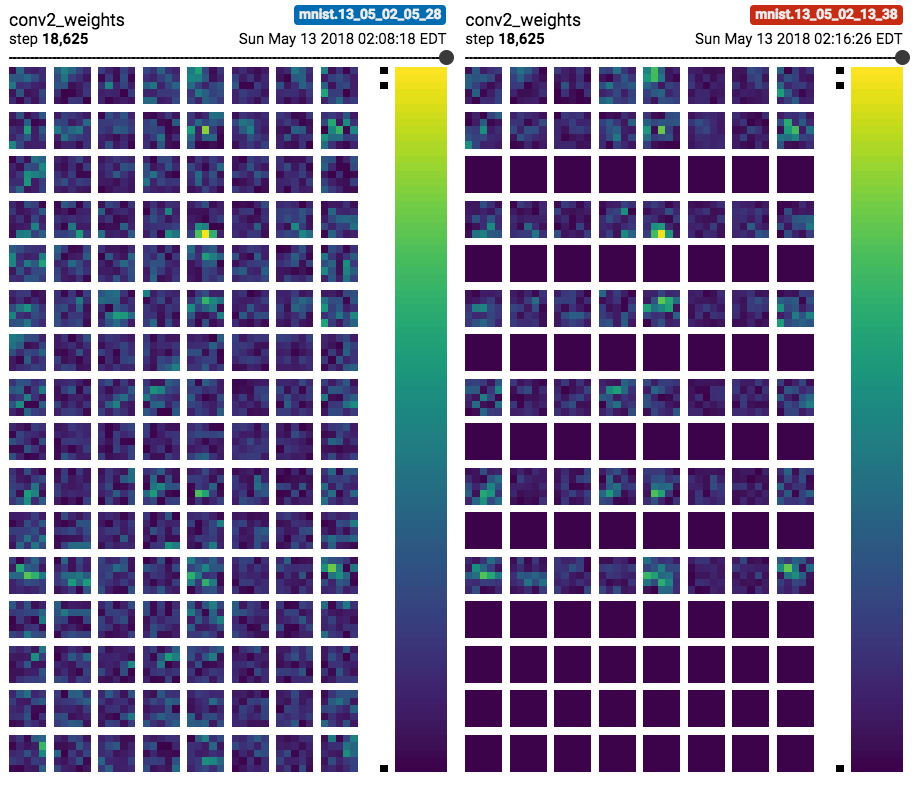}
  \end{center}%
\caption[Magnitude colormap for conv2]{Magnitude colormaps of the weight tensor of the conv2 layer with(right) and without{left} pruning. Colormaps are generated after the training and all values are normalized such that yellow represents the max-value and blue represents the minimum value.}
\label{fig:4.8}
\end{figure}

\begin{figure}[ht]
  \begin{center}
    \includegraphics[width=.8\linewidth]{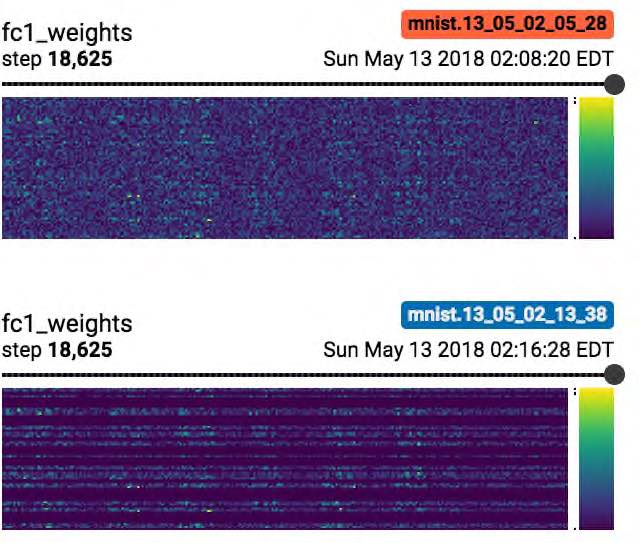}
  \end{center}%
\caption[Magnitude colormap for fc1]{Magnitude colormaps of the weight tensor of the fc1 layer with(right) and without{left} pruning. Colormaps are generated after the training and all values are normalized such that yellow represents the max-value and blue represents the minimum value.}
\label{fig:4.9}
\end{figure}

\begin{figure}[ht]
  \begin{center}
    \includegraphics[width=.8\linewidth]{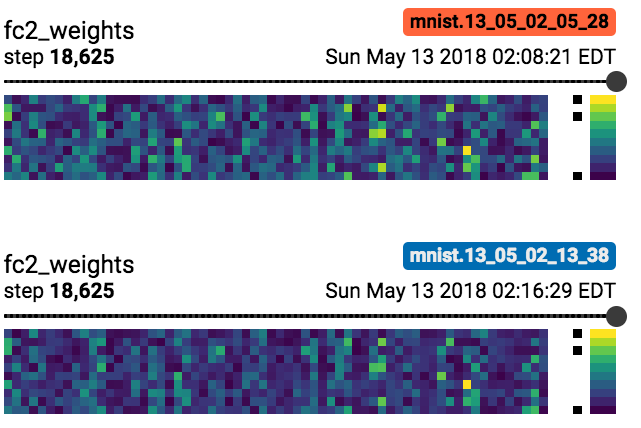}
  \end{center}%
\caption[Magnitude colormap for fc2]{Magnitude colormaps of the weight tensor of the fc2 layer with(right) and without{left} pruning. Colormaps are generated after the training and all values are normalized such that yellow represents the max-value and blue represents the minimum value.}
\label{fig:4.10}
\end{figure}

Colormaps in Figure \ref{fig:4.7},\ref{fig:4.8},\ref{fig:4.9} and \ref{fig:4.10} depict magnitudes after the training for the the first four layers of our network. We observe that the magnitudes of the parameters in the surviving units seem unaffected by the pruning.


\appendix
\chapter{a Pytorch Pruner Library: pytorchpruner \label{app1}}
\texttt{pytorchpruner} is a \href{https://pytorch.org/}{pytorch} package
for pruning neural networks. It is intended for research and its main
objective is not to provide fastest pruning framework, however, it is
relatively efficient and fast. It uses masking idea to simulate pruning
and supports two main pruning strategies. It also implements various
second-order functions to calculate Hessian and Hessian-vector products.\begin{center}
\url{https://github.com/evcu/pytorchpruner}
\end{center}

There are 5 main parts of this library.

\begin{enumerate}
\def\labelenumi{\arabic{enumi}.}
\item
  \textbf{Parameter Pruning (pytorchpruner.scorers)}: has saliency measures
  that return a same-sized-tensor of scores for each parameter in the
  provided parameter tensor.
\item
  \textbf{Unit Pruning (pytorchpruner.unitscorers)}: has saliency measures
  that return a vector of scores for each unit of the provided parameter
  tensor.
\item
  \textbf{Pruners (pytorchpruner.pruners)}: has two different pruner
  engine for the two different pruning strategies (parameter vs unit).
  \texttt{remove\_empty\_filters} function in this file reduces the size
  of the network by copying the parameters into smaller tensors if
  possible.
\item
  \textbf{Auxiliary Modules (pytorchpruner.modules)}: implements
  \texttt{meanOutputReplacer} and \texttt{maskedModule}, the two
  important wrappers for \texttt{torch.nn.Module} instances. The first
  one replaces its output with the mean value if enabled. And the second
  one simulates the pruning layers.
\item
  \textbf{Various first/second-order functionality
  (pytorchpruner.utils)}: implements hessian calculation, Hessian-vector
  product, some search functionality and some other utility functions.
\end{enumerate}

\chapter{Experiment Bootstrapper Library\label{app2}}
This package includes the bootstrap code for generating experiments and includes helper functions/scripts/examples for pytorch training and slurm job scheduling. The release can be cloned from the following public GitHub repo.

\begin{center}
\url{https://github.com/evcu/exp.bootstrp}
\end{center}

The very basic/main idea is:
\begin{itemize}
  \item Create a python script(experiment) which accepts command line arguments
  \item Provide some argument lists and generate SLURM jobs using the cross product of the given argument lists.
  \item Make all above with ease and allow easy reproducibility.
\end{itemize}

A quick walkthrough of library can be found in the repository above.

\subsection{Features}\label{features}
\begin{itemize}
\item
  \href{https://github.com/evcu/exp.bootstrp/blob/master/experiments/exp_utils.py\#L224}{\textbf{read\_yaml\_args}}
  reads conf.yaml and creates a type-checked argParser out of the
  definition. Write the conf, read and overwrite with CLI args.
\item
  \href{https://github.com/evcu/exp.bootstrp/blob/master/experiments/exp_utils.py\#L175}{\textbf{Customizable
  eval-prefixes}} inside yaml file, which enables defining programatic
  eval-able arguments. i.e.~the string `+range(5)' would be evaluated
  and read as a list.
\item
  Current configuration of the experiment
  \href{https://github.com/evcu/exp.bootstrp/blob/master/experiments/create_experiment_jobs.py\#L34}{\textbf{is copied
  to the experiment folder}} such that you can always change experiments
  default\_args after submission
\item
  \href{https://github.com/evcu/exp.bootstrp/blob/master/experiments/exp_utils.py\#L83}{\textbf{ClassificationTrainer}}/\href{https://github.com/evcu/exp.bootstrp/blob/master/experiments/exp_utils.py\#L9}{\textbf{ClassificationTester}}
  which wraps the main training/testing functionalities and provides
  hooks for loggers.
\item
  tensorboardX
  \href{https://github.com/evcu/exp.bootstrp/blob/master/experiments/exp_loggers.py}{\textbf{logging
  utils}} and examples.
\item
  \href{https://github.com/evcu/exp.bootstrp/blob/master/experiments/exp_models.py\#L67}{\textbf{convNetGeneric}}
  implementation
\item
  \href{https://github.com/evcu/exp.bootstrp/blob/master/experiments/default_conf.yaml\#L6}{\textbf{Multiple
  experiment definitions}} through yaml lists.
\end{itemize}

\chapter{Model Definitions\label{app3}}

\section{small CNN for MNIST\label{secA3.1}}
Following model is used for experiments in Chapter \ref{chap2} and Chapter \ref{chap4}. Note that the first layer is wrapped with \textit{meanOutputReplacer} wrapper to simulate mean-replacement during forward pass and/or calculate MRS score.

\begin{lstlisting}[language=Python]
ConvNet_generic(
  (conv1): meanOutputReplacer(
  	module=Conv2d (1, 8, kernel_size=(5, 5), stride=(1, 1))
  	,is_mean_replace=False
  	,enabled=False)
  (conv2): Conv2d (8, 16, kernel_size=(5, 5), stride=(1, 1))
  (fc1): Linear(in_features=256, out_features=32)
  (fc2): Linear(in_features=32, out_features=10))
  \end{lstlisting}

\section{small CNN for CIFAR\label{secA3.2}}
Following model is used for experiments in Chapter \ref{chap3}
\begin{lstlisting}[language=Python]
ConvNet_generic(
  (conv1): Conv2d(3, 8, kernel_size=(5, 5), stride=(1, 1))
  (conv2): Conv2d(8, 16, kernel_size=(5, 5), stride=(1, 1))
  (fc1): Linear(in_features=400, out_features=64, bias=True)
  (fc2): Linear(in_features=64, out_features=10, bias=True)
)
  \end{lstlisting}

\chapter{Loss curves of attacked units\label{app4}}
In this section we provide the loss curves of the experiments shared in Chapter \ref{chap3}. Each plot has a caption telling which figure they match in Chapter \ref{chap3}.

\begin{figure}
  \begin{center}
    \includegraphics[width=.95\linewidth]{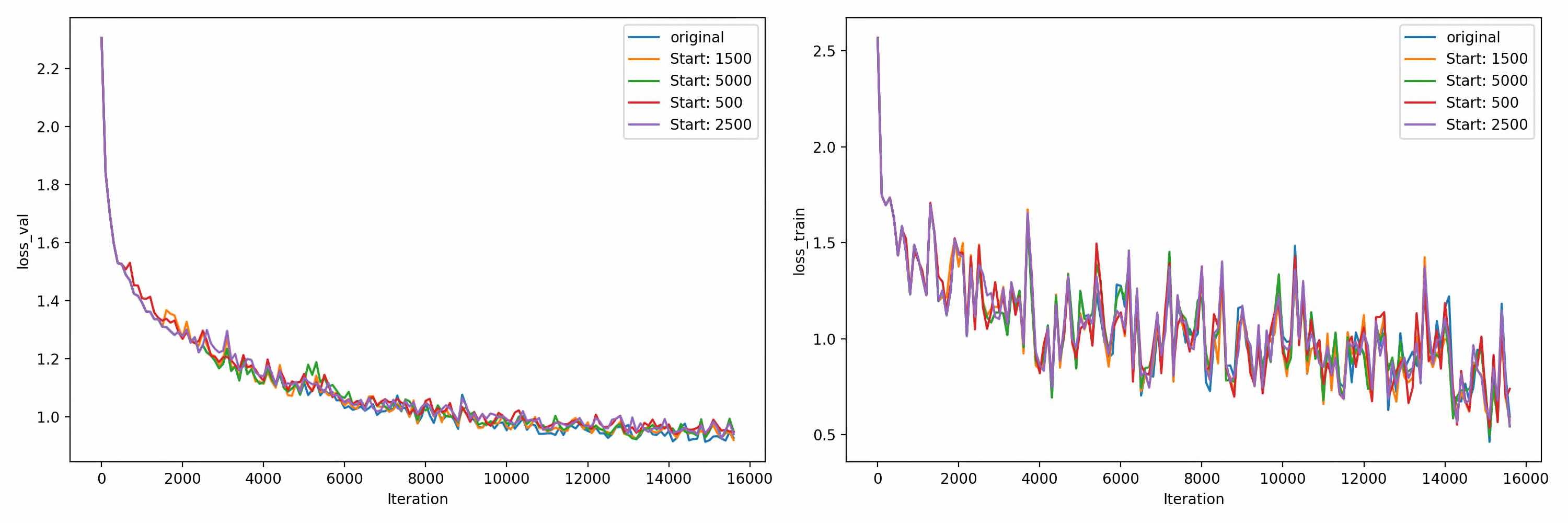}
  \end{center}%
\caption[Loss curves of the Figure \ref{fig:3.3}\subref{fig:3.3.1}]{Loss curves of the Figure \ref{fig:3.3}\subref{fig:3.3.1}}
\label{fig:A4.1}
\end{figure}

\begin{figure}
  \begin{center}
    \includegraphics[width=.95\linewidth]{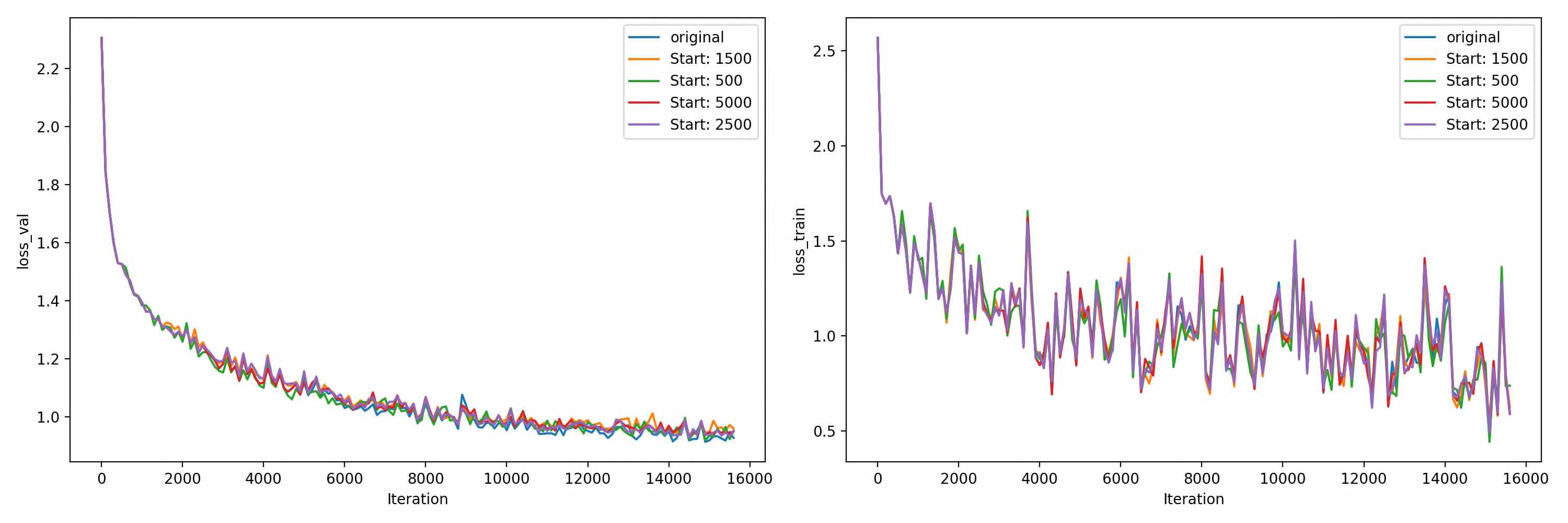}
  \end{center}%
\caption[Loss curves of the Figure \ref{fig:3.3}\subref{fig:3.3.2}]{Loss curves of the Figure \ref{fig:3.3}\subref{fig:3.3.2}}
\label{fig:A4.2}
\end{figure}

\begin{figure}
  \begin{center}
    \includegraphics[width=.95\linewidth]{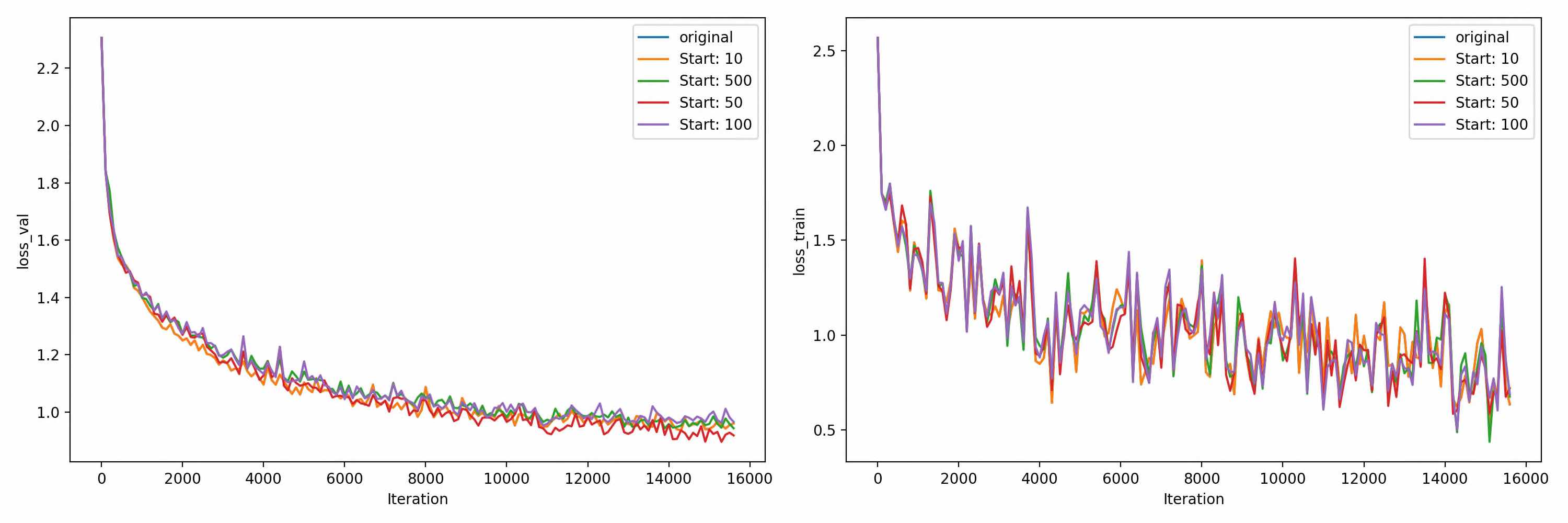}
  \end{center}%
\caption[Loss curves of the Figure \ref{fig:3.4}\subref{fig:3.4.1}]{Loss curves of the Figure \ref{fig:3.4}\subref{fig:3.4.1}}
\label{fig:A4.3}
\end{figure}

\begin{figure}
  \begin{center}
    \includegraphics[width=.95\linewidth]{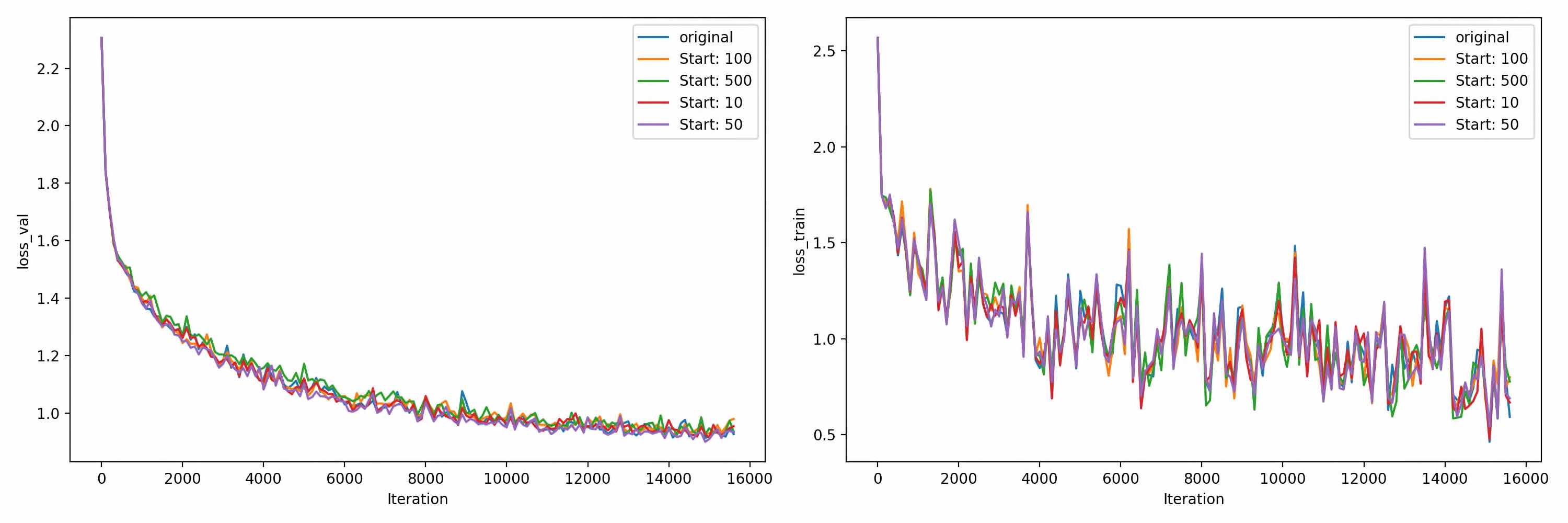}
  \end{center}%
\caption[Loss curves of the Figure \ref{fig:3.4}\subref{fig:3.4.2}]{Loss curves of the Figure \ref{fig:3.4}\subref{fig:3.4.2}}
\label{fig:A4.4}
\end{figure}

\begin{figure}
  \begin{center}
    \includegraphics[width=.95\linewidth]{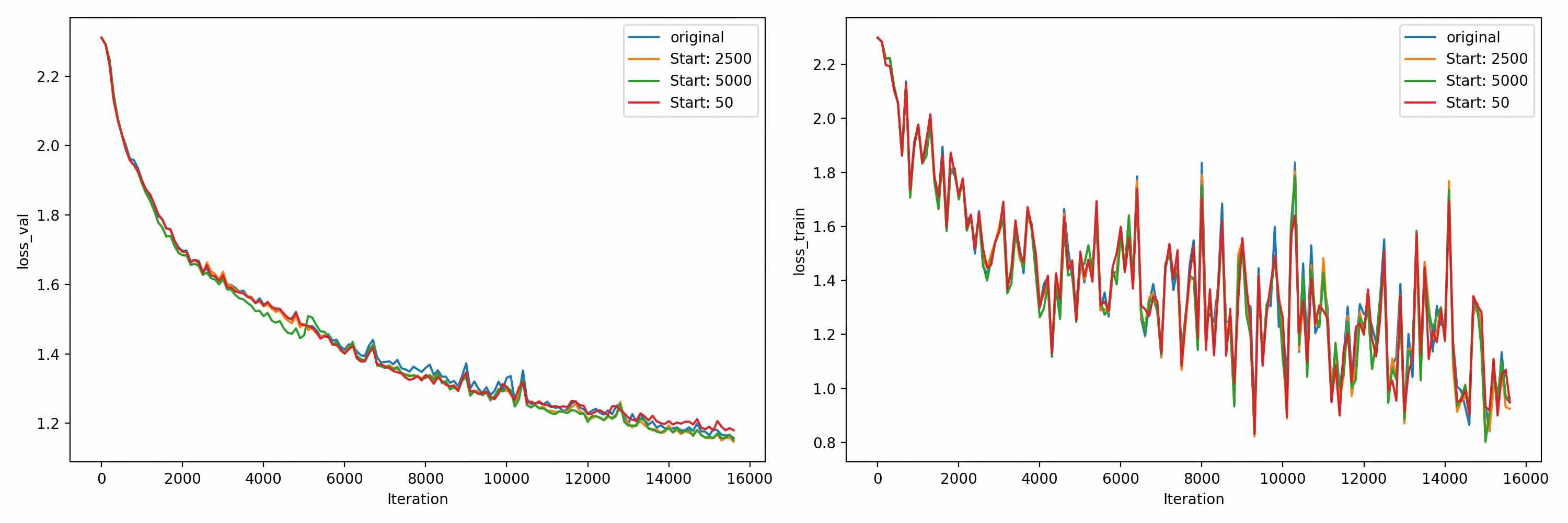}
  \end{center}%
\caption[Loss curves of the Figure \ref{fig:3.7}\subref{fig:3.7.1}]{Loss curves of the Figure \ref{fig:3.7}\subref{fig:3.7.1}}
\label{fig:A4.5}
\end{figure}

\begin{figure}
  \begin{center}
    \includegraphics[width=.95\linewidth]{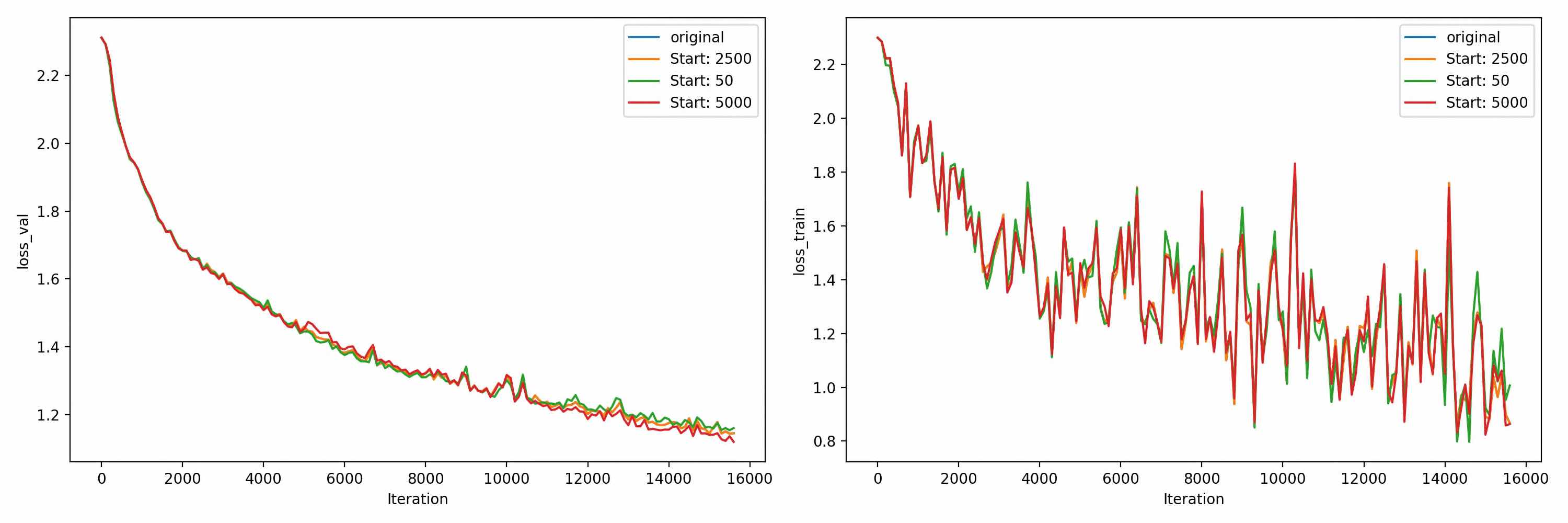}
  \end{center}%
\caption[Loss curves of the Figure \ref{fig:3.7}\subref{fig:3.7.2}]{Loss curves of the Figure \ref{fig:3.7}\subref{fig:3.7.2}}
\label{fig:A4.6}
\end{figure}

\chapter{Unit Saliency Measure Comparison: Additional Plots\label{app5}}
In this section we provide additional results for Section \ref{sec4.1}. We

\begin{figure}
  \begin{center}
    \includegraphics[width=.95\linewidth]{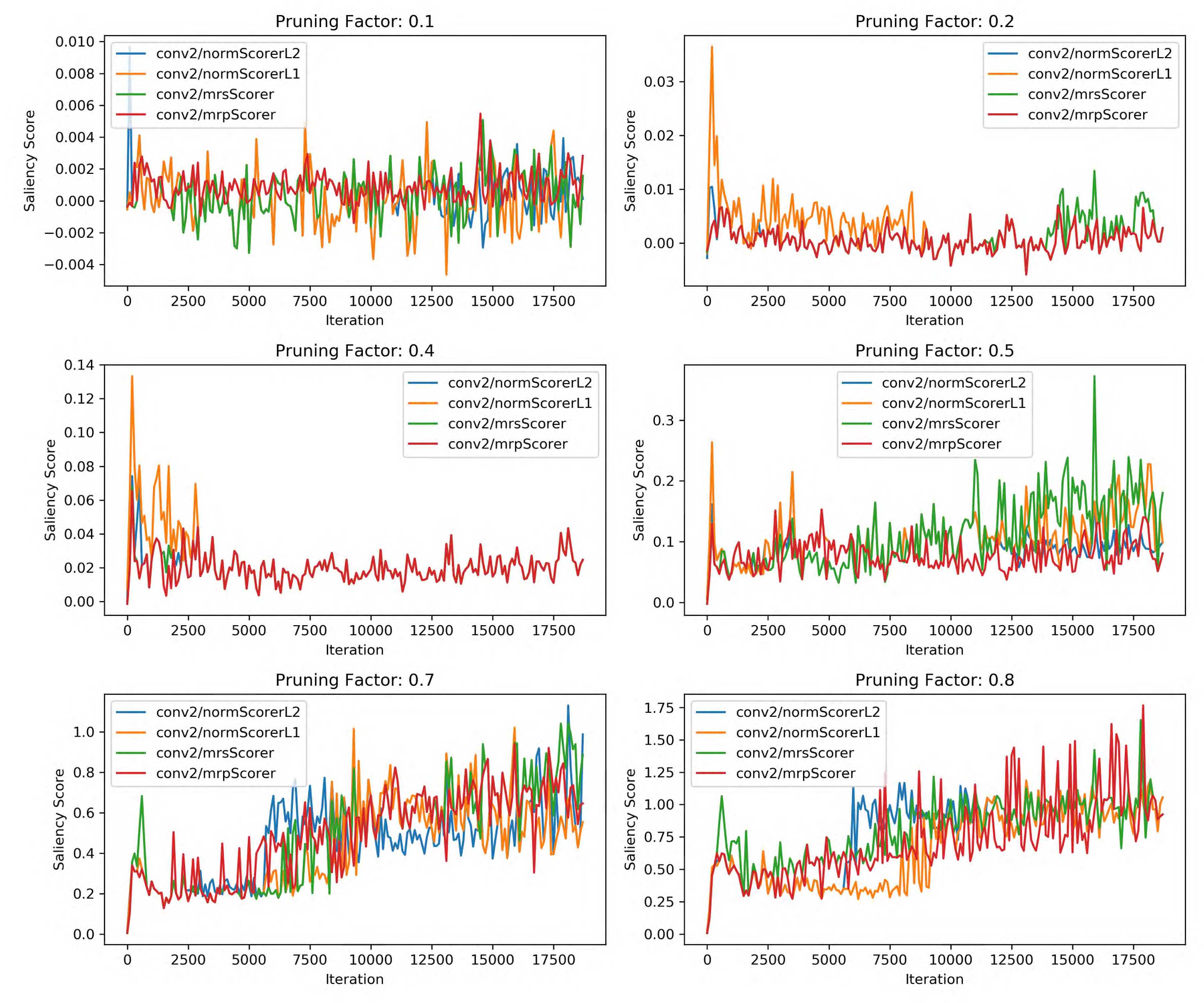}
  \end{center}%
\caption[Comparison of unit saliency scores: second layer (ReLU)]{We calculate the change in the loss value after pruning the second convolutional layer using various saliency measures. Each plot employs a different pruning factor, i.e. fraction of parameters in that layer to be pruned. $\Delta$ Training Loss is calculated on a randomly sampled training set of size 1000.}
\label{fig:A5.1}

\end{figure}

\begin{figure}
  \begin{center}
    \includegraphics[width=.95\linewidth]{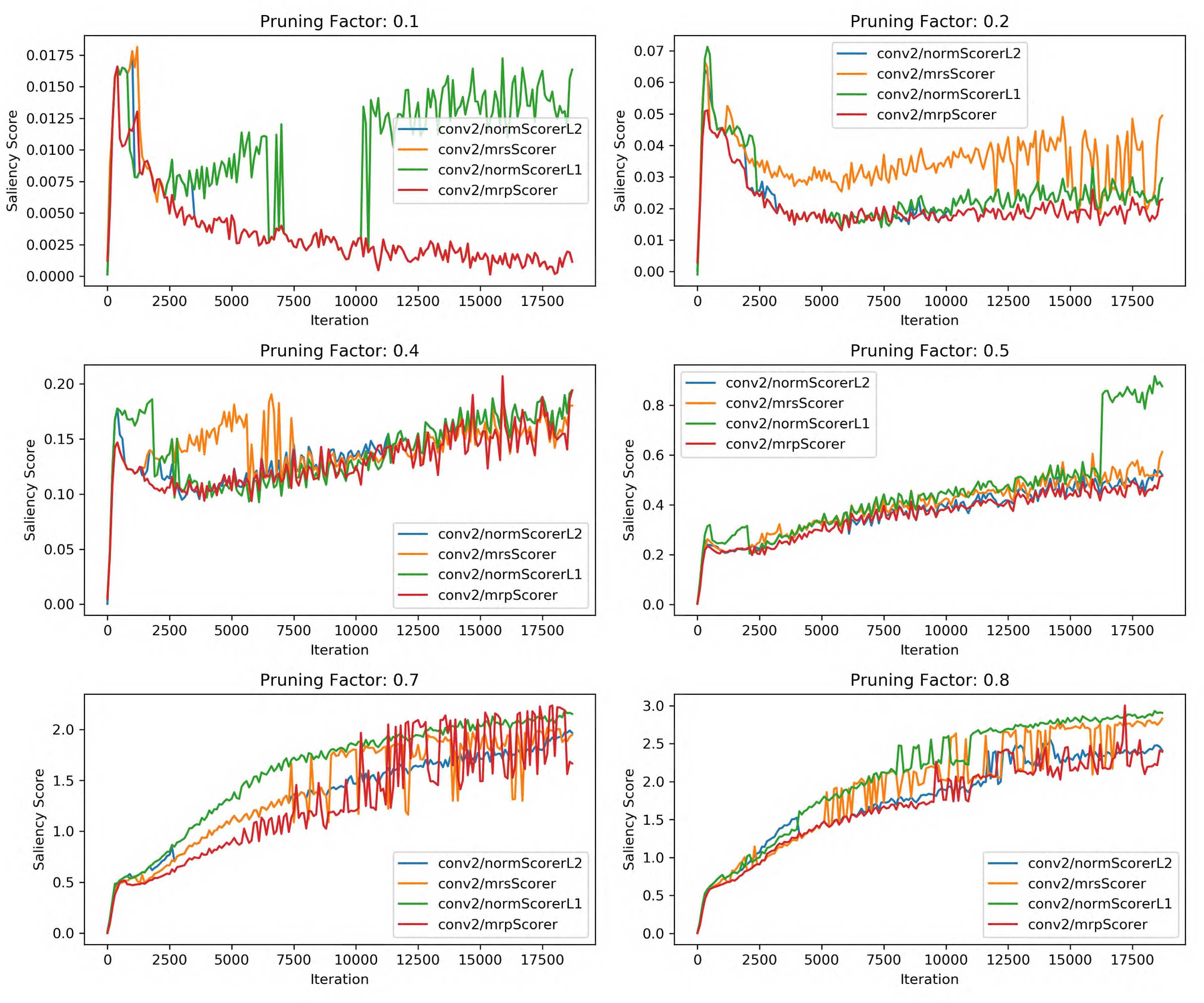}
  \end{center}%
\caption[Comparison of unit saliency scores: second layer (Tanh)]{Same setup as the Figure \ref{fig:A5.1} except units have Tanh activation.}
\label{fig:A5.2}

\end{figure}

\bibliography{thesis}

\end{document}